\def\preprintversion{}
\documentclass[sn-mathphys-num]{sn-jnl}

\usepackage{graphicx}%
\usepackage{multirow}%
\usepackage{amsmath,amssymb,amsfonts}%
\usepackage{amsthm}%
\usepackage{mathrsfs}%
\usepackage{mathtools}
\usepackage[title]{appendix}%
\usepackage{xcolor}%
\usepackage{textcomp}%
\usepackage{manyfoot}%
\usepackage{booktabs}%
\usepackage{algorithm}%
\usepackage{algorithmicx}%
\usepackage{algpseudocode}%
\usepackage{listings}%
\usepackage{mathtools}
\usepackage{comment}
\usepackage{rotating} 
\usepackage{tabularx}
\usepackage{array}
\usepackage{booktabs}
\usepackage{rotating}
\usepackage{cleveref}

\theoremstyle{thmstyleone}%
\theoremstyle{thmstyletwo}%

\theoremstyle{thmstylethree}%

\usepackage{graphicx}

\newif\ifproduction
\productiontrue
\newif\ifincludesupplement
\ifdefined\preprintversion
  \includesupplementtrue
  \productionfalse
\fi
\ifincludesupplement
  \usepackage{bbm}
  \usepackage{siunitx}
\fi
\ifproduction
  \usepackage[figuresonly,nolists,nomarkers]{endfloat}

  \makeatletter
  \AtBeginDelayedFloats{%
    \renewenvironment{figure}[1][]{\par\begingroup\def\@captype{figure}}{\par\endgroup}%
    \renewenvironment{figure*}[1][]{\par\begingroup\def\@captype{figure}}{\par\endgroup}}
  \makeatother
\fi

\begin{document}
\title[]{Soft yet Effective Robots via Holistic Co-Design}


\author*[1,2,7]{\fnm{Maximilian} \sur{Stölzle}}\email{mstolzle@mit.edu}
\equalcont{These authors contributed equally to this work.}

\author[3]{\fnm{Niccolò} \sur{Pagliarani}}\email{niccolo.pagliarani@santannapisa.it}
\equalcont{These authors contributed equally to this work.}

\author[4,5]{\fnm{Francesco} \sur{Stella}}\email{f.stella@embodiedai.ch}

\author[5]{\fnm{Josie} \sur{Hughes}}\email{josie.hughes@epfl.ch}

\author[6]{\fnm{Cecilia} \sur{Laschi}}\email{mpeclc@nus.edu.sg}

\author[7]{\fnm{Daniela} \sur{Rus}}\email{rus@csail.mit.edu}

\author[3]{\fnm{Matteo} \sur{Cianchetti}}\email{matteo.cianchetti@santannapisa.it}

\author[2]{\fnm{Cosimo} \sur{Della Santina}}\email{C.DellaSantina@tudelft.nl}

\author[1]{\fnm{Gioele} \sur{Zardini}}\email{gzardini@mit.edu}

\affil*[1]{\orgdiv{Laboratory for Information \& Decision Systems}, \orgname{Massachusetts Institute of Technology}, \orgaddress{ 
\city{Cambridge}, \postcode{02139}, \state{MA}, \country{USA}}}
\affil[2]{\orgdiv{Cognitive Robotics}, \orgname{Delft University of Technology}, \orgaddress{
\street{Mekelweg 2}, 
\city{Delft}, \postcode{2628 CD}, \country{Netherlands}}}

\affil[3]{\orgdiv{The BioRobotics Institute}, \orgname{Scuola Superiore Sant'Anna}, \orgaddress{\city{Pisa}, \postcode{56025}, \country{Italy}}}
\affil[4]{\orgname{Embodied AI AG}, \orgaddress{\city{Lausanne}, \postcode{1015}, \country{Switzerland}}}
\affil[5]{\orgdiv{CREATE Lab}, \orgname{EPFL}, \orgaddress{\city{Lausanne}, \postcode{1015}, \country{Switzerland}}}

\affil[6]{\orgdiv{Advanced Robotics Centre, Department of Mechanical Engineering}, \orgname{National University of Singapore},\orgaddress{ 
\postcode{117575}, \country{Singapore}}. On leave from \orgdiv{The BioRobotics Institute}, \orgname{Scuola Superiore Sant'Anna}, \orgaddress{\city{Pisa}, \postcode{56025}, \country{Italy}}}

\affil[7]{\orgdiv{Computer Science and Artificial Intelligence Laboratory}, \orgname{Massachusetts Institute of Technology},\orgaddress{ 
\city{Cambridge}, \postcode{02139}, \state{MA}, \country{USA}}}


\abstract{

Soft robots promise inherent safety via their material compliance for seamless interactions with humans or delicate environments.
Despite progress, the field struggles to balance task-specific performance with broader factors like durability and manufacturability—a difficulty that we find is compounded by traditional sequential design processes with their lack of feedback loops.
In this perspective, we review emerging co-design approaches that simultaneously optimize the soft robot’s body and brain, enabling the discovery of unconventional designs highly tailored to the given tasks.
Their adoption is limited by narrow objectives, gaps between simulated and real performance, and computational cost.
To address these challenges, we propose a holistic co-design framework that incorporates a broader range of design values, integrates real-world prototyping to refine evaluations, and boosts efficiency through surrogate metrics and model-based control strategies.
Finally, we outline research priorities in design priors and metrics, AI-assisted evaluation, and balancing computational refinement with physical testing and safety with performance.
}




\maketitle

\section{Introduction}

Robots are increasingly moving from structured industrial settings to human-centered environments such as homes and public spaces~\cite{todd1986fundamentals,nahavandi2019industry,royakkers2015literature}. In these contexts, physical intelligence can enable safer and more adaptive interaction than conventional rigid robotic systems~\cite{villani2018survey,zhao2024potential,haddadin2009requirements}. \emph{Soft robotics} offers a transformative approach by embedding safety directly into a robot’s intentionally compliant and deformable materials and/or structure~\cite{rus2015design, laschi2016soft, della2021softencyclopedia, yasa2023overview}, reducing reliance on complex computational algorithms for ensuring safety and instead leveraging the embodied intelligence of compliant bodies~\cite{pfeifer2007self, mengaldo2022concise}, particularly during physical interaction with the environment. In this perspective, we focus on the co-design of \emph{soft robots} according to this definition, in contrast to the (co-)design of predominantly rigid systems.
This inherent compliance enables safe human interaction and operation in sensitive settings such as personal assistance, caregiving, and handling delicate items~\cite{abidi2017intrinsic, pasquier2025study}. 
However, soft robot development is inherently complex, requiring the seamless integration of materials, geometry, actuation, sensing, compliant continuum dynamics, perception, and control systems. 
It is notably difficult to predict how morphological changes affect closed-loop motion, as the design space is much larger than for rigid robots, and the capabilities of control and task \& motion planning are tightly coupled to morphology. 
Crucially, the body is not a fixed ``plant'': morphology and materials directly shape the effective kinematics/dynamics, the informativeness of sensing, and the feasibility of stable control, so even small design changes can significantly change closed-loop behavior.
Consequently, we observe that traditional sequential design processes~\cite{van2020delft}—which move from conceptual design through mechatronic design to the development of autonomy and control systems—struggle with this complexity, leading to robots that exhibit imprecise, oscillatory motions, limited payload capacity, and inadequate force output~\cite{iida2011soft, cianchetti2013stiff, mazzolai2022roadmap, majidi2014soft, hawkes2017soft}.

\begin{figure*}[tb]
    \centering
    \ifproduction\else
      \includegraphics[width=\textwidth]{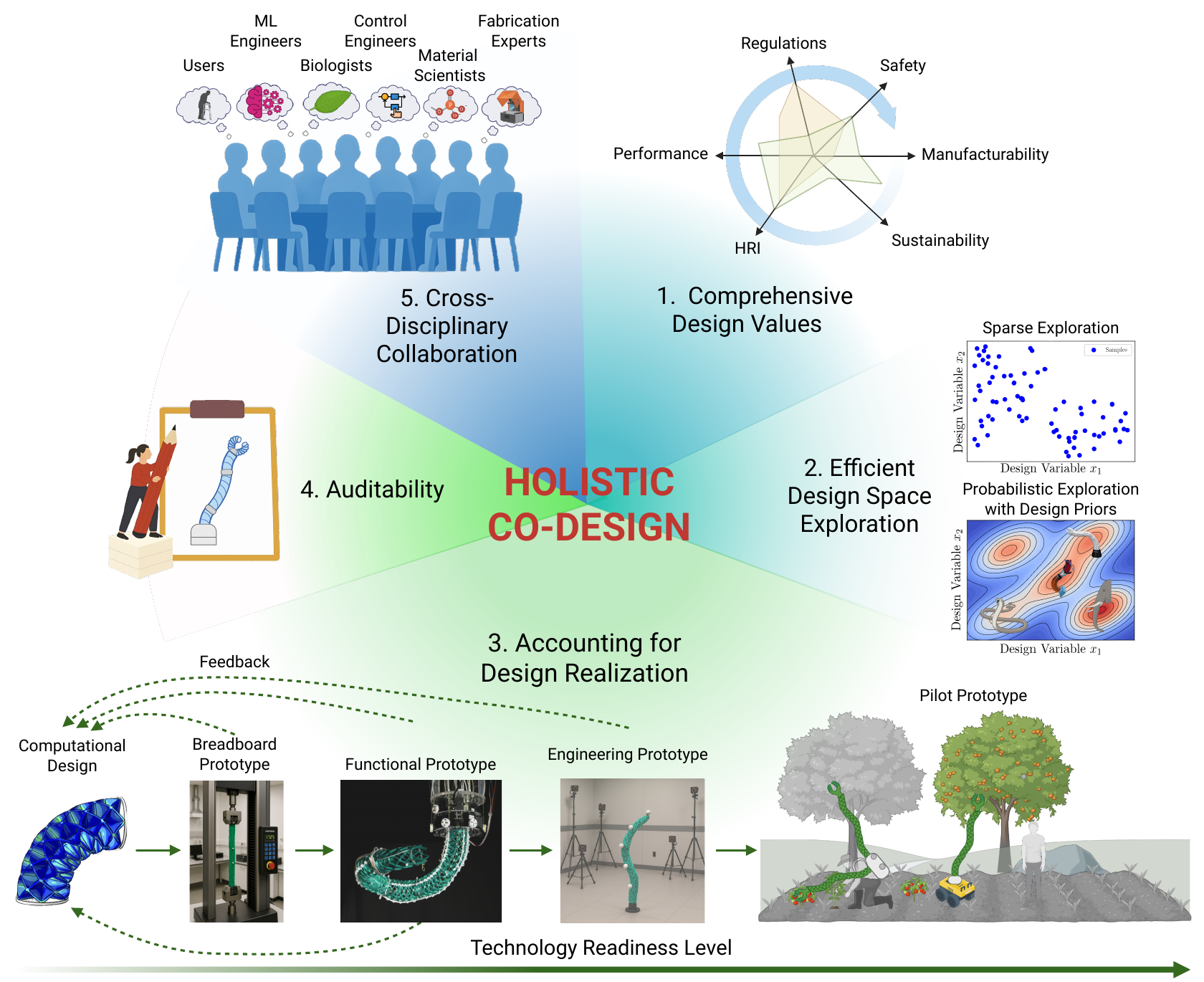}
    \fi
    
    \caption{
    \textbf{Holistic Co-Design of Soft Robots.}
    The five pillars of holistic co‑design are (1) Incorporating comprehensive design requirements and values directly into multi‑objective optimization; (2) Efficiently exploring the full design space by melding design priors (e.g., biological inspirations, existing solutions) with computationally efficient co‑design routines; (3) Explicitly accounting for design realization (e.g., prototyping and testing) via a probabilistic treatment of evaluation metrics and by formalizing the refinement‑vs‑realization trade‑off—enabling targeted prototyping to reduce metric uncertainty and narrow the sim‑to‑real gap; 
    (4) Ensuring auditability to preserve design knowledge and guarantee reproducibility.
    (5) Fostering cross‑disciplinary collaboration and involving all relevant stakeholders in defining design values and providing iterative feedback;
    These pillars are inherently interconnected (see green arrows)—for example, stakeholders and the design team co‑define design requirements, certain design values (e.g., HRI) demand evaluation in the real world (i.e., enabled by realization), and both the exploration of the design space and the fulfillment of design requirements remain fully auditable.
    }
    \label{fig:overview}
\end{figure*}

\paragraph*{Related Work and Its Limitations}
Co-design strategies have proven effective for multi-objective problems in hierarchical systems~\cite{zardini2023co}, and recent studies have introduced algorithms for jointly optimizing soft robot morphology and control~\cite{van2018spatial, spielberg2019learning, chen2020design, bhatia2021evolution, spielberg2021co, wang2023preco, medvet2021biodiversity, wang2023curriculum, junge2022leveraging, legrand2023reconfigurable, wang2023diffusebot, navez2024contributions}. However, several limitations still hinder broader applicability.

Most frameworks remain computationally expensive because they rely on high-dimensional morphology representations—such as voxel- or particle-based discretizations~\cite{spielberg2019learning, medvet2021biodiversity, medvet2022impact, wang2023curriculum, legrand2023reconfigurable, wang2023softzoo, wang2023preco, wang2023diffusebot}—and optimization strategies that scale poorly with dimensionality and non-convexity, including evolutionary strategies~\cite{chen2020design, rieffel2014growing, hiller2012automatic, bhatia2021evolution, medvet2021biodiversity, medvet2022impact, li2025generating} and first-order gradient-based strategies~\cite{spielberg2019learning, wang2023softzoo, wang2023diffusebot}. Training a controller for each candidate design, often through reinforcement learning~\cite{bhatia2021evolution, wang2023curriculum, wang2023softzoo, wang2023preco}, further increases cost, especially when evaluation depends on high-fidelity closed-loop simulations~\cite{spielberg2019learning, medvet2021biodiversity, medvet2022impact, wang2023curriculum, legrand2023reconfigurable, wang2023softzoo, wang2023preco, wang2023diffusebot}.

Many studies optimize only a narrow set of easily computable metrics, such as locomotion speed~\cite{wang2023diffusebot} or workspace~\cite{guan2023trimmed}, while overlooking broader objectives that determine real-world viability, including manufacturability~\cite{rieffel2005crossing, howison2020reality, kim2025generative}, safety~\cite[Chapter 3]{stolzle2025phdthesis}, cost, environmental sustainability, usability, and regulatory compliance~\cite{junge2022leveraging}.

Even when such values are included, they are usually treated as fixed simulation-only quantities. Realization is seldom integrated into the design loop~\cite{junge2022leveraging}, limiting feedback from fabrication and field testing. As a result, uncertainty in simulation-based evaluations is rarely quantified~\cite{chen2020design}, and designs that appear optimal in silico may fail once physically realized—a persistent manifestation of the \emph{reality} / sim-to-real gap~\cite{mouret2017realitygap, schaff2023sim}.

Promising yet currently disjointed directions include category theory--based co-design for complex systems~\cite{zardini2021co,zardini2022task,milojevic2025codei}, generative models for reduced-order optimization~\cite{zeng2022lion, song2024morphvae, wang2023diffusebot}, reduced-order dynamic models for model-based control~\cite{hewing2020learning, alora2023data, della2023model, stolzle2024input, alkayas2025soft}, inexpensive metrics such as observability, controllability, and safety~\cite{griffith1971observability, zheng2019controllability, haddadin2013towards, iso2016collaborative}, and uncertainty-aware methods in Bayesian optimization, reinforcement learning, and composable co-design~\cite{hernandez2014predictive, garnett2023bayesian, sutton1998reinforcement, huang2025uncertainty, furter2026composable}.

Therefore, the core scientific challenge is system-level co-design: jointly shaping all components so that the resulting soft robot meets task requirements and design values such as safety, manufacturability, cost, and usability under substantial modeling/metric uncertainty and realization constraints.

\paragraph*{Proposed Framework}
This perspective outlines a holistic co-design framework that integrates design components, stakeholder values, design processes, and optimization strategies through five core advances, as illustrated in \Cref{fig:overview}.

(1) The framework broadens the range of considered objectives and constraints to include safety, fabrication and operational costs, environmental impact, and regulatory compliance.

(2) It improves computational co-design efficiency through reduced-order design spaces decoded into full morphologies, co-optimized reduced-order dynamical models, fast surrogate metrics, and model-based control methods that reduce reliance on costly RL training~\cite{howison2020reality, wang2023diffusebot, armanini2023soft, mathew2025reduced, liu2024physics, stolzle2024input, valadas2025learning, alkayas2025soft, alkayas2025structure, navez2025modeling, della2023model,stolzle2024input}.

(3) The framework incorporates purposeful physical prototyping to reduce uncertainty in computational evaluations and the corresponding sim-to-real gap, treating co-design probabilistically and formalizing the trade-off between computational refinement and physical realization across varying TRLs~\cite{dubied2022sim, NASA_TRL}.

(4) The framework supports reproducibility~\cite{baines2024need} by maintaining an auditable design trail.

Finally (5), structured stakeholder participation ensures that diverse values and requirements are reflected throughout the design process.

This paradigm holds promise for human-centric domains, such as caregiving, which we emphasize in this perspective. However, the framework could also have an impact in other areas, including space robotics~\cite{zhang2023progress}, agriculture~\cite{elfferich2022soft, armanini2024soft}, and inspection and maintenance~\cite{milana2022soft, groo2024toward}.

We first review existing soft robot design processes and introduce the proposed holistic co-design framework. We then outline computational metrics and illustrate a possible application through an indoor tomato harvesting case study, before discussing conclusions, limitations, and open challenges.

\section{The Past and Present Soft Robot Design Process}\label{sec:past_and_present_design_process}

\begin{figure}[tb]
    \centering
    \ifproduction\else
      \includegraphics[width=1\linewidth]{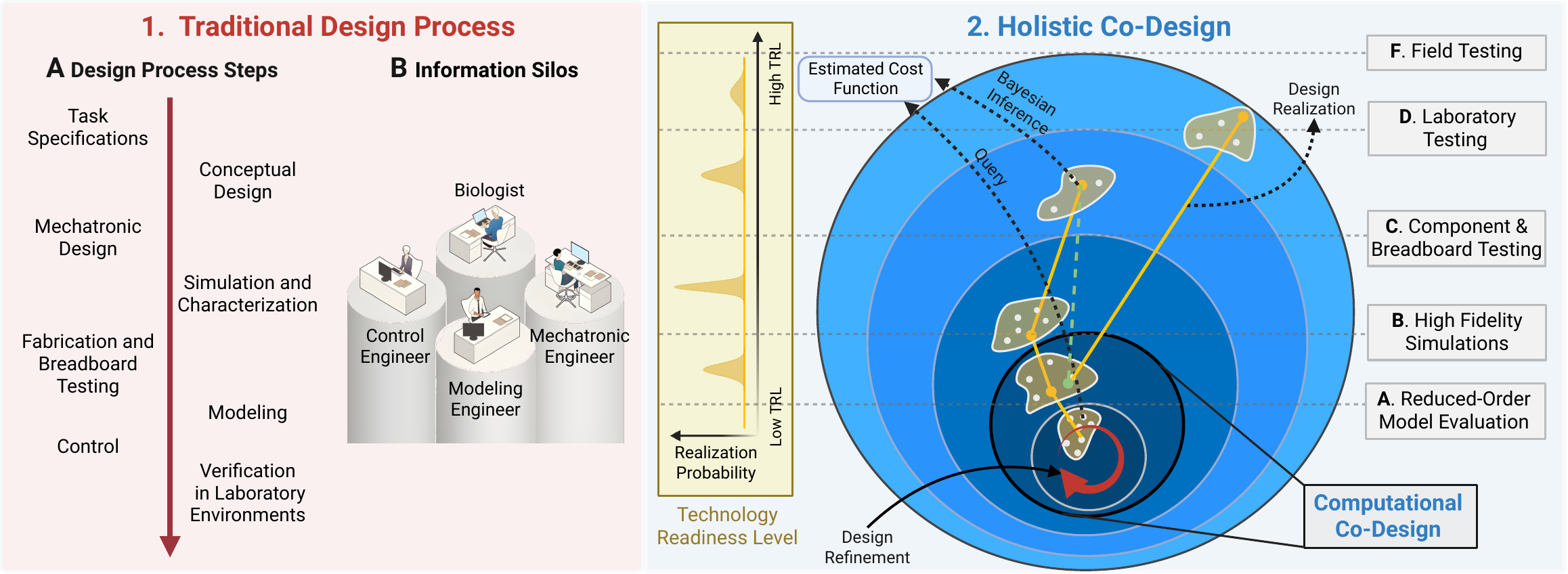}
    \fi
    
    \caption{
        \textbf{The traditional soft robot design process vs. holistic co-design.}
        \textit{Left:} A traditional sequential design cycle~\cite{van2020delft} applied to soft robots. This sequential workflow lacks iterative feedback loops, leading to information silos and preventing regular, bidirectional sharing of insights and data across the development team and the relevant stakeholders.
        \textit{Right:} A holistic co-design framework that complements computational refinement with prototyping across various levels of design readiness - visualized as ``onion'' layers in this graphic. Solid yellow lines denote the prototyping pathway, while dashed orange arrows illustrate how empirical findings continuously inform and update the estimates $\hat{c}_j$ for the real‑world design values $c_j$.
    }
    \label{fig:holistic_co_design}
\end{figure}

\subsection{Traditional Design Process}\label{sub:traditional_design_process}

We depict the traditional design cycle~\cite{roozenburg1995product, van2020delft}, which is applied within soft robotics, in Fig.~\ref{fig:holistic_co_design}~(Left).
It begins with task specifications and proceeds to a conceptual design, followed by mechatronic development. The open-loop behavior of the design is verified through simulation, fabrication, and testing on prototypes of Technology Readiness Levels (TRLs)~\cite{NASA_TRL}. 
Typically, control design and reduced-order modeling occur only after validating a viable soft robotic prototype, enabling verification of closed-loop behavior within laboratory settings. Furthermore, steps (2)-(4) often involve iterative cycles of diverging and converging phases~\cite{feldhusen2013pahl}. During the \emph{diverging} phase, new design concepts and implementations are generated. Conversely, the \emph{converging} phase evaluates these through simulations or prototype tests to identify, for example, via selection matrices~\cite{ulrich2016product}, promising designs for further development.
The traditional design cycle has several deficiencies:
(i) Iterations that could close the loop are often prohibitively expensive, with no systematic way to feed performance insights back into earlier stages.
(ii) Key downstream constraints, such as manufacturability and regulatory compliance, are often neglected early on.
(iii) Sequential workflows across design, perception, modeling, and control create silos and weak feedback, hindering iterative refinement.

\subsection{First Steps Towards Co-Design of Soft Robots}\label{sub:co_design_review}

Given the complexity of soft robots and the limitations of traditional design processes, the community has recently begun exploring how co-design algorithms could support soft robot development~\cite{spielberg2019learning, cianchetti2021embodied, bhatia2021evolution, van2022co, wang2023curriculum, wang2023preco, wang2023diffusebot, junge2022leveraging}.

The term ``\textit{co}-design'' conveys several distinguishing attributes compared to traditional design approaches, including a \textit{co}mpositional and hierarchical nature, a \textit{co}llaborative approach involving relevant stakeholders, and a \textit{co}ntinuous process of design improvement~\cite{junge2022leveraging, zardini2023co}. In soft robotics, \emph{co-design} refers to methods that jointly optimize the soft robot’s body and brain to meet task requirements and design values~\cite{spielberg2019learning, chen2020design, junge2022leveraging}. Joint physical and control design is also studied within multidisciplinary design optimization~\cite{allison2014multidisciplinary}.

Specifically, we argue that the general definition of co-design by Zardini \textit{et al.}~\cite{zardini2023co} naturally extends to soft robots. Let $f\in\mathcal{F}$ denote \emph{functionalities} (e.g., task performance, safety, durability, or controllability) that are typically to be maximized, and let $r\in\mathcal{R}$ denote \emph{resources} (e.g., unit cost or environmental impact) that should be kept as low as possible.
We consider a parameterized design $x\in\mathcal{X}$, also referred to as an implementation~\cite{zardini2023co}, together with mappings $\operatorname{prov}:\mathcal{X}\to\mathcal{F}$ and $\operatorname{req}:\mathcal{X}\to\mathcal{R}$ that assign to each design the provided functionality and required resources, respectively.

In practice, the co-design variables are not arbitrary: only some parameter
values correspond to physically meaningful, manufacturable, and deployable
systems. We therefore distinguish between a raw parameterization domain
$\mathcal X_0$ and a strict feasible design set $\mathcal X\subseteq\mathcal X_0$.
The set $\mathcal X$ collects non-negotiable feasibility requirements, such as
valid geometry, material admissibility, manufacturability limits, kinematic
compatibility, regulatory exclusion rules, and other conditions that should not
be traded against performance. When useful, these strict constraints can be
written explicitly as
\begin{equation}
    \mathcal X
    =
    \left\{
        x\in\mathcal X_0
        \;\middle|\;
        h(x)\leq 0,\;
        g(x)=0
    \right\},
    \label{eq:strict_design_space}
\end{equation}
where $h:\mathcal X_0\to\mathbb R^{n_{\mathrm{ineq}}}$ collects deterministic
inequality constraints and $g:\mathcal X_0\to\mathbb R^{n_{\mathrm{eq}}}$
collects deterministic equality constraints. Such constraints may also arise
inside subproblems used to evaluate a design, for example, dynamics constraints
in trajectory optimization, friction-cone constraints in contact modeling, or
observability constraints in reduced-order sensing.

In addition to strict feasibility, practical co-design problems often include
acceptance requirements that specify minimum deployability levels. These
requirements are not objectives to be optimized along the Pareto front, but
gates that determine whether a feasible design is acceptable for the intended
use case. We denote their margins by
\begin{equation}
    b(x)\leq 0,
    \qquad
    b:\mathcal X\to\mathbb R^{n_b}.
    \label{eq:acceptance_requirements}
\end{equation}
Examples include bounded probability of bruising, minimum task-success rate,
maximum allowable cleaning time, maximum failure probability, certification
readiness, or minimum operational autonomy. Unlike the strict constraints that
define $\mathcal X$, these requirements may depend on uncertain real-world
metrics and may therefore be imposed through conservative estimates or
probabilistic constraints during computational co-design.

Having formalized the core elements of (soft-robot) co-design, we can express common co-design queries. 
For example, under resource limits, the query $\operatorname{FixResMaxFun}$
searches for the maximal achievable functionality subject to resource budgets
and acceptance requirements:
\begin{equation}
\begin{aligned}
    \max_{x}\quad & \operatorname{prov}(x)\\
    \textrm{s.t.}\quad
    & x\in\mathcal X,\\
    & \operatorname{req}(x)\leq r_{\max},\\
    & b(x)\leq 0.
\end{aligned}
\label{eq:FixResMaxFun}
\end{equation}

\emph{Computational co-design} refers to methodologies in which design evaluation and optimization are carried out computationally~\cite{carlone2019robot, wang2023softzoo}. In this setting, the optimization uses an estimated value $\hat{c}_j(x)$--for instance, from simulation--instead of the true design value $c_j(x)$, thereby avoiding the need to prototype and test every candidate design.

\section{Holistic Co-Design of Soft Robots}\label{sec:holistic_co_design}
Holistic co-design reframes soft robot design as the concurrent optimization of morphology, control, and user interaction within an application-driven process, rather than a \emph{sequential} one. This modular framework (Fig.~\ref{fig:overview}) expands the design values considered, improves tractability, explicitly incorporates \emph{realization}, and accounts for uncertainty through targeted prototyping and staged validation.

\subsection{Multi-Objective Co-Design for Comprehensive Design Requirements and Values}
Current co-design approaches for soft robots typically consider only one~\cite{spielberg2019learning, spielberg2021co, medvet2022impact, wang2023preco, wang2023diffusebot, navez2024design}, or at most a few~\cite{navez2025modeling}, evaluation metrics\footnote{Commonly, such evaluation metrics are also referred to as costs, losses, or negative rewards.} per task. However, as soft robots transition from prototypes to commercial products, broader design values become essential, including manufacturing costs and scalability~\cite{miriyev2017soft, schmitt2018soft, majidi2014soft, junge2022leveraging}, material availability, operational lifespan, and ecological impacts~\cite{mazzolai2020vision}.

In particular, manufacturability costs can act as implicit regularizers on shape complexity. By assigning a higher cost to features requiring variable cross-sections or intricate tooling, the optimizer favors uniform cross-sections and symmetric chamber layouts. These designs inherently exhibit fewer strain discontinuities, reducing the effective order of the dynamics (without imposing a constant-curvature assumption) and simplifying strain-based sensing. As a result, both proprioception and feedback control become more tractable.

Additionally, early integration of regulatory compliance checks (e.g., FDA/EMA or ISO-type requirements) can be posed as explicit inequality constraints or feasibility filters, enforced via conservative surrogates and progressively tightened through higher-fidelity simulation and staged prototyping, reducing late-stage redesign while keeping the optimization tractable. More broadly, holistic co-design operationalizes hard-to-quantify values by starting from coarse but computable feasibility checks, optimizing remaining objectives in a multi-objective manner, and refining uncertain metrics through realization (\Cref{sub:probabilistic_co_design_metrics}), conducting late-stage assessments earlier without freezing progress.

Safety--performance is a key trade-off in soft robotics: although soft robots are often described as intrinsically safe, safety remains poorly quantified, and excessive compliance can compromise accuracy, payload, and force output~\cite{abidi2017intrinsic, laschi2014soft, rus2015design, yasa2023overview, stolzle2025phdthesis, pasquier2025study, iida2011soft, cianchetti2013stiff, mazzolai2022roadmap, majidi2014soft, hawkes2017soft}. 
Holistic co-design treats safety as a core design value that is defined by the soft robot's morphology, control~\cite{stolzle2025phdthesis, wong2025contact}, and perception. Our framework could jointly maximize safety and performance while enforcing worst-case safety constraints based on injury severity criteria grounded in biomechanical studies ~\cite[Chapter 3]{stolzle2025phdthesis}; at runtime, safety-aware control can further regulate this trade-off through explicit constraints and CBF-based safety filters~\cite{wong2025contact, dickson2025safe, dickson2026force}.

\begin{figure}[h!]
    \centering
    \ifproduction\else
      \includegraphics[width=1\linewidth]{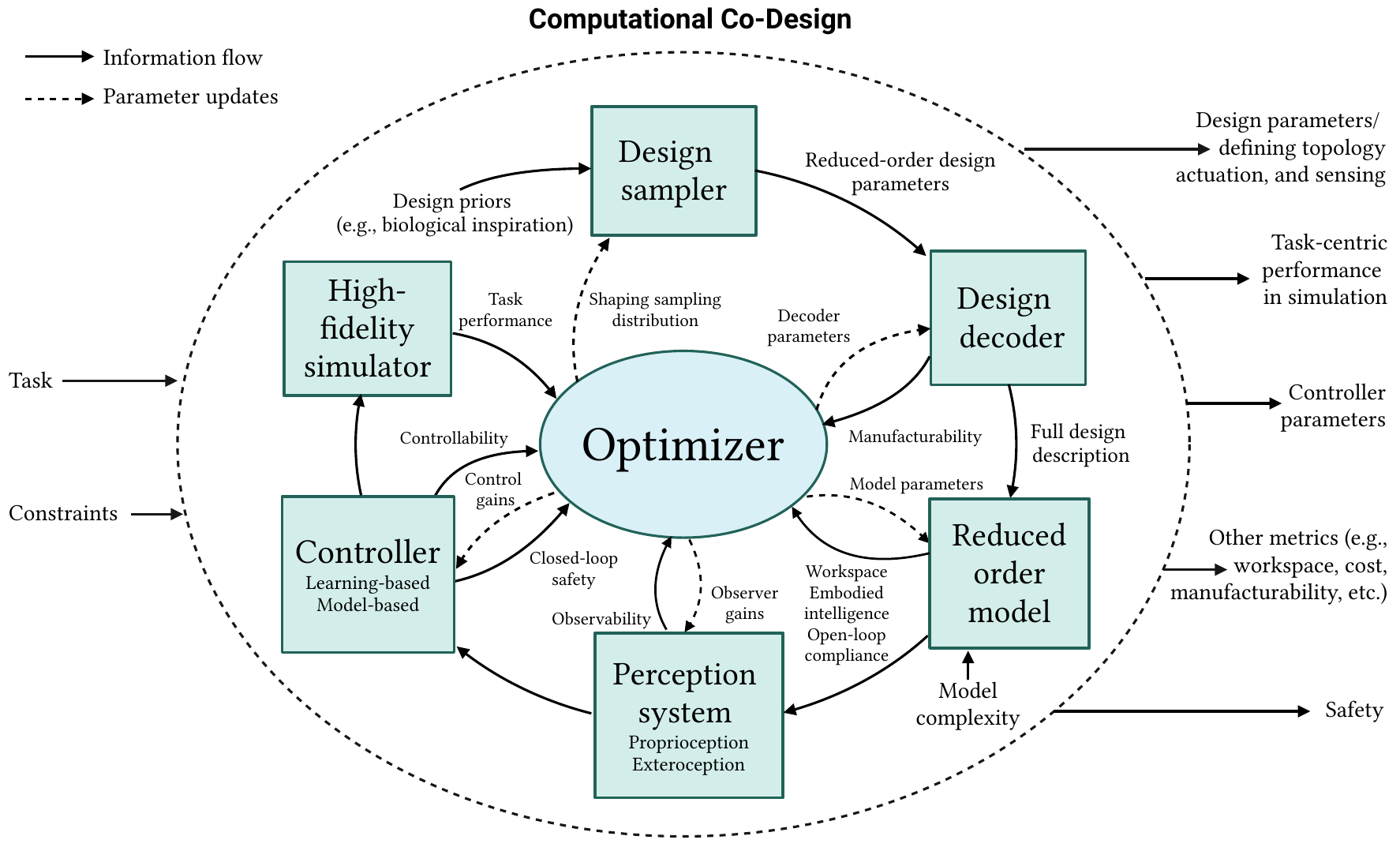}
    \fi
    
    \caption{
        \textbf{A Framework for Efficient Computational Co-Design of Soft Robots.}
        We sample reduced-order design parameters $x$ from an initial distribution that can include design priors from biological inspiration~\cite{mazzolai2020vision, chen2020design, laschi2024bioinspiration}, known mechanisms, or existing soft robot designs. This design space—either learned or explicitly defined by physical or geometric values (i.e., size optimization)—is translated by a design decoder into a detailed robot body description (such as a 3D mesh with sensor and actuator placements) that provides immediate manufacturability feedback. A reduced-order model is then derived~\cite{alora2023data, stolzle2024input, valadas2025learning, alkayas2025soft} or learned to efficiently assess workspace, open-loop compliance, and embodied intelligence~\cite{cianchetti2021embodied, mengaldo2022concise, vihmar2023measure}, offering inexpensive feedback to the optimizer. Similarly, perception and control systems can be developed using this model (via model-based control~\cite{della2023model}, MPC~\cite{alora2023data}, or differentiable physics~\cite{spielberg2019learning, wang2023softzoo, stolzle2026soromox}), while observability and controllability are evaluated without costly simulations. Finally, closed-loop simulations (low-fidelity using the reduced-order model or high-fidelity FEM-based~\cite{coevoet2017software}) assess integrated performance metrics, which in turn optimize the sampling distribution, decoder, and all relevant system parameters.
    }
    \label{fig:computational_co_design}
\end{figure}

\subsection{Efficient Computational Co-Design for Exploring the Entire Design Space}
One reason co-design approaches are not yet widely adopted in soft robotics is their computational inefficiency, which restricts design-space exploration and reduces the chance of finding practically useful solutions~\cite{chen2020design}. Current co-design methods are often too expensive to support broad design-space exploration. A major bottleneck is the use of high-dimensional morphology representations (e.g., voxelized geometries), which makes optimization hard to scale and global search unreliable. Cost is further amplified because designs are commonly assessed via closed-loop simulations, and high-fidelity physics turns each evaluation into a heavy computation~\cite{spielberg2019learning, medvet2021biodiversity, wang2023curriculum, wang2023softzoo, wang2023diffusebot}.
Finally, closed-loop performance cannot be assessed without first deriving a controller, yet current approaches remain limited: shared policies across many morphologies~\cite{zardini2021seeking, boekel2025learning} are more efficient but may fail to represent the best achievable behavior of each design, while deriving a dedicated controller for every candidate is typically too expensive, especially with RL from scratch~\cite{bhatia2021evolution, wang2023curriculum, wang2023softzoo, wang2023preco} or gradient-based optimization in differentiable simulators~\cite{spielberg2019learning, bacher2021design, wang2023softzoo, wang2023diffusebot, visentin2026digital}, and design-conditioned RL~\cite{schaff2023sim} still struggles to handle diverse morphologies while achieving near-optimal performance for each.

Our framework, depicted in Fig.~\ref{fig:computational_co_design}, addresses these challenges through four complementary modifications.

\subsubsection{Optimizing in a Reduced-Order Design Space}

Direct optimization of soft robots requires parameterizing the design via $n_\mathrm{x}$ variables $x \in \mathcal{X}$, with $\mathcal{X} = \mathbb{R}^{n_\mathrm{x}}$\footnote{While a continuous space is assumed here, discrete design choices can be naturally embedded into this framework.}. In soft robot co-design, these parameters encode spatial geometry, actuator~\cite{wang2023diffusebot} and sensor placements~\cite{spielberg2021co, junge2022leveraging}, material choices, and related structural traits. Traditionally, design has relied either on size optimization with explicit geometric/material parameters~\cite{chen2020design, guan2023trimmed, calisti2011octopus, pagliarani2024variable, polygerinos2015modeling, navez2024design, junge2022leveraging} or on discretization-based shape/topology optimization using voxels or particles~\cite{caasenbrood2020computational, pinskier2024diversity, bhatia2021evolution, medvet2022impact, wang2023curriculum, nadizar2022schedule}. The former can be suboptimal or redundant, whereas the latter is computationally expensive and often yields impractical designs.



Recent studies address these issues by optimizing within a reduced-order design space~\cite{howison2020reality} and then using spatial basis functions~\cite{candiello2026unified} or a generative decoder~\cite{song2024morphvae, wang2023diffusebot, navez2024contributions, li2025generating} to reconstruct the complete design (e.g., a 3D mesh with sensor and actuator placements). In our framework, a design sampler in the reduced-order space and a design decoder work in tandem: a probabilistic design policy $\vartheta(x): \mathcal{X} \to [0,1]$ maps parameters to probabilities, reduced-order parameters $x \sim \vartheta(\cdot)$ are sampled during co-design, and then expanded into a full design $d \in \mathcal{D}$ via the decoder $d = \psi(x)$. The decoder can be model-based (e.g., parametric CAD) or implemented using learned generative models such as VAEs~\cite{song2024morphvae, navez2024contributions}, GANs~\cite{hu2022modular}, or Diffusion models~\cite{wang2023diffusebot}. 
Learned decoders can be pretrained offline—either on soft-robot libraries or on general 3D/CAD foundation models—and then rapidly decode reduced-order parameters to full designs, with the one-time training cost amortized across many co-design runs\cite{sanghi2024wala,wu2024direct3d,li2025cadllama}.  Within our framework, such a pretrained decoder $\psi$ can be initialized from existing models and optionally fine-tuned alongside the design sampler to adapt to the specific task.
Design priors—derived from proven soft robot examples or bio-inspired concepts~\cite{mazzolai2020vision, chen2020design, laschi2024bioinspiration}—inform the initial distribution, while the optimizer iteratively refines $\vartheta(x)$~\cite{song2024morphvae, sutton1998reinforcement}.
While priors can initially bias the search, they only initialize $\vartheta(x)$; online updates (with exploration) can shift probability mass beyond the prior toward higher-performing designs.
Additionally, sensitivity analysis can reduce the dimensionality of the sampled parameters and improve the decoder’s conditioning~\cite{chen2020design, guan2023trimmed, maestre2023toros, navez2024contributions}, enabling the optimizer to focus on critical variables by deferring less sensitive parameters or fixing them. 

\subsubsection{Co-Optimizing a Control-Oriented Reduced-Order Model}\label{ssub:rom}

A control-oriented Reduced-Order Model (ROM) is valuable for observers, control, and motion planning~\cite{sadati2018ritz, sadati2023reduced, bruder2021data, armanini2023soft, menager2023direct, alora2023data, stolzle2024input, alkayas2025soft, valadas2025learning, mathew2025reduced}. While deriving such models is relatively straightforward for rigid manipulators~\cite{siciliano2025foundations, zhao2020robogrammar}, soft robots exhibit a complex four-way interplay between morphology (e.g., mass and stiffness distribution, gravitational forces, sensing, actuation), proprioception, control, and tasks (e.g., payload, contact forces) that directly shape their deformations. Namely, the DOFs of the reduced-order model are upper-bounded by the information that can be extracted from the sensors, such that all shape parameters can be estimated online by the proprioception module. This motivates the joint synthesis of morphology, proprioception, the (reduced-order) kinematic model, and the controller.

Traditionally, the soft robotics community has relied on expert intuition and trial-and-error to define finite-dimensional backbone models such as PCC, PCS, and GVS~\cite{webster2010design, renda2018discrete, renda2020geometric, mathew2025reduced, stolzle2026soromox}. In co-design, however, the goal is to automatically identify a suitable control-oriented ROM. Promising approaches include data-driven models~\cite{thuruthel2017learning, bern2020soft, alora2023data, chen2025data, menager2023direct} and methods that adapt existing finite-dimensional strain approximations to a given design~\cite{alkayas2025soft, alkayas2025structure, valadas2025learning, navez2025modeling}. Irrespective of the specific formulation, the behavior of the reduced-order model can be compared against the high-fidelity simulation during optimization, ensuring that the deformation modes most important for the task are captured by the reduced-order model. Integrating reduced-order models into the co-design formulation then ensures that the shape information required by the reduced-order model can be estimated from the available sensor information via the proprioception algorithm, and regularization of (too) high-dimensional reduced-order models as they do not provide added value for the controller given the highly underactuated nature of the system.

\subsubsection{Surrogate Metrics for Rapid Feedback to the Optimizer}

To reduce dependence on expensive closed-loop simulations, we propose using computationally efficient surrogate metrics that provide early feedback during optimization and help discard low-fitness designs. Combining these inexpensive evaluations with selective higher-fidelity assessment draws on ideas from multi-fidelity optimization~\cite{peherstorfer2018survey}. Candidate surrogate metrics include controllability and observability measures derived from reduced-order models, as well as proxies for expected control performance~\cite{sethuramalingam2025design}, embodied intelligence~\cite{cianchetti2021embodied, mengaldo2022concise, vihmar2023measure}, open-loop compliance~\cite{guan2023trimmed}, safety~\cite[Chapter 3]{stolzle2025phdthesis}, and manufacturability.

Controllability and observability can be assessed through nonlinear tools~\cite{griffith1971observability, zheng2019controllability} or by linearizing the dynamics around an equilibrium. Closed-loop stability should also be considered; for instance, PD+Feedforward control is locally stable around $q^\mathrm{d}$ when $\frac{\partial^2 \mathcal{U}}{\partial q^2} + K_\mathrm{p} \succ 0$~\cite{della2023model}. Since low proportional gains are generally preferred~\cite{della2017controlling}, the design should ideally satisfy $\frac{\partial^2 \mathcal{U}}{\partial q^2} \succ 0$ across the desired workspace.

\subsubsection{Deriving the Controller in a Model-Based Fashion}\label{ssub:control_policy_derivation}
The same ROM from \Cref{ssub:rom} can then support model-based controller derivation, including LQR, PID+Feedforward, PD+, MPC, or model-based RL, depending on the model structure~\cite{bruder2021data, della2023model, stolzle2024experimental, stolzle2024input, della2020model, alora2023data, thuruthel2019model}. Compared with training a new RL policy for every design~\cite{bhatia2021evolution, wang2023curriculum, wang2023softzoo, wang2023preco}, these approaches can substantially reduce computational cost.


\begin{figure}[ht]
    \centering
    \ifproduction\else
      \includegraphics[width=1.0\linewidth]{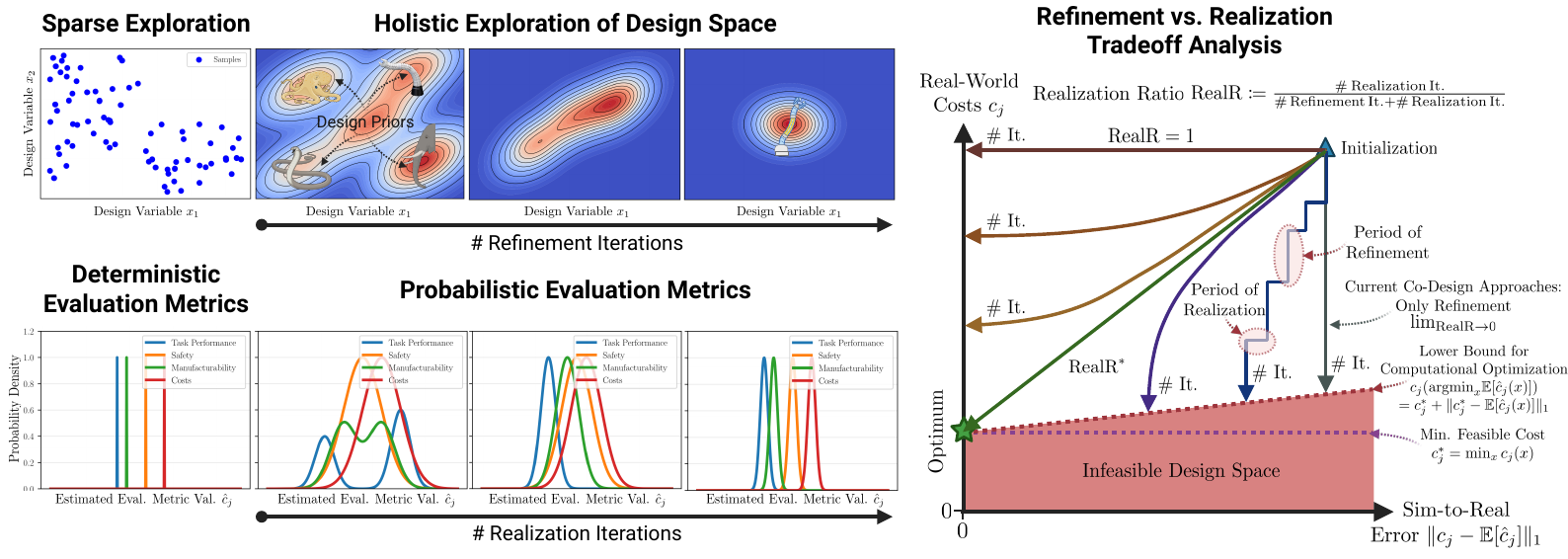}
    \fi
    \caption{\textbf{Refinement vs. Realization.}
    \textit{Top Left:} This panel shows how refinement affects the design sampling distribution $\vartheta(x)$. Unlike (most) conventional co‑design methods that sparsely explore the design space, we condition $\vartheta(x)$ on design priors (e.g., biological inspiration and existing soft robot designs) and iteratively update it via the co‑design optimizer until the optimum is reached.
    \textit{Bottom Left:} Here, we demonstrate how iterative realization refines probabilistic evaluation metrics. Rather than using deterministic metrics that ignore inherent simulation uncertainty, we define a probabilistic distribution $\hat{m}_j(\hat{c}_j, x)$ for each metric, with purposeful prototyping reducing the uncertainty over iterations.
    \textit{Right:} This panel examines the tradeoff between refinement and realization. We define the realization ratio $\mathrm{RealR}$ as the fraction of realization iterations relative to the total iterations and the sim‑to‑real error as $\lVert c_j - \mathbb{E}[\hat{c}_j] \rVert$. We then plot the evolution of real-world cost $c_j$ versus sim‑to‑real error as iterations increase for a fixed $\mathrm{RealR}$: when $\lim{\mathrm{RealR} \to 0}$, designs are optimized computationally with realization occurring only at the end, while $\mathrm{RealR}=1$ means only the confidence in the evaluation metrics is increased. The optimal ratio $\mathrm{RealR}^*$ efficiently minimizes the real-world cost to its optimum $\min_x c_j(x)$ via refinement and reduces the sim‑to‑real error to zero by updating $\hat{m}_j(\hat{c}_j, x)$ via realization.
    }\label{fig:refinement_vs_realization_tradeoff}
\end{figure}

\subsection{Probabilistic Co-Design Metrics for Balancing Refinement and Realization}\label{sub:probabilistic_co_design_metrics}

Holistic co-design evaluates soft robot designs across multiple objectives, many of which cannot be assessed exactly from simulation alone. In practice, simulation-based metrics are often only proxies for true design value and are affected by a well-known reality gap~\cite{mouret2017realitygap, howison2020reality, schaff2023sim, dubied2022sim}. Moreover, important objectives such as manufacturability, cost, ecological sustainability, and operational lifetime often require fabrication and real-world testing. As a result, the design that appears optimal under computational evaluation may differ from the design that is best in practice. Holistic co-design should therefore account explicitly for uncertainty in how design values are estimated.

Connecting to optimization under uncertainty and uncertain control co-design~\cite{azad2023overview}, we represent each evaluation metric probabilistically, as a belief over the true cost or performance of a design conditioned on its specifications. Rather than assigning a fixed value to a metric, we model uncertainty in the estimate, for example through a distribution $p(\hat{c}_j \mid x)$ over the $j$th estimated design cost $\hat{c}_j$ for design $x$. This formulation allows co-design to reason directly about uncertainty during optimization, while also incorporating objectives that are difficult or impossible to evaluate from simulation alone. Importantly, uncertainty arising from simplified physics, model mismatch, sensing and actuation effects, and environmental variability can be absorbed into these probabilistic metric estimates. High-fidelity simulations, prototypes, and experiments then serve not only to validate designs, but also to calibrate and update the metric beliefs used during co-design.

Within this view, we distinguish between \emph{refinement} and \emph{realization}. Refinement refers to computationally improving the design using the current probabilistic metric estimates, for example, by searching for designs that are predicted to perform well under the current belief. Realization, in contrast, refers to allocating resources to more faithful evaluation---such as higher-fidelity simulation, fabrication, or physical experiments---in order to reduce uncertainty and improve the metric estimates themselves. In other words, refinement improves the design under the current model, whereas realization improves the model used to assess the design. As illustrated in Fig.~\ref{fig:holistic_co_design}, refinement is represented by iterative design cycles in the innermost layers (A \& B), whereas realization is shown by selected designs moving to the outer layers (i.e., C-F).

A central challenge in holistic co-design is therefore to balance refinement and realization. Excessive refinement risks over-optimizing against inaccurate metric estimates, whereas excessive realization can consume substantial time, cost, and experimental effort without directly improving the design. This tradeoff is analogous to the exploration--exploitation dilemma in reinforcement learning: refinement exploits the current belief to search for better designs, while realization explores to improve the reliability of that belief. Methods from Bayesian optimization and related areas may provide principled ways to decide when a design should be further refined and when it should instead be realized and tested. Over time, shared datasets and benchmarks for soft robotic designs could further reduce uncertainty in these metric estimates and lessen the dependence on expensive realization.

\subsection{Synergistic Cross-Disciplinary Collaboration}
Collaboration is a core pillar of holistic co-design, engaging engineers, end-users, material scientists, and domain experts from the outset to ensure alignment with real-world requirements. For instance, in robotic harvesting, growers’ knowledge of crop fragility and handling practices can guide the design of soft end-effectors that minimize damage while maximizing yield.

In contrast to traditional sequential workflows, co-design inherently fosters communication across stakeholders and the full design and product development team. It requires the team to jointly define design values and requirements, agree on design variables and priors, and derive metrics, heuristics, and surrogate models for estimating design values. Co-design also makes the selection of trade-offs along the Pareto front an explicit, shared decision (e.g., via preference elicitation or periodic design reviews), rather than handing the soft robot design from one subteam to the next and having each group work and take (subjective) design decisions in isolation.
Importantly, this makes holistic co-design a mixed-initiative process rather than a frozen optimization routine: designers can iteratively update objectives, constraints, priors, and the set of optimized variables as new computational and experimental evidence becomes available.

\begingroup

\section{Operationalizing Holistic Co-Design}

\subsection{From Design Values to Quantifiable Metrics and Surrogates}
To make holistic design values actionable within computational co-design, they must be translated into quantifiable metrics and corresponding surrogates that can be evaluated, compared, and optimized across candidate robot designs. We distinguish between the true metric value $c_j=m_j(x)$, which is obtained from real-world measurement or sufficiently accurate experimental characterization of design $x$, and its computational estimate/surrogate $\hat c_j=\hat m_j(x)$, which is evaluated from design variables, virtual twins, simulated deformation and motion data, material and component data sheets, or limited empirical calibration. The following subsections therefore provide an initial outline of computational metric estimates, with more detailed derivations and assumptions provided in the Supplementary Information~\cite{SI}.

\subsubsection{Safety}
For physical human--robot interaction, safety can be quantified by injury-severity criteria grounded in biomechanical contact limits. Following ISO/TS~15066 and related pHRI literature, relevant limits are typically expressed in terms of maximum contact force or contact pressure for a given body region~\cite{iso2016collaborative,muttray2014collaborative}. The true design-level safety cost $c_{\mathrm{safety}}=m_{\mathrm{safety}}(x)$ would be obtained from systematic collision testing, for instance as the supremum of measured peak contact pressure over admissible contact scenarios. Since such exhaustive testing is infeasible during early co-design, we use a computational estimate $\hat c_{\mathrm{safety}}=\hat m_{\mathrm{safety}}(x)$ based on a projected one-dimensional collision model.

We consider a local blunt-contact scenario with contact geometry $G_\mathrm{c}=(s_\mathrm c,n_\mathrm c)$, where $s_\mathrm c$ denotes the contact location and $n_\mathrm c$ the contact normal. Projecting the robot dynamics onto the contact normal yields a one-dimensional penetration coordinate
\[
\delta_\mathrm c(t)=x_\mathrm R(t)-x_\mathrm H(t),
\qquad
\dot{\delta}_\mathrm c(t)=J_\mathrm c(q,G_\mathrm{c}) \, \dot q(t)+v_\mathrm H .
\]
Under a local linear contact model and short-duration collision approximation, the projected dynamics reduce to
\[
m_\mathrm R\ddot{\delta}_\mathrm c+
(k_\mathrm R+k_\mathrm c)\delta_\mathrm c
=
f_\mathrm u-f_{\mathcal U}^{\mathrm c0},
\]
where $m_\mathrm R$ is the reflected robot inertia, $k_\mathrm R$ the dynamically projected robot stiffness, $k_\mathrm c$ the contact stiffness, $f_\mathrm u$ the projected actuation contribution, and $f_{\mathcal U}^{\mathrm c0}$ the projected elastic/preload force at contact onset. This scalar system admits a closed-form upper bound $\overline f_\mathrm c$ on the peak contact force. We define the scenario-level computational estimate of the Soft Robot Injury Severity Criterion (SRISC) as
\[
\widehat{\mathrm{SRISC}}(\sigma_\mathrm{c})
=
\frac{\overline f_\mathrm c(q,v_\delta,f_\mathrm u,G_\mathrm{c})}
{A_\mathrm c^{\min}(G_\mathrm{c})},
\qquad
\sigma_\mathrm{c}=(q,\dot q,G_\mathrm{c},u),
\]
and lift it to the estimate of the Soft Robot Design Hazardousness Criterion (SRDHC)
\[
\hat c_{\mathrm{safety}}(x)
=
\widehat{\mathrm{SRDHC}}(x)
=
\sup_{\sigma_\mathrm{c}\in\Sigma_\mathrm{c}(x)}
\widehat{\mathrm{SRISC}}(\sigma_\mathrm{c}),
\]
where $\Sigma_\mathrm{c}(x)$ denotes the feasible collision scenarios of design $x$, including admissible states, contact locations, velocities, and inputs. 
This conservative surrogate directly connects to the safety certification criteria in ISO/TS 15066~\cite{iso2016collaborative} for maximum contact forces and pressures, which are grounded in biomechanical studies~\cite{muttray2014collaborative}, and supports early-stage comparisons of morphologies, materials, actuation strategies, and controllers.
It does not replace physical collision testing or safety validation, but enables safety to enter the co-design loop before hardware realization.

\subsubsection{Manufacturability}
Manufacturability should enter co-design as an expected fabrication cost rather than a qualitative score. The true manufacturability cost $c_{\mathrm{mfg}}=m_{\mathrm{mfg}}(x)$ is the realized cost per accepted robot, measured from fabrication batches, including actual material use, labor, equipment occupancy, energy, and yield losses. During computational co-design, we estimate this quantity for a design $x$, and batch size $B$ as
\[
\hat c_{\mathrm{mfg}}(x,\phi;B)
=
\frac{\hat C^{\mathrm{tool}}(x,\phi)}{B}
+
\sum_{o\in\mathcal O_\phi}
\hat\chi_o(x,\phi)\hat C_o(x,\phi),
\]
where $\hat C^{\mathrm{tool}}/B$ amortizes molds, fixtures, and jigs, and $\mathcal O_\phi$ is the ordered set of fabrication, assembly, and inspection operations. 
Depending on the soft robot design class, the process route $\phi$ may either be fixed a priori, such as silicone casting or 3D printing, or selected from a set of feasible routes, with each candidate design evaluated using the route associated with its specific morphology, materials, and fabrication constraints.
The estimated nominal operation cost is
\[
\hat C_o
=
\hat C_o^{\mathrm{mat}}+
\hat C_o^{\mathrm{buy}}+
a_o^{\mathrm{lab}}\hat t_o^{\mathrm{lab}}+
a_o^{\mathrm{equip}}\hat t_o^{\mathrm{equip}}+
c_o^{\mathrm{energy}}\hat E_o,
\]
in line with standard manufacturing-cost and design-for-manufacture-and-assembly models~\cite{boothroyd2010product,ostwald2004cost,niazi2006cost}. The estimated yield-loss multiplier
\[
\hat\chi_o(x,\phi)
=
\left(\prod_{r\succeq o}\hat y_r(x,\phi)\right)^{-1}
\]
penalizes designs for which operation $o$ or downstream operations have low pass probability. For pneumatic soft robots, the estimator connects directly to computational design parameters such as material volume, mold volume, support volume, curing time, number of chambers, tubing length, minimum wall thickness $t_{\mathrm{wall,min}}$, minimum draft angle $\theta_{\mathrm{draft,min}}$, core slenderness, bond width, and proof-pressure margin.

\subsubsection{Durability}
Durability captures whether a soft robot preserves repeatable behavior over prolonged operation, rather than merely whether it survives a single task. The true durability metric $c_{\mathrm{dur}}=m_{\mathrm{dur}}(x)$ should ultimately be obtained from repeated-use or accelerated-life experiments that quantify degradation, fatigue, hysteresis, recovery, or failure over time~\cite{stella2024toward,ouyoucef2025durability}. As an actionable computational proxy, we use the trajectory-induced strain-energy demand. For simulated or model-predicted strain coordinates $\hat q(t)$, zero-strain configuration $\bar q$, stiffness matrix $K$, and task horizon $H$, define
\[
\hat E_K(t)=
\frac{1}{2} \, (\hat q(t)-\bar q)^\top K(\hat q(t)-\bar q),
\qquad
\hat c_\mathrm{dur}(x;H)
=
\frac{1}{H}\int_0^H \hat E_K(t)\,dt .
\]
This estimate penalizes designs and controllers that repeatedly drive the soft body through energetically demanding deformations. When stress concentrations dominate failure, local or maximum strain-energy variants can be used instead. The key point is that durability is controller- and task-dependent: the same morphology may be benign under one policy but fatigue-inducing under another. When comparing designs with different sizes or material volumes, normalized or local strain-energy-density variants may be more appropriate.

\subsubsection{Sustainability}
Sustainability is a multidimensional concept, commonly understood to encompass environmental, social, and economic dimensions~\cite{purvis2019three}. Here, we focus on the environmental dimension through carbon footprint, expressed in $\mathrm{kg\,CO_2e}$, because it can be assembled from physical quantities and life-cycle inventories~\cite{iso14067,ghgprotocol2011product}. The true sustainability cost $c_{\mathrm{sust}}=m_{\mathrm{sust}}(x)$ corresponds to the measured or fully accounted life-cycle carbon footprint per functional unit. During co-design, we estimate it as
\begin{equation}\small
\hat c_{\mathrm{sust}}(x;H)
=
\frac{
\sum_{j\in M_\mathrm{rep}(x)} \hat n_j(x;H) \hat C_j^{\mathrm{mod}}(x)
+
\hat C^{\mathrm{ops}}(x;H)
}
{\hat Y(x;H)},
\end{equation}\small
where $\hat Y(x;H)$ is the expected useful output of the robot over the mission horizon $H$ and
\[
\hat C_j^{\mathrm{mod}}(x)
=
\hat C_j^{\mathrm{make}}(x)+\hat C_j^{\mathrm{eol}}(x),
\quad
\hat C^{\mathrm{ops}}(x;H)
=
\hat C^{\mathrm{use}}(x;H)
+
\hat C^{\mathrm{repair}}(x;H)
+
\hat C^{\mathrm{trans}}(x;H).
\]
Here, $M_\mathrm{rep}(x)$ are replaceable modules, $\hat n_j$ is the expected number of manufactured copies, $\hat C_j^{\mathrm{make}}$ and $\hat C_j^{\mathrm{eol}}$ are estimated manufacturing and end-of-life footprints, and the remaining terms capture estimated use, repair, and transport footprints.

Several terms are especially relevant for soft robots. For tendon-driven actuation, the use-phase energy consumption can be estimated from tendon tensions $u$, tendon velocities $v_{\mathrm{ten}}=A_\mathrm{u}^\top(q) \, \dot q$, spool radius $r_k$, gear ratio $g_k$, and motor-driver power maps $\Pi_k$:
\begin{equation}
\hat E_{\mathrm{elec}}^{\mathrm{ten}}(x;H)
=
\sum_{\ell=1}^{N_H}
\int_0^{T_\ell(x)}
\sum_{k=1}^m
\Pi_k\!\left(
\frac{r_k u_k}{g_k\eta_{\mathrm{gb},k}},
\,
g_k\frac{v_{\mathrm{ten},k}}{r_k}
\right)
dt .
\end{equation}
Repairable or self-healing materials reduce replacement by increasing the probability that damage is repaired locally rather than through module replacement~\cite{terryn2017selfhealing,giordano2023sustainable}. Degradable or on-demand degradable materials can reduce end-of-life burdens by changing disposal, separation, or recovery pathways~\cite{oh2023photodegradable}. Thus, $\hat c_{\mathrm{sust}}$ exposes design levers such as material choice, actuation efficiency, lifetime, repairability, modularity, and end-of-life separability within one carbon-grounded computational estimate.

\subsection{Case Study of Holistic Co-Design for Indoor Tomato Harvesting}
\label{sub:case_study_tomato}

We include this case study to make the preceding perspective actionable rather than purely conceptual. Indoor tomato harvesting is a useful stress test for holistic co-design because the robot must coordinate morphology, perception, control, materials, hygiene, economics, and deployment logistics while interacting with fragile and variable biological objects. The goal is not to prescribe a single optimal harvester, but to show how the proposed framework can be instantiated as a stepwise recipe: stakeholders and values are translated into metrics, strict constraints and tradeable objectives are separated, uncertainty is represented explicitly, and refinement is coupled to selective realization before a deployment-ready system is chosen.

\paragraph{Step 1 Define the problem setting and stakeholders}
The first step is to identify the operating environment, the relevant stakeholders, and the deployment constraints. In greenhouse tomato harvesting, fruits grow in clusters at variable heights, are often occluded by leaves and stems, and are surrounded by delicate structures such as support strings, branches, and neighboring fruits. Commercial greenhouses also impose narrow aisle geometries, high humidity, cleaning requirements, and strict limits on downtime and maintenance. Relevant stakeholders include growers, operators, agronomists, system integrators, safety assessors, quality-control personnel, and fruit buyers. Each stakeholder contributes different requirements: growers prioritize throughput and cost per harvested unit; operators emphasize ease of use, cleaning, and maintenance; agronomists provide knowledge of cultivar variability, peduncle strength, ripening patterns, and plant damage risks; and safety or quality assessors define constraints related to hygiene, food-contact materials, and safe operation around workers and crops.

\paragraph{Step 2 Translate needs into design values and metrics}
Stakeholder needs must then be converted into explicit design values and corresponding metrics. For tomato harvesting, relevant design values include fruit integrity, harvest throughput, robustness, hygiene, maintainability, energy efficiency, and economic viability. These values can be translated into metrics such as pick success rate, cycle time, bruise probability after storage, dropped-fruit rate, plant disturbance, cleaning time per shift, energy per pick, system availability, and cost per harvested unit. This translation is essential because it defines what the co-design process should optimize, constrain, or monitor before specifying which parts of the robotic system are allowed to vary.

\paragraph{Step 3 Define computational surrogates and probabilistic metrics}
After defining the relevant design values and measurable requirements, the next step is to specify how each metric will be estimated during co-design. Some quantities can be approximated through computational surrogates that provide rapid feedback to the optimizer. For example, reachability, estimated cycle time, energy per pick, perception confidence, collision risk, contact-pressure proxies, and control-related metrics can be evaluated using reduced-order models, simplified contact models, domain-randomized simulation, or data-driven surrogate models. These surrogates do not need to be perfectly accurate at early stages; rather, they provide inexpensive estimates that allow the design process to compare candidate solutions and discard clearly unsuitable ones. Other quantities are much harder to model or simulate directly and should therefore be represented as explicitly probabilistic metrics. In tomato harvesting, bruise probability depends on contact pressure, fruit maturity, storage conditions, and cultivar-specific tissue properties; and cleaning performance, plant disturbance, or long-term reliability depend on deployment conditions that are difficult to capture in simulation. Instead of assigning deterministic values to these quantities, holistic co-design treats them as uncertain estimates that are progressively updated through realization. Physical experiments, such as artificial-peduncle tests, instrumented fruit trials, washdown tests, and pilot greenhouse trials, then reduce uncertainty in the most deployment-critical metrics.

\paragraph{Step 4 Distinguish strict constraints and acceptance requirements from resources and functionalities}
A practical co-design formulation should distinguish between constraints that cannot be violated, minimum requirements for deployment, and objectives that can be traded against one another. Strict feasibility constraints act as filters: designs that cannot fit within greenhouse aisles, cannot be cleaned, cannot operate in humid environments, or do not use food-contact-compatible materials are discarded rather than traded against performance. Acceptance requirements define minimum deployability requirements, such as a target probability of successful picking, bounded bruising, limited plant damage, acceptable cleaning time, and sufficient autonomy during a harvest window. Within this feasible and acceptable set, the optimizer can explore trade-offs among throughput, energy per pick, fruit quality, robustness, and economic viability. In this sense, harvesting throughput represents a functionality provided by the system, whereas cost, energy, maintenance time, cleaning labor, and consumables represent resources required by the system.

\paragraph{Step 5 Define the design space}
Once the design values, measurable requirements, probabilistic metrics, and constraint classes have been defined, the next step is to specify the design variables through which those requirements can be addressed. In the tomato-harvesting platform, the design space may include base footprint and bin-handling layout, manipulator length, degrees of freedom and compliance distribution, end-effector geometry, number of contact elements, contact-material stiffness and friction, actuator placement, sensing modalities and sensor locations, perception parameters, contact-control gains and force thresholds, and sealing or cleanability features. 

The design space is therefore the set of physical, sensing, control, and operational variables that the optimizer or design team may modify. Deployment requirements, such as aisle width, food-contact compatibility, humidity tolerance, maximum allowable bruising, or minimum pick success, are not themselves design parameters; rather, they bound the feasible region of the design space $\mathcal{X}$ or become constraints in the previous step. Importantly, not all variables need to remain free throughout the entire process. At early technology-readiness levels, the search may focus on variables related to gentle detachment and fruit retention; at later stages, validated modules can be frozen while the optimizer refines end-effector geometry, compliance placement, sensing configuration, and control thresholds.

\paragraph{Step 6 Select the optimization strategy}
The optimization strategy should be chosen after the design space, metric estimators, uncertainty model, and constraint classes have been specified, since different subproblems have different mathematical structures. Design priors can guide the search as initial sampling distributions, warm starts, restricted decoder families, candidate motion primitives, or architecture proposals. In tomato harvesting, such priors may include biologically inspired grasping strategies, torsional detachment, and proven gripper morphologies. These priors improve sample efficiency, but the optimizer should still be able to compare them against alternatives generated by the broader design space.

Smooth and differentiable subproblems, such as the refinement of control gains, force thresholds, compliance distribution within a fixed topology, or trajectory parameters, can be addressed with gradient-based methods. By contrast, discrete architecture choices, non-smooth contact events, mixed material selections, and expensive physical evaluations are often better handled with derivative-free methods, including Bayesian optimization, evolutionary or genetic algorithms, and multi-objective acquisition functions~\cite{jones1998efficient,hernandez2014predictive,garnett2023bayesian,chen2020design}. A practical implementation may therefore use a multi-hierarchy optimizer~\cite{antonova2023rethinking}: a Bayesian layer allocates refinement or realization queries under metric uncertainty, an evolutionary layer explores gripper and sensor-layout families and local gradient-based solvers refine continuous controller and morphology parameters once a promising architecture has been identified. The optimizer should also exploit the composability of the co-design problem. For example, perception, arm reachability, end-effector mechanics, detachment primitives, cleaning constraints, and bin-handling logistics can first be modeled as coupled subsystem design problems and then composed into a system-level trade-off relation. When these subsystems can be expressed as monotone design problems with implementation variables (MDPIs), the formulation can propagate functionality--resource relations, eliminate dominated implementations, and preserve Pareto-efficient trade-offs across subsystem boundaries~\cite{zardini2023co,zardini2021co,zardini2022task}. This compositional approach is particularly useful when validated modules can be fixed while others remain open to refinement.

\paragraph{Step 7 Refine computationally then realize selectively}
The refinement--realization process should proceed in stages. Early in the process, low-cost computational models and physical twins, such as artificial peduncles with calibrated detachment forces, provide rapid feedback on candidate end-effectors and detachment primitives. Instrumented fruit trials then measure contact forces, slip events, dropped-fruit rates, and visible or delayed bruising after storage. Finally, pilot greenhouse trials update the metric distributions under realistic lighting, humidity, occlusion, plant motion, and operator-maintenance conditions. This staged process avoids two common failure modes: over-optimizing against simulation-only metrics that may not predict real fruit damage, and fabricating every candidate design, which would be prohibitively expensive. Instead, realization is allocated to designs or metrics for which uncertainty is high and deployment relevance is critical. For instance, if two end-effector geometries achieve similar simulated pick success but the uncertainty in bruise risk or cleaning performance is high, targeted fruit trials and washdown tests can be used before selecting a point on the Pareto front.

\paragraph{Step 8 Select a deployment-ready system through auditable design choices}
The final design is not necessarily the best-performing gripper under a single metric, but the system that offers the best compromise within the feasible and acceptable design set. In tomato harvesting, this means choosing a platform that can harvest reliably while limiting bruising and plant damage, remaining cleanable and maintainable, and keeping the cost per harvested unit within an acceptable range. A prototype may perform well in simplified trials but still be unsuitable if it is difficult to clean, damages plants under occlusion, requires frequent maintenance, or uses materials that degrade in humid and chemically cleaned environments. In such cases, a slightly lower-performing design may be preferable if it is more robust, easier to service, and more compatible with hygiene and safety requirements. To make this choice auditable, the design team should preserve the rationale for strict-constraint filtering, acceptance-threshold selection, uncertainty updates from realization, and final Pareto-front selection. By treating constraints, resources, objectives, and uncertain metrics explicitly, holistic co-design reduces the risk that practical limitations emerge only at the end of development.

\endgroup

\section{Conclusions and Outlook}

\subsection{Conclusion}

High-performing soft robots require moving beyond simulation-centered design and co-design approaches that often overlook realization, safety, and regulatory constraints. To address these gaps, we propose a holistic co-design framework that optimizes broader values, boosts computational efficiency to enable tractable exploration, and increases confidence in computational metrics through targeted realization.

Holistic co-design spans a spectrum: when instantiated with broad priors, learned/generative decoders, and exploration-heavy optimization, it can enable the discovery of uncommon morphologies; when instantiated with (existing) design priors, parametric decoders, and efficient mechanics-aware models/controllers, it more directly supports reliable real-world adoption by optimizing performance \emph{together with} realizability and reliability values. In this perspective, our primary contribution is the latter methodological blueprint for systematic adoption, while remaining compatible with the former exploratory setting.


\subsection{Limitations}

This article is a \emph{perspective} that outlines a holistic co-design blueprint, not a technical paper with a complete implementation. Practical pipelines will therefore remain robot- and modality-specific, depending on the sensing/actuation stack and manufacturing route.

A key limitation is the absence of \emph{fully derived and validated} metrics that are both faithful and computationally tractable. Developing such metrics will require joint efforts across domain experts and computational optimization researchers. Moreover, several adoption-critical values (e.g., sustainability) are context-dependent and will initially rely on conservative proxies whose usefulness depends on credible test protocols and data.

Although many ingredients already exist in isolation, their \emph{coupled} optimization (e.g., morphology--perception--reduced-order model--controller co-optimization) remains an open challenge. Closing the sim-to-real gap and reducing metric uncertainty through prototyping can be resource-intensive~\cite{howison2020reality}, motivating further work in automated fabrication and scalable manufacturing workflows~\cite{vogt2018shipboard, gepner2025flexprinter, aygul2025soft}. Likewise, formalizing the refinement--realization trade-off and defining useful \emph{design priors} remain open problems.

Finally, optimization remains a bottleneck because of the dimensionality, coupling, and mixed discrete/continuous structure of holistic co-design. Non-convexity is challenging \emph{for both} sampling-based and gradient-based methods, making hybrid strategies promising~\cite{antonova2023rethinking}. Two key directions to improve tractability are: (i) exploiting the \emph{compositionality} of co-design to decompose the search into modular subproblems~\cite{zardini2023co}, and (ii) \emph{convexifying} parts of the pipeline, or using convex surrogates/relaxations, where possible. Together with shared benchmarks, standardized protocols, and reusable software, these directions could reduce the marginal cost of holistic co-design over time.

\subsection{Next Steps and Open Challenges}

Key priorities are:
(1) deriving computationally evaluable metrics for manufacturability, fabrication and operation costs, durability, and sustainability;
(2) incorporating design priors from existing soft robotic designs~\cite{navez2024design} or biological inspirations~\cite{laschi2012soft, krieg2015design, chen2020design, laschi2024bioinspiration}, including through generative models that propose task-specific morphologies~\cite{grattafiori2024llama, hurst2024gpt, stella2023can, wang2023diffusebot, ghasemi2025vision};
(3) using VLMs as design critics~\cite{ghasemi2025vision} to assess ergonomics, versatility, and perceived safety/friendliness;
(4) optimizing the refinement--realization trade-off by accounting for resource requirements;
(5) addressing control-policy optimization through X-Embodiment policies~\cite{o2024open} that generalize across multiple soft robotic designs without retraining; and
(6) characterizing the Pareto front between safety and performance.

We emphasize that the modular nature of the proposed framework lets the community build on concepts and technical approaches scattered across otherwise disjoint literatures, and develop the required components step by step, rather than implementing a monolithic algorithm all at once.

Finally, we invite the community to perform a controlled, initially restricted benchmark study---as a pragmatic first step toward broader evaluations---that compares a (holistic) co-design approach against a traditional design process in terms of total product-development cost (including human labor, computation, and prototyping) and achieved Pareto performance.

\bibliography{main}

\section*{Acknowledgements}
We thank Yujun Huang and Marius Furter from the Zardini Lab at MIT and Andrew Fletcher from UC San Diego for reviewing the manuscript draft and providing feedback.

\section*{Funding}
The work by Maximilian Stölzle was supported under the European Union's Horizon Europe Program from Project EMERGE - Grant Agreement No. 101070918, and by the Cultuurfonds Wetenschapsbeurzen 2024 and the Rudge (1948) and Nancy Allen Chair for his research visit to LIDS/Zardini Lab at MIT. The work by Cecilia Laschi was supported by the National University of Singapore through the start-up grant RoboLife and by the Singapore-Italy collaborative project DESTRO (Grant R23I0IR043).

\section*{Author contributions}
M.S. and N.P. contributed equally to this work. M.S., N.P. and F.S. conceived the perspective and developed the holistic co-design framework. M.S. led the literature review. M.S. and N.P. wrote the manuscript and revised it in response to co-author and reviewer feedback. M.S. and N.P. prepared Figures~1 and~3; F.S. and M.S. prepared Figure~2; M.S. prepared Figure~4; and F.S. and N.P. prepared Supplementary Figure~1. M.S. prepared Supplementary Note~1 on probabilistic co-design metrics and the refinement--realization tradeoff and developed the safety metric in Supplementary Note~2. M.S. and N.P. jointly developed all other metrics in Supplementary Note~2, the case study in Supplementary Note~3 and the co-design formulation demonstration in Supplementary Note~4. C.L., D.R., M.C., C.D.S. and G.Z. provided feedback on the manuscript. G.Z., C.D.S. and M.C. supervised the work. C.D.S., G.Z., M.C. and D.R. provided funding for the work. All authors reviewed and approved the final manuscript.

\section*{Competing interests}
Francesco Stella, Josie Hughes and Cosimo Della Santina hold equity in Embodied AI. Niccolò Pagliarani and Matteo Cianchetti are involved in technology transfer through the emerging startup Capio Robotics. The remaining authors declare no competing interests.

\ifincludesupplement
  \clearpage
  \newgeometry{textwidth=6.5in,textheight=9in,centering,bindingoffset=0mm}
  \section*{Supplementary Information}
  \setcounter{section}{0}
  \setcounter{subsection}{0}
  \setcounter{figure}{0}
  \setcounter{table}{0}
  \setcounter{equation}{0}
  \renewcommand{\theequation}{S\arabic{equation}}
  \renewcommand{\theHsection}{supplement.\arabic{section}}
  \renewcommand{\theHfigure}{supplement.\arabic{figure}}
  \renewcommand{\theHtable}{supplement.\arabic{table}}
  \renewcommand{\theHequation}{supplement.\arabic{equation}}
  \renewcommand{\figurename}{Supplementary Fig.}
  \renewcommand{\tablename}{Supplementary Table}
  \crefname{figure}{Supplementary Figure}{Supplementary Figures}
  \Crefname{figure}{Supplementary Figure}{Supplementary Figures}
  \crefname{table}{Supplementary Table}{Supplementary Tables}
  \Crefname{table}{Supplementary Table}{Supplementary Tables}
\newcommand{\suppnote}[1]{%
  \refstepcounter{section}%
  \section*{Supplementary Note \thesection: #1}%
  \addcontentsline{toc}{section}{Supplementary Note \thesection: #1}%
}

\crefname{section}{Supplementary Note}{Supplementary Notes}
\Crefname{section}{Supplementary Note}{Supplementary Notes}
\crefname{subsection}{Supplementary Section}{Supplementary Sections}
\Crefname{subsection}{Supplementary Section}{Supplementary Sections}
\crefname{subsubsection}{Supplementary Section}{Supplementary Sections}
\Crefname{subsubsection}{Supplementary Section}{Supplementary Sections}

\suppnote{Probabilistic Co-Design Metrics and the Refinement--Realization Tradeoff}\label{si:note:probabilistic_codesign}

This note expands the probabilistic view of holistic co-design introduced in the main text. The central motivation is that many design values of a soft robot cannot be known exactly from computation alone. Closed-loop simulations, reduced-order models, learned surrogates, and analytical estimates are indispensable for rapid search, but they are still imperfect proxies for real-world behavior. This is especially important for metrics affected by the reality gap---for example, task success, contact safety, durability, manufacturability, operational lifetime, and cost~\cite{mouret2017realitygap,howison2020reality,schaff2023sim,dubied2022sim}. A design that is optimal under a deterministic simulation metric may therefore not be optimal after fabrication, field testing, or stakeholder evaluation. Holistic co-design should consequently optimize not only over predicted metric values, but also over the uncertainty attached to those predictions.

\subsection{True metrics, surrogate metrics, and metric beliefs}
Let $x\in\mathcal X$ denote the design variables considered by the co-design optimizer. These variables may directly parameterize a robot, or they may be reduced-order variables that are decoded into a full design $d=\psi(x)$. We associate each design value with an evaluation metric
\begin{equation}
    c_j = m_j(x) \in \mathcal C_j, \qquad j=1,\ldots,J,
\end{equation}
where $c_j$ is written as a cost to be minimized. Functionality metrics that are naturally maximized can be converted to costs by a sign change or by introducing a task-specific loss. The true metric $m_j(x)$ is the value that would be obtained from sufficiently faithful real-world evaluation of the design, such as fabrication, benchtop testing, field trials, user studies, certification tests, or operational measurements at the relevant Technology Readiness Level (TRL)~\cite{NASA_TRL}. High-fidelity simulations may provide more accurate surrogates, but they do not replace physical realization.

In practice, obtaining $m_j(x)$ exactly is often infeasible during early design. Instead, the optimizer has access to a computational estimate or surrogate. We write the uncertain metric estimate at co-design iteration $\iota$ as a probability distribution
\begin{equation}
    \widehat m_{j,\iota}(\cdot,x)
    :=
    p_{\iota}(\hat c_j\mid x,\mathcal D_{\iota}).
    \label{eq:si_metric_belief}
\end{equation}
where $\hat c_j$ is a possible value of the $j$th metric and $\mathcal D_{\iota}$ denotes all computational and realized evidence collected so far. This evidence may include high-fidelity simulation results, measurements on partial prototypes, material tests, calibration data, expert annotations, field-trial results, and historical benchmark data from related designs. The mean of \Cref{eq:si_metric_belief} can be used as a point estimate, while its variance or credible interval quantifies uncertainty due to limited evidence, rollout stochasticity, and model mismatch.

Metric uncertainty is often inherited from the prediction pipeline used to evaluate a design. Let $z\in\mathcal Z$ denote an intermediate model prediction, such as a simulated rollout, reduced-order-model (ROM) state trajectory, deformation field, contact history, material response, or learned dynamics prediction. Let $\theta\in\Theta$ denote uncertain simulator, ROM, or learned-model parameters, including discrepancy terms that represent the reality gap. The predictive distribution over such intermediate quantities can be written as
\begin{equation}
    p_{\iota}(z\mid x,\mathcal D_{\iota})
    =
    \int p(z\mid x,\theta)\,p_{\iota}(\theta\mid\mathcal D_{\iota})\,\mathrm{d}\theta .
    \label{eq:si_rollout_belief}
\end{equation}
A metric is then obtained by applying a metric readout to this uncertain prediction. If $\eta_j$ denotes metric-specific nuisance variables, such as measurement noise, threshold uncertainty, environmental variation, manufacturing variability, or operator behavior, one convenient factorization is
\begin{equation}
    p_{\iota}(\hat c_j\mid x,\mathcal D_{\iota})
    =
    \int
        p(\hat c_j\mid z,x,\eta_j)\,
        p_{\iota}(z\mid x,\mathcal D_{\iota})\,
        p_{\iota}(\eta_j\mid\mathcal D_{\iota})
        \,\mathrm{d}z\,\mathrm{d}\eta_j .
    \label{eq:si_composed_metric_belief}
\end{equation}
More generally, the product $p_{\iota}(z\mid x,\mathcal D_{\iota})p_{\iota}(\eta_j\mid\mathcal D_{\iota})$ can be replaced by a joint belief if model and metric uncertainties are correlated. The important point is that the metric belief is a push-forward distribution: uncertainty in the simulation, ROM, learned surrogate, environmental assumptions, and metric readout is propagated through the metric definition. This subsection defines the current belief at iteration $\iota$; the realization-based update of that belief is introduced in the realization subsection below. If the metric is deterministic once a rollout is known, then $p(\hat c_j\mid z,x,\eta_j)$ reduces to a Dirac distribution centered at a metric function $g_j(z,x,\eta_j)$, but the resulting metric belief can still be broad because the predicted rollout or model parameters are uncertain.


A useful diagnostic is the metric-specific sim-to-real error
\begin{equation}\label{eq:si_sim_to_real_error}
    e_{j,\iota}(x)
    =
    \left|m_j(x)-\mathbb E_{\iota}[\hat c_j\mid x]\right|.
\end{equation}
This error can be evaluated only for realized designs but can be modeled across the design space. Across metrics, the collection $\{e_{j,\iota}(x)\}_{j=1}^J$ indicates which metric beliefs are still poorly calibrated. 

\subsection{Refinement: improving the design under the current belief}
Refinement denotes computational improvement of the design while holding the current metric beliefs fixed, except for inexpensive surrogate queries. At iteration $\iota$, a design sampler $\vartheta_{\iota}(x)$ proposes candidate designs, possibly biased by design priors such as biological inspiration, proven soft-robot morphologies, or previous successful prototypes. 

Uncertainty can enter the co-design optimization in several ways. A risk-neutral formulation may optimize the posterior mean $\widetilde c_{j,\iota}(x) = \mathbb E_{\iota}[\hat c_j\mid x]$. A risk-aware formulation may optimize an upper confidence bound
\begin{equation}
    \widetilde c_{j,\iota}(x)
    =
    \mathbb E_{\iota}[\hat c_j\mid x]
    + \beta_j\sqrt{\mathrm{Var}_{\iota}[\hat c_j\mid x]},
\end{equation}
or impose chance constraints such as
\begin{equation}
    \Pr_{\iota}\!\left[\hat c_j(x)\leq \tau_j\right] \geq 1-\delta_j,
    \label{eq:si_chance_constraint}
\end{equation}
where $\tau_j$ is an acceptance threshold and $\delta_j$ is an allowable probability of violation. For metrics that are naturally maximized, such as pick success, the inequality direction is reversed, or the metric is first converted into a cost. Such chance constraints are particularly useful for deployment-critical requirements, for example, maximum contact pressure, maximum bruise probability, minimum pick success, bounded cleaning time, or minimum operational uptime.
From now on, let $\widetilde c_{j,\iota}(x)$ denote the chosen risk summary of the $j$th metric.

In practice, for a candidate design $x$, each probabilistic metric can, for example, be queried by Monte Carlo sampling,
\begin{equation}
    \hat c_j^{(k)} \sim \widehat m_{j,\iota}(\cdot,x),
    \qquad k=1,\ldots,K,
\end{equation}
which makes it tractable to obtain samples or risk summaries. 

To emphasize that refinement is multi-objective and constrained, we separate
the probabilistic metric beliefs into functionality estimates $\widehat f_{\iota}$,
resource estimates $\widehat r_{\iota}$, and acceptance-margin estimates
$\widehat b_{\iota}$. After applying suitable risk summaries, e.g.,
lower-confidence bounds for functionalities, upper-confidence bounds for
resources, and conservative upper-confidence bounds for acceptance margins, the
current belief induces
\[
    \widetilde f_{\iota}(x),\qquad
    \widetilde r_{\iota}(x),\qquad
    \widetilde b_{\iota}(x).
\]
Strict feasibility is enforced by the design set $x\in\mathcal X$, where
$\mathcal X$ may itself be defined through deterministic equality and inequality
constraints $g(x)=0$ and $h(x)\leq0$. Belief-dependent acceptance requirements
then define the current admissible set
\begin{equation}
    \mathcal X_{\iota}^{\mathrm{adm}}
    =
    \left\{
        x\in\mathcal X
        \;\middle|\;
        \widetilde b_{\iota}(x)\leq 0
    \right\}.
    \label{eq:si_belief_dependent_admissible_set}
\end{equation}
A refinement step for updating the design sampler and the decoder parameters can therefore be written compactly as
\begin{equation}
    (
    \vartheta_{\iota+1},
    \psi_{\iota+1}
    )
    =
    \operatorname{Refine}_{\mathcal Q}
    \!\left(
        \vartheta_{\iota},
        \psi_{\iota},
        \widetilde f_{\iota},
        \widetilde r_{\iota},
        \mathcal X_{\iota}^{\mathrm{adm}}
    \right),
    \label{eq:si_refinement_update}
\end{equation}
where $\mathcal Q$ denotes the co-design query, such as maximizing
functionalities under fixed resource budgets, minimizing resources subject to
minimum functionality requirements, approximating a Pareto set, or optimizing a
scalarized objective. 

This operation is analogous to exploitation in reinforcement learning (RL)~\cite{sutton1998reinforcement}. In RL, an agent may exploit a current value estimate $Q(s,a)$ by greedily selecting actions that appear valuable under the present model. In co-design, the probabilistic metric belief $\widehat m_{j,\iota}$ plays a role analogous to a value estimate, and the design sampler $\vartheta_{\iota}$ plays a role analogous to a policy over candidate designs. Refinement exploits the current belief to search for designs predicted to perform well. It should still retain some exploration over the design space, but its primary purpose is design improvement under the current model rather than physical uncertainty reduction.

Finally, we can assign a computational acquisition value to the next refinement step. Let $\xi_\mathrm{ref}\in\Xi_\mathrm{ref}$ denote a computational refinement action, such as an optimizer update, a batch of surrogate evaluations, a gradient step, or a candidate-generation step, and let $\rho_{\mathrm{ref}}(\xi_\mathrm{ref})$ denote its computational cost. The best refinement action can be selected according to
\begin{equation}
    \alpha_{\mathrm{ref},\iota}(\xi_\mathrm{ref})
    =
    \frac{\Delta_{\mathrm{ref},\iota}(\xi_\mathrm{ref})}
         {\rho_{\mathrm{ref}}(\xi_\mathrm{ref})},
    \qquad
    \xi_{\mathrm{ref},\iota+1}^{\star}
    \in
    \arg\max_{\xi_\mathrm{ref}\in\Xi_\mathrm{ref}}\alpha_{\mathrm{ref},\iota}(\xi_\mathrm{ref}),
    \label{eq:si_refinement_acquisition}
\end{equation}
where $\Delta_{\mathrm{ref},\iota}(\xi_\mathrm{ref})$ denotes the predicted improvement of the current co-design tradeoff induced by the computational refinement action. For example, it may quantify an expected improvement of the predicted admissible Pareto set defined by the current functionality estimates $\widetilde f_{\iota}$, resource estimates $\widetilde r_{\iota}$, and constraint estimates $\widetilde b_{\iota}$. Thus, $\alpha_{\mathrm{ref},\iota}$ measures the expected design improvement per unit computational cost, whereas $\alpha_{\mathrm{real},\iota}$ measures the expected value of realization per unit experimental or validation cost.

\subsection{Realization: improving the metric belief}
Realization denotes allocating resources to more faithful evaluation in order to reduce uncertainty in the metrics. A realization action may be a higher-fidelity simulation, a material coupon test, fabrication of a subcomponent, a benchtop experiment, a physical twin, a full prototype, a field trial, a user study, or a certification-oriented test. Let $\xi_{\mathrm{real}}\in\Xi_\mathrm{real}$ denote the realization action and let $\rho_\mathrm{real}(x,\xi_{\mathrm{real}})$ denote its resource cost, including time, money, experimental effort, fabrication labor, and opportunity cost.

With the composed metric belief in \Cref{eq:si_composed_metric_belief}, realization can update the metric belief through different types of observations. We denote the outcome of a realization action $\xi_{\mathrm{real}}$ applied to design $x$ by $y$. This outcome may be a direct noisy measurement of one or more true metrics, for example
\begin{equation}
    y_j
    =
    m_j(x)
    +
    \varepsilon_j(x,\xi_{\mathrm{real}}),
    \label{eq:si_noisy_metric_observation}
\end{equation}
where $\varepsilon_j$ represents measurement noise and action-dependent test variability. More generally, however, $y$ may contain lower-level calibration signals rather than a final scalar metric value, such as forces, deformations, slip events, trajectories, pressures, material hysteresis, environmental variables, or manufacturing deviations. These observations improve the metric belief by updating the posterior over the simulator, ROM, learned model, discrepancy, and metric-readout parameters. Thus, direct metric measurements correspond to the special case in which the realization outcome is itself a noisy observation of $c_j=m_j(x)$, whereas richer realization data can calibrate the predictive model and thereby improve many downstream metric beliefs simultaneously.

For a realization action $\xi_{\mathrm{real}}$ applied to design $x$ that produces an observation $y$, a real-to-sim calibration update can be written as
\begin{equation}
    p_{\iota+1}(\theta,\eta_j\mid \mathcal D_{\iota}\cup\{x,\xi_{\mathrm{real}},y\})
    \propto
    p(y\mid x,\xi_{\mathrm{real}},\theta,\eta_j)
    p_{\iota}(\theta,\eta_j\mid\mathcal D_{\iota}).
    \label{eq:si_real_to_sim_update}
\end{equation}
The updated metric belief for any candidate design $x'$ is then obtained by propagating this calibrated posterior through the prediction pipeline,
\begin{equation}
    p_{\iota+1}(\hat c_j\mid x',\mathcal D_{\iota}\cup\{x,\xi_{\mathrm{real}},y\})
    =
    \int
        p(\hat c_j\mid z,x',\eta_j)
        p(z\mid x',\theta)
        p_{\iota+1}(\theta,\eta_j\mid \mathcal D_{\iota}\cup\{x,\xi_{\mathrm{real}},y\})
        \,\mathrm{d}z\,\mathrm{d}\theta\,\mathrm{d}\eta_j .
    \label{eq:si_composed_metric_update}
\end{equation}
This is why one physical experiment can improve many metric predictions: the experiment calibrates shared simulator or ROM parameters, and all metrics that depend on those parameters inherit the reduced uncertainty. Realization, therefore, changes the probabilistic model used to evaluate designs, whereas refinement improves the design under that model.

This distinction clarifies the connection to exploration in RL. In RL, exploration collects interaction data to improve an uncertain model, value function, or policy estimate; the resulting data then changes subsequent policy-improvement steps. Analogously, realization collects physical evidence about designs, metrics, or fidelity levels to improve the metric belief; the resulting calibrated belief then changes subsequent co-design refinement steps. In this sense, realization plays a role similar to exploration or system-identification rollouts in model-based RL, while refinement plays a role similar to planning or policy improvement under the current learned model.

The analogy is useful, but it is not exact. A realization action in co-design is often much more expensive than an exploratory action in RL, because it may require fabrication, material testing, subsystem integration, human-subject evaluation, or field deployment. Moreover, holistic co-design is typically multi-objective and constrained: realization should reduce uncertainty not only about a scalar return, but also about feasibility, safety, robustness, resource use, and the predicted functionality--resource tradeoff. Thus, the question is not merely how to explore, but which design, metric, and realization fidelity should be selected to most improve the downstream co-design decision.

\subsection{Choosing what to realize}
The key algorithmic question is not simply whether prototypes are useful, but \emph{which} design should be realized, \emph{which} metric should be measured, and \emph{at what fidelity}. We denote a realization action by $(x,\xi_{\mathrm{real}})$, where $x\in\mathcal X$ is the design to be evaluated and $\xi_{\mathrm{real}}\in\Xi_{\mathrm{real}}$ specifies the realization mode, such as a material test, high-fidelity simulation, subsystem prototype, full prototype, field trial, or user study. The corresponding resource cost is denoted by $\rho_\mathrm{real}(x,\xi_{\mathrm{real}})$.

For a scalar cost-minimization problem, one classical way to select the next design is Expected Improvement (EI)~\cite{jones1998efficient,garnett2023bayesian}. Let $c_{\iota}^{\star}$ denote the current incumbent cost, for example the best realized cost observed so far or the best risk-aware predicted cost among admissible designs. Before realizing a candidate $x$, its cost is uncertain and represented by the current metric belief. The expected improvement is
\begin{equation}
    \operatorname{EI}_{\iota}(x)
    =
    \mathbb E_{\iota}
    \!\left[
        \left(c_{\iota}^{\star}-\hat c_j(x)\right)_+
    \right],
    \qquad
    (z)_+=\max\{z,0\}.
    \label{eq:si_expected_improvement}
\end{equation}
Thus, EI favors designs that are either predicted to improve upon the incumbent or uncertain enough that a large improvement remains plausible. However, holistic co-design is generally not a scalar cost-minimization problem: realization may be valuable because it improves the predicted functionality--resource tradeoff, clarifies feasibility, or reduces uncertainty in deployment-critical constraints.

We therefore write the realization acquisition directly in terms of the multi-objective co-design quantities introduced above. Let $\mathcal P^{\star}$ denote the unknown realized feasible Pareto set induced by the true functionalities, resources, and constraints, and let $\mathcal P_{\iota}$ denote the corresponding Pareto set predicted under the current metric beliefs. The information-gain component of realization can be written abstractly as the expected reduction in uncertainty over $\mathcal P^{\star}$,
\begin{equation}
    \Delta_{\mathrm{info},\iota}(x,\xi_{\mathrm{real}})
    =
    H[p_{\iota}(\mathcal P^{\star})]
    -
    \mathbb E_{y\sim p_{\iota}(y\mid x,\xi_{\mathrm{real}},\mathcal D_{\iota})}
    \left[
        H\!\left[
            p_{\iota+1}
            \left(
                \mathcal P^{\star}
                \mid
                \mathcal D_{\iota}\cup\{x,\xi_{\mathrm{real}},y\}
            \right)
        \right]
    \right],
    \label{eq:si_pareto_information_gain}
\end{equation}
where $H[\cdot]$ denotes entropy. This term prioritizes realization actions that are expected to reduce uncertainty in the parts of the feasible Pareto set that could change the co-design decision.

The performance-improvement component can be defined analogously as the expected improvement of the predicted functionality--resource tradeoff,
\begin{equation}
    \Delta_{\mathrm{perf},\iota}(x,\xi_{\mathrm{real}})
    =
    \mathbb E_{y\sim p_{\iota}(y\mid x,\xi_{\mathrm{real}},\mathcal D_{\iota})}
    \left[
        \operatorname{Imp}
        \left(
            \mathcal P_{\iota+1}(y),
            \mathcal P_{\iota}
        \right)
    \right],
    \label{eq:si_expected_pareto_improvement}
\end{equation}
where $\mathcal P_{\iota+1}(y)$ is the predicted feasible Pareto set after incorporating the hypothetical realization outcome $y$. The improvement measure $\operatorname{Imp}$ can be instantiated as hypervolume improvement, improvement in a stakeholder-relevant region of the Pareto set, or reduction of predicted constraint violation.

A cost-aware realization acquisition is then
\begin{equation}
    \alpha_{\mathrm{real},\iota}(x,\xi_{\mathrm{real}})
    =
    \frac{
        \Delta_{\mathrm{info},\iota}(x,\xi_{\mathrm{real}})
        +
        \lambda_{\mathrm{perf}}\,
        \Delta_{\mathrm{perf},\iota}(x,\xi_{\mathrm{real}})
    }
    {\rho_\mathrm{real}(x,\xi_{\mathrm{real}})},
    \label{eq:si_realization_acquisition}
\end{equation}
and the next acquisition-optimal realization action is selected as
\begin{equation}
    \left(
        x_{\mathrm{real},\iota+1}^{\star},
        \xi_{\mathrm{real},\iota+1}^{\star}
    \right)
    \in
    \arg\max_{x\in\mathcal X,\,\xi_{\mathrm{real}}\in\Xi_{\mathrm{real}}}
    \alpha_{\mathrm{real},\iota}(x,\xi_{\mathrm{real}}).
    \label{eq:si_realization_acquisition_maximizer}
\end{equation}
Here, $\lambda_{\mathrm{perf}}\geq 0$ controls how strongly realization is directed toward immediate improvement of the predicted co-design tradeoff versus uncertainty reduction. The denominator makes explicit that a full field trial, a material coupon test, and a low-cost simulation refinement should not be treated as equivalent measurements. This formulation also captures targeted realization: if two designs have similar predicted performance but one has high uncertainty in a deployment-critical metric, such as bruising, cleaning performance, or contact safety, then realization should prioritize the action that is more likely to change the selected design, the admissible set, or the predicted feasible Pareto front.

\subsection{Balancing refinement and realization}
The refinement--realization tradeoff can be summarized by the realization ratio
\begin{equation}
    \operatorname{RealR}_{\iota}
    =
    \frac{N_{\mathrm{real}}(\iota)}
         {N_{\mathrm{real}}(\iota)+N_{\mathrm{ref}}(\iota)},
    \label{eq:si_realization_ratio}
\end{equation}
where $N_{\mathrm{real}}(\iota)$ and $N_{\mathrm{ref}}(\iota)$ denote the numbers of realization and refinement iterations up to co-design iteration $\iota$, respectively. 


The limiting case $\operatorname{RealR}\rightarrow 0$ corresponds to a mostly simulation-only co-design process in which realization occurs only after an apparently optimal design has been found. This can be efficient if the surrogate is accurate, but it risks over-optimizing against the wrong metric. Indeed, even a perfect computational optimizer can only optimize the metric belief, not the true realized metric. For a scalar metric, suppose that refinement selects
\[
    x_{\iota}^{\mathrm{sim}}
    \in
    \arg\min_{x\in\mathcal X}
    \mathbb E_{\iota}[\hat c_j\mid x],
    \qquad
    x_j^{\star}
    \in
    \arg\min_{x\in\mathcal X}
    m_j(x),
\]
where $x_{\iota}^{\mathrm{sim}}$ is optimal under the current surrogate belief and $x_j^{\star}$ is optimal under the true metric. The realized suboptimality
\[
    m_j(x_{\iota}^{\mathrm{sim}})-m_j(x_j^{\star})
\]
is then controlled by the sim-to-real calibration error from \Cref{eq:si_sim_to_real_error}.
For the illustrative scalar case, if $e_{j,\iota}(x)\leq \bar e_{j,\iota}$ for all $x\in\mathcal X$, then
\[
    m_j(x_{\iota}^{\mathrm{sim}})-m_j(x_j^{\star})
    \leq
    2\bar e_{j,\iota}.
\]
Thus, in the purely computational limit, realized performance can be bounded away from the true optimum by metric-model error rather than by optimizer error. Moreover, the mean error $e_{j,\iota}$ captures only bias in the current metric estimate. The remaining belief uncertainty,
\[
    \sigma_{j,\iota}(x)
    =
    \sqrt{\operatorname{Var}_{\iota}[\hat c_j\mid x]},
\]
indicates where the optimizer may still be relying on poorly resolved surrogate predictions. Realization reduces both effects: it can decrease the calibration error $e_{j,\iota}$ by aligning the simulator, reduced-order model, or learned surrogate with the physical system, and it can decrease $\sigma_{j,\iota}$ by reducing uncertainty in the metric belief. The opposite extreme, $\operatorname{RealR}=1$, spends all iterations on measurement and validation; it may reduce uncertainty but does not actively improve the design.

Therefore, this realization ratio is not prescribed a priori. Instead, it emerges from the outer-loop scheduling decision between refinement and realization. Let
\begin{equation}
    \alpha_{\mathrm{real},\iota}^{\star}
    =
    \max_{x\in\mathcal X,\,\xi_{\mathrm{real}}\in\Xi_\mathrm{real}}
    \alpha_{\mathrm{real},\iota}(x,\xi_{\mathrm{real}}),
    \qquad
    \alpha_{\mathrm{ref},\iota}^{\star}
    =
    \max_{\xi_\mathrm{ref}\in\Xi_\mathrm{ref}}
    \alpha_{\mathrm{ref},\iota}(\xi_\mathrm{ref})
    \label{eq:si_best_ref_real_acquisitions}
\end{equation}
denote the best available realization and refinement acquisitions at iteration $\iota$. To compare refinement and realization actions, their acquisition values must be normalized to a common decision-utility scale. A simple resource-aware scheduling rule is
\begin{equation}
    s_{\iota}
    =
    \begin{cases}
        \text{realization}, &
        \text{if } \alpha_{\mathrm{real},\iota}^{\star}
        \geq
        \alpha_{\mathrm{ref},\iota}^{\star}
        \text{ and the required realization budget is available},\\[2mm]
        \text{refinement}, &
        \text{otherwise},
    \end{cases}
    \label{eq:si_ref_real_scheduler}
\end{equation}
where $\alpha_{\mathrm{real},\iota}$ already accounts for the realization cost $\rho_\mathrm{real}(x,\xi_{\mathrm{real}})$ and $\alpha_{\mathrm{ref},\iota}$ accounts for the computational cost $\rho_{\mathrm{ref}}(\xi_\mathrm{ref})$. For this comparison, the acquisition values are assumed to be normalized to a common decision-utility scale, so that information gain, predicted Pareto improvement, and computational improvement can be compared per unit resource. The realization ratio can then be written as the empirical fraction of scheduling decisions assigned to realization,
\begin{equation}
    \operatorname{RealR}_{\iota}
    =
    \frac{
        \sum_{\kappa=1}^{\iota}
        \mathbbm{1}\!\left[s_{\kappa}=\mathrm{real}\right]
    }{
        \sum_{\kappa=1}^{\iota}
        \mathbbm{1}\!\left[s_{\kappa}=\mathrm{real}\right]
        +
        \sum_{\kappa=1}^{\iota}
        \mathbbm{1}\!\left[s_{\kappa}=\mathrm{ref}\right]
    }.
    \label{eq:si_realization_ratio_scheduler}
\end{equation}
A field trial, a material coupon test, and a computational optimizer step therefore affect the schedule through their respective costs and expected value, rather than by being counted as equivalent resource expenditures.

\subsection{Practical implications}
This probabilistic formulation highlights several practical consequences for soft-robot development.

First, metric definitions should include both a computational implementation and a realization protocol. A metric such as contact safety, bruise probability, or durability is incomplete if the manuscript only states the surrogate. The co-design team should also specify how the corresponding true value will be measured, at which TRL, and how the result will update the metric belief.

Second, not all metrics require the same amount of realization. Metrics that are cheap, low-uncertainty, or only weakly relevant to the final Pareto decision can remain mostly computational. Metrics that are uncertain, safety-critical, deployment-critical, or likely to change the chosen Pareto point should receive more realization budget.

Third, shared benchmark datasets can reduce the amount of realization needed in each project. If researchers report not only final designs but also design variables, surrogate predictions, realized measurements, fabrication details, failure modes, and operating conditions, future projects can start with better metric priors. This aligns with the broader push for benchmarking and reproducibility in soft robotics~\cite{bhatia2021evolution,wang2023softzoo,baines2024need} and is particularly important for expensive metrics such as operational lifetime, manufacturing yield, cleaning robustness, and serial production cost.

Finally, the refinement--realization tradeoff should be treated as part of the optimization problem rather than as an informal project-management choice. The central decision is whether the next unit of budget is more valuable for computational search or for reducing uncertainty through realization. Framing this decision probabilistically connects holistic co-design to Bayesian optimization, active learning, and the exploration--exploitation literature in RL~\cite{sutton1998reinforcement,garnett2023bayesian}, while preserving the distinctive constraints of soft-robot development: physical fabrication, multi-objective stakeholder values, TRL gates, and deployment accountability.

\suppnote{A Selection of Holistic Co-Design Metrics}

\subsection{Safety}\label{si:sub:safety}
In this subsection, we provide a concise overview of a computational safety estimator for \emph{blunt} physical human--robot contacts that is suitable for use in a holistic co-design framework. A complete and detailed derivation is available in \cite[Chapter~3]{stolzle2025phdthesis}. The derivation proceeds in three steps: first, we introduce a local single-point contact model and derive one-dimensional projected collision dynamics; second, we define a scenario-level computational estimate, the \textbf{estimated Soft Robot Injury Severity Criterion} ($\widehat{\mathrm{SRISC}}$); and third, we lift this local estimate to a design-level computational estimate, the \textbf{estimated Soft Robot Design Hazardousness Criterion} ($\widehat{\mathrm{SRDHC}}$).

\begin{figure}[h!]
    \centering
    \includegraphics[width=0.65\linewidth]{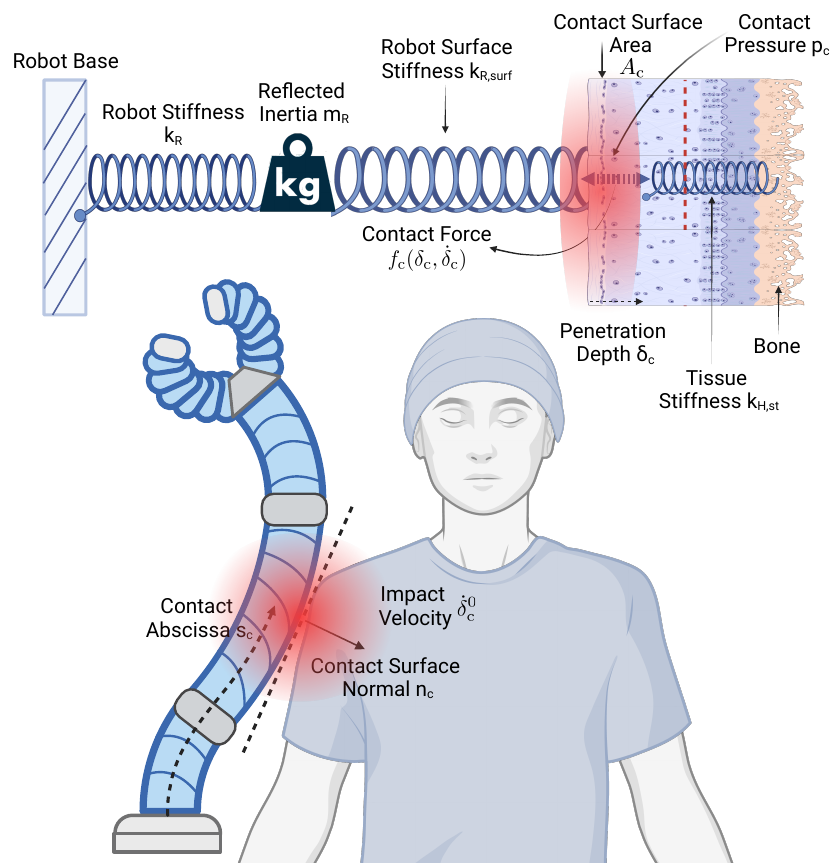}
    \caption{Illustration of the proposed safety metric for quantifying the injury severity of a local blunt contact between a soft robot and a human body part. The maximum contact pressure/stress $\max_t p_\mathrm{c}(t)=\max_t f_\mathrm{c}(t)/A_\mathrm{c}(t)$ is used as a proxy for injury risk~\cite{iso2016collaborative}. The contact force $f_\mathrm{c}(t)$ is obtained from a one-dimensional projected collision model along the contact normal $n_\mathrm c$, with penetration coordinate $\delta_\mathrm c(t)$. To obtain a conservative estimate, the human-side motion is prescribed, corresponding to the limiting case in which the human body part is effectively constrained or has inertia much larger than the reflected robot inertia $m_\mathrm R$.}
    \label{fig:injury_severity_criterion_illustration}
\end{figure}


\paragraph{Goal and true metric.}
The physical safety metric $c_{\mathrm{safety}}=m_{\mathrm{safety}}(x)$ is the hazardousness of a realized soft robot design under admissible blunt-contact scenarios. In principle, it could be determined by exhaustive collision testing as
\[
    c_{\mathrm{safety}}(x)
    =
    \sup_{\sigma_\mathrm{c}\in\Sigma_\mathrm{c}(x)}
    \max_t p_{\mathrm c}^{\mathrm{real}}(t;\sigma_\mathrm{c},x),
\]
where $p_{\mathrm c}^{\mathrm{real}}$ is the measured contact pressure during a physical collision scenario $\sigma_\mathrm{c}$. This true metric accounts for all real effects, including distributed compliance, stored elastic energy, actuation through the deformable structure, controller behavior, contact geometry, damping, material nonlinearities, and manufacturing variability. Since exhaustive real-world collision testing is infeasible during early co-design, we approximate it with a computational estimate $\hat c_{\mathrm{safety}}=\hat m_{\mathrm{safety}}(x)$ obtained from projected collision dynamics. The scenario-level estimate $\widehat{\mathrm{SRISC}}$ evaluates one compact collision scenario, and the design-level estimate $\widehat{\mathrm{SRDHC}}$ aggregates this quantity over feasible states, contact geometries, and admissible inputs.

\paragraph{Local contact geometry.}
A potential collision is represented by a single active point contact with contact geometry
\[
    G_\mathrm{c} := (s_\mathrm c,n_\mathrm c) \in \mathcal G_\mathrm{c},
    \qquad
    \mathcal C := (0,L]\times\mathbb S^2,
\]
where $s_\mathrm c$ is the closest backbone coordinate, $n_\mathrm c$ is the local contact normal, and $L$ is the robot length. Let $r(q,s)\in\mathbb R^3$ denote the translational part of the forward kinematics and let $r_\mathrm c$ denote the contact point. The scalar robot-side normal contact Jacobian is
\begin{equation}\label{eq:summary_contact_jacobian}
    J_\mathrm c(q,G_\mathrm{c})
    =
    n_\mathrm c^\top
    \left(
        J_\mathrm r(q,s_\mathrm c)
        -
        \left[
            r_\mathrm c-r(q,s_\mathrm c)
        \right]_\times
        J_\Omega(q,s_\mathrm c)
    \right)
    \in\mathbb R^{1\times n},
\end{equation}
so that the robot-side normal contact velocity is
\[
    \dot{x}_\mathrm R = J_\mathrm c(q,G_\mathrm{c})\dot q .
\]
By virtual work, the scalar normal contact force $f_\mathrm{c}$ contributes the generalized force
\[
    \tau_\mathrm c = -J_\mathrm c^\top(q,G_\mathrm{c}) f_\mathrm{c} .
\]

\paragraph{Projected collision dynamics.}
The soft robot dynamics are written as
\[
    M(q)\ddot q
    + C(q,\dot q)\dot q
    + \partial_q \mathcal U(q)
    + D\dot q
    =
    A_\mathrm u(q)u
    + \tau_\mathrm c ,
\]
where $u$ is the control input and $A_\mathrm u(q)$ maps it into generalized actuation forces. Projecting these dynamics onto the contact normal gives a scalar collision coordinate. We prescribe the human-side motion during contact as
\[
    \dot{x}_\mathrm H(t)=-v_\mathrm H,
    \qquad
    v_\mathrm H\in\mathbb R_{\ge 0},
\]
which yields the penetration coordinate
\[
    \delta_\mathrm c(t)=x_\mathrm R(t)-x_\mathrm H(t),
    \qquad
    \dot{\delta}_\mathrm c(t)=J_\mathrm c(q,G_\mathrm{c})\dot q(t)+v_\mathrm H .
\]
During active compression, the normal contact law is approximated by a linear spring-damper law
\[
    f_\mathrm{c}(\delta_\mathrm c,\dot\delta_\mathrm c)
    =
    k_\mathrm c\delta_\mathrm c+d_\mathrm c\dot\delta_\mathrm c,
    \qquad
    \delta_\mathrm c>0,
\]
where $k_\mathrm c$ is the effective normal contact stiffness of the robot--human interface. For the closed-form safety metric, the velocity-dependent terms are neglected over the short collision interval, including contact damping, structural damping, and Coriolis/centrifugal effects. The contact event is also linearized around the contact-onset configuration $q_\mathrm c^0$.

Let
\[
    m_\mathrm R
    :=
    \Lambda_\mathrm c(q_\mathrm c^0,G_\mathrm{c}),
    \qquad
    \Lambda_\mathrm c(q,G_\mathrm{c})
    :=
    \left(
        J_\mathrm c(q,G_\mathrm{c})M^{-1}(q)J_\mathrm c^\top(q,G_\mathrm{c})
    \right)^{-1}
\]
denote the reflected inertia at the contact point. Further define the dynamically consistent pseudo-inverse
\[
    J_{\mathrm c,M}^{+}(q,G_\mathrm{c})
    :=
    M^{-1}(q)J_\mathrm c^\top(q,G_\mathrm{c})
    \left(
        J_\mathrm c(q,G_\mathrm{c})M^{-1}(q)J_\mathrm c^\top(q,G_\mathrm{c})
    \right)^{-1}.
\]
The projected potential force at contact onset is
\[
    f_{\mathcal U}^{\mathrm c0}
    :=
    J_{\mathrm c,M}^{+\top}(q_\mathrm c^0,G_\mathrm{c}) \, \partial_q\mathcal U(q_\mathrm c^0),
\]
and the dynamically projected local stiffness in the collision direction is
\[
    k_\mathrm R
    :=
    \left.
    \frac{\partial}{\partial q}
    \left[
        J_{\mathrm c,M}^{+\top}(q,G_\mathrm{c})\partial_q\mathcal U(q)
    \right]
    \right|_{q=q_\mathrm c^0}
    J_{\mathrm c,M}^{+}(q_\mathrm c^0,G_\mathrm{c}).
\]
Finally, the worst-case projected actuation during the contact interval is bounded by
\[
    f_\mathrm u
    :=
    \sup_{t\in\mathcal I_\mathrm c}
    \left|
        A_{\mathrm u,\mathrm c}u(t)
    \right|,
    \qquad
    A_{\mathrm u,\mathrm c}
    :=
    J_{\mathrm c,M}^{+\top}(q_\mathrm c^0,G_\mathrm{c})A_\mathrm u(q_\mathrm c^0).
\]
The simplified scalar collision dynamics are then
\begin{equation}\label{eq:summary_simplified_collision_dynamics}
    m_\mathrm R\ddot{\delta}_\mathrm c
    +
    (k_\mathrm R+k_\mathrm c)\delta_\mathrm c
    =
    f_\mathrm u
    -
    f_{\mathcal U}^{\mathrm c0}.
\end{equation}

\paragraph{Closed-form peak contact force.}
To express the solution compactly, define the total scalar collision stiffness, static penetration offset, and collision frequency as
\begin{equation}\label{eq:summary_collision_stiffness_offset}
\begin{split}
    k_\Sigma(q,G_\mathrm{c})
    &:=
    k_\mathrm R(q,G_\mathrm{c})+k_\mathrm c(G_\mathrm{c}),\\
    \delta_\mathrm{eq}(q,f_\mathrm u,G_\mathrm{c})
    &:=
    \frac{
        f_\mathrm u-f_{\mathcal U}^{\mathrm c0}(q,G_\mathrm{c})
    }{
        k_\Sigma(q,G_\mathrm{c})
    },\\
    \omega_\mathrm c(q,G_\mathrm{c})
    &:=
    \sqrt{
        \frac{k_\Sigma(q,G_\mathrm{c})}{m_\mathrm R(q,G_\mathrm{c})}
    } .
\end{split}
\end{equation}
With initial penetration depth $\delta_\mathrm c^0$ and initial approaching penetration velocity $v_\delta=\dot\delta_\mathrm c^0$, the closed-form upper bound on the maximum elastic contact force is
\begin{equation}\label{eq:summary_peak_contact_force}
    \overline f_\mathrm{c}(q,v_\delta,f_\mathrm u,G_\mathrm{c})
    =
    k_\mathrm c
    \left[
        \delta_\mathrm{eq}
        +
        \sqrt{
            \left(
                \delta_\mathrm c^0-\delta_\mathrm{eq}
            \right)^2
            +
            \left(
                \frac{v_\delta}{\omega_\mathrm c}
            \right)^2
        }
    \right].
\end{equation}
Unless stated otherwise, the nominal contact-onset value $\delta_\mathrm c^0=0$ is used. This closed-form expression avoids simulating the transient collision dynamics while retaining the dependence on local reflected inertia, local stiffness, contact stiffness, actuation, and relative impact velocity.

\paragraph{Soft Robot Injury Severity Criterion (SRISC): scenario-specific severity.}
For a compact local collision scenario
\[
    \sigma_\mathrm{c} := (q,\dot q,G_\mathrm{c},u),
\]
we define the projected actuation map
\[
    B_\mathrm u(q,G_\mathrm{c})
    :=
    J_{\mathrm c,M}^{+\top}(q,G_\mathrm{c}) \, A_\mathrm u(q)
    \in\mathbb R^{1\times m},
\]
the approaching penetration velocity
\[
    v_\delta(q,\dot q,G_\mathrm{c})
    :=
    \left[
        J_\mathrm c(q,G_\mathrm{c}) \, \dot q+v_\mathrm H
    \right]_+,
    \qquad
    [a]_+ := \max\{a,0\},
\]
and the projected actuation magnitude
\[
    f_\mathrm u(q,G_\mathrm{c},u)
    :=
    \left|
        B_\mathrm u(q,G_\mathrm{c})u
    \right|.
\]
Let
\[
    A_\mathrm c^{\min}(G_\mathrm{c})
    :=
    \min_{t\in\mathcal I_\mathrm c} A_\mathrm c(t)
\]
denote the minimum contact area during the local contact event. The SRISC is the resulting upper bound on maximum contact pressure:
\begin{equation}\label{eq:summary_srisc_definition}
\begin{split}
    \widehat{\mathrm{SRISC}}(\sigma_\mathrm{c})
    &:=
    \frac{
        \overline f_\mathrm{c}(q,v_\delta,f_\mathrm u,G_\mathrm{c})
    }{
        A_\mathrm c^{\min}(G_\mathrm{c})
    }.\\
\end{split}
\end{equation}
Thus,
\[
    \max_t p_\mathrm c(t)
    =
    \max_t
    \frac{f_\mathrm{c}(t)}{A_\mathrm c(t)}
    \leq
    \frac{
        \overline f_\mathrm{c}(q,v_\delta,f_\mathrm u,G_\mathrm{c})
    }{
        A_\mathrm c^{\min}(G_\mathrm{c})
    }
    =
    \widehat{\mathrm{SRISC}}(\sigma_\mathrm{c}).
\]
The SRISC is therefore a local injury-severity proxy for one collision scenario.

\paragraph{Closed-loop systems: how the controller enters SRISC.}
For an integrated robot--controller system, the controller affects SRISC in two ways. First, it determines the projected actuation magnitude $f_\mathrm u(q,G_\mathrm{c},u)$. If the controller is treated as a black box, actuator limits or input bounds $u\in\mathbb U$ can be used to compute a worst-case projected actuation bound. Second, for structured feedback controllers, the feedback law can change the effective local collision stiffness.

For example, in a fully actuated setting with $A_\mathrm u(q)=I^n$, consider a PD+feedforward regulator
\[
    u(q,\dot q)
    =
    \partial_q\mathcal U(q^\mathrm d)
    +
    K_\mathrm p(q^\mathrm d-q)
    -
    K_\mathrm d\dot q .
\]
The closed-loop dynamics become
\[
    M\ddot q
    +
    C\dot q
    +
    \partial_q\mathcal U(q)
    +
    K_\mathrm p q
    +
    (D+K_\mathrm d)\dot q
    =
    \partial_q\mathcal U(q^\mathrm d)
    +
    K_\mathrm p q^\mathrm d
    +
    \tau_\mathrm c .
\]
Neglecting velocity-dependent terms during the short collision interval, the simplified collision dynamics take the form
\[
    m_\mathrm R\ddot{\delta}_\mathrm c
    +
    (k_\mathrm R+k_\mathrm c)\delta_\mathrm c
    =
    J_{\mathrm c,M}^{+\top}(q_\mathrm c^0,G_\mathrm{c})
    \left(
        \partial_q\mathcal U(q^\mathrm d)
        +
        K_\mathrm p q^\mathrm d
        -
        \partial_q\mathcal U(q_\mathrm c^0)
    \right).
\]
The proportional feedback term contributes to the effective local stiffness through
\[
    k_\mathrm R
    =
    \left.
    \frac{\partial}{\partial q}
    \left[
        J_{\mathrm c,M}^{+\top}(q,G_\mathrm{c})
        \left(
            \partial_q\mathcal U(q)
            +
            K_\mathrm p q
        \right)
    \right]
    \right|_{q=q_\mathrm c^0}
    J_{\mathrm c,M}^{+}(q_\mathrm c^0,G_\mathrm{c}).
\]
Consequently, increasing proportional gains can increase the local stiffness felt during impact and thereby increase the peak force/pressure, while softer gains can reduce this contribution.

\paragraph{Soft Robot Design Hazardousness Criterion (SRDHC): worst-case design risk.}
The SRISC evaluates one local collision scenario. For design evaluation, however, we are interested in the worst-case hazardousness of the robot over its feasible states, contact locations, contact orientations, and admissible inputs. Let $\mathcal Q$ and $\dot{\mathcal Q}$ denote the feasible configuration and velocity sets, and let $\mathbb U$ denote the admissible input set. The feasible collision-scenario set is
\[
    \Sigma_\mathrm{c}
    :=
    \mathcal Q
    \times
    \dot{\mathcal Q}
    \times
    \mathcal C
    \times
    \mathbb U .
\]
The SRDHC is defined as
\begin{equation}\label{eq:summary_srdhc_definition}
    \widehat{\mathrm{SRDHC}}
    :=
    \sup_{\sigma_\mathrm{c}\in\Sigma_\mathrm{c}}
    \widehat{\mathrm{SRISC}}(\sigma_\mathrm{c}).
\end{equation}

\paragraph{Computable conservative SRDHC bound.}
For efficient design-level evaluation, global bounds on the approaching penetration velocity and projected actuation can be used. Define
\[
    \Sigma_{\mathrm{c},v}
    :=
    \mathcal Q\times\dot{\mathcal Q}\times\mathcal C,
    \qquad
    \bar v_\delta
    :=
    \sup_{(q,\dot q,G_\mathrm{c})\in\Sigma_{\mathrm{c},v}}
    v_\delta(q,\dot q,G_\mathrm{c}).
\]
A simple norm bound is
\[
\begin{split}
    \bar v_\delta
    &\leq
    \bar J_\mathrm c\,\bar{\dot q}+v_\mathrm H,\\
    \bar J_\mathrm c
    &:=
    \sup_{(q,G_\mathrm{c})\in\mathcal Q\times\mathcal C}
    \|J_\mathrm c(q,G_\mathrm{c})\|_2,\\
    \bar{\dot q}
    &:=
    \sup_{\dot q\in\dot{\mathcal Q}}
    \|\dot q\|_2 .
\end{split}
\]
Similarly, the projected actuation bound is
\[
\begin{split}
    \bar f_\mathrm u(q,G_\mathrm{c})
    &:=
    \sup_{u\in\mathbb U}
    |B_\mathrm u(q,G_\mathrm{c})u|,\\
    \bar f_\mathrm u(q,G_\mathrm{c})
    &\leq
    \|B_\mathrm u(q,G_\mathrm{c})\|_2\,\bar u,
    \qquad
    \bar u
    :=
    \sup_{u\in\mathbb U}\|u\|_2 .
\end{split}
\]
This yields the conservative design-level upper bound
\begin{equation}\label{eq:summary_srdhc_upper_bound}
    \overline{\widehat{\mathrm{SRDHC}}}
    :=
    \sup_{(q,G_\mathrm{c})\in\mathcal Q\times\mathcal C}
    \widehat{\mathrm{SRISC}}
    \big(
        q,
        \bar v_\delta,
        \bar f_\mathrm u(q,G_\mathrm{c}),
        G_\mathrm{c}
    \big),
    \qquad
    \widehat{\mathrm{SRDHC}}
    \leq
    \overline{\widehat{\mathrm{SRDHC}}} .
\end{equation}

\paragraph{Interpretation and use.}
The computational safety metric $\hat c_{\mathrm{safety}}(x)=\overline{\widehat{\mathrm{SRDHC}}}(x)$ is an efficiently evaluable estimate of the true physical hazardousness $c_{\mathrm{safety}}(x)$. It is suitable for screening morphologies, materials, actuator selections, discretizations, and closed-loop controllers before physical testing. The estimator remains grounded in the soft-robot dynamics through the reflected inertia $m_\mathrm R$, projected stiffness $k_\mathrm R$, projected actuation $f_\mathrm u$, and contact geometry $c$, while its computational efficiency follows from the closed-form peak-force expression in~\eqref{eq:summary_peak_contact_force}. Physical collision experiments can then be used to calibrate or validate the gap between $\hat c_{\mathrm{safety}}$ and $c_{\mathrm{safety}}$.

\subsection{Manufacturability}
\label{si:sub:manufacturability}

Within the holistic co-design framework, manufacturability is treated as an
expected cost rather than as a qualitative score. The true manufacturability
metric $c_{\mathrm{mfg}}=m_{\mathrm{mfg}}(x)$ is the realized manufacturing cost
per accepted robot, obtained from actual fabrication batches and accounting for
real material use, labor, equipment occupancy, energy consumption, and yield
losses. During computational co-design, this physical quantity is replaced by an
estimate $\hat c_{\mathrm{mfg}}=\hat m_{\mathrm{mfg}}(x)$. For a soft robot
design $x$, process route $\phi$, and batch size $B$, we define
\[
\hat c_{\mathrm{mfg}}(x,\phi;B)
=
\frac{\hat C^{\mathrm{tool}}(x,\phi)}{B}
+
\sum_{o\in\mathcal{O}_{\phi}}
\hat\chi_o(x,\phi) \hat C_o(x,\phi),
\]
where $\mathcal{O}_{\phi}$ is the ordered set of manufacturing, assembly, and
inspection operations. The tooling term $\hat C^{\mathrm{tool}}/B$ amortizes molds,
fixtures, jigs, and other non-recurring resources over the batch size. This
decomposition follows standard manufacturing-cost and design-for-manufacture-
and-assembly reasoning, in which product cost is assembled from material,
purchased components, labor, machine time, overhead, and inspection
\cite{boothroyd2010product,ostwald2004cost,niazi2006cost}. For additively
manufactured parts, related cost models further emphasize machine occupancy,
build volume, support material, ancillary process steps, and build failure
\cite{alexander1998part,ruffo2006cost,thompson2016design,baumers2017adoption}.

The estimated nominal cost of operation $o$ is
\[
\hat C_o(x,\phi)
=
\hat C_o^{\mathrm{mat}}(x,\phi)
+
\hat C_o^{\mathrm{buy}}(x,\phi)
+
a_o^{\mathrm{lab}}\hat t_o^{\mathrm{lab}}(x,\phi)
+
a_o^{\mathrm{equip}}\hat t_o^{\mathrm{equip}}(x,\phi)
+
c_o^{\mathrm{energy}}\hat E_o(x,\phi).
\]
Here, $\hat C_o^{\mathrm{mat}}$ is estimated process material cost, $\hat C_o^{\mathrm{buy}}$ is the
estimated cost of purchased components, $a_o^{\mathrm{lab}}$ and $a_o^{\mathrm{equip}}$
are labor and equipment rates, $\hat t_o^{\mathrm{lab}}$ and
$\hat t_o^{\mathrm{equip}}$ are estimated active labor time and equipment occupancy time, and
$\hat E_o$ is estimated consumed energy. 

The factor $\hat\chi_o(x,\phi)\geq 1$ is an estimated yield-loss multiplier. It converts the
estimated nominal cost of operation $o$ into its expected contribution per accepted robot.
If $\hat y_r(x,\phi)\in(0,1]$ is the estimated pass probability of operation $r$, then
\[
\hat\chi_o(x,\phi)
=
\left(
\prod_{r\succeq o} \hat y_r(x,\phi)
\right)^{-1},
\]
where $r\succeq o$ denotes operation $o$ and all downstream operations. Yield is
therefore absorbed into the scalar cost estimate: low-yield designs become expensive
because more material, labor, machine time, and inspection effort are expected to
be consumed per accepted robot.

The required inputs are restricted to quantities that can be obtained from
CAD/BOM data, supplier catalogs, fabrication handbooks, slicer outputs, simple
heuristic process rules, or limited empirical tests. For example, material
volumes and part counts come from CAD and the bill of materials; print times and
support volumes come from slicer output; material and component prices come from
supplier catalogs; curing and degassing times come from material datasheets or
fabrication protocols; and yield multipliers can be calibrated from simple
fabrication tests such as leak tests, proof-pressure tests, bond tests, or
demolding trials.

\paragraph{Specific instantiation: cast pneumatic soft robot.}

We now instantiate the metric for a cast pneumatic soft robot, a representative
class of soft robots whose fabrication often involves elastomer casting,
pneumatic chambers, embedded channels, bonds, seals, and pressure testing
\cite{rus2015design,shepherd2011multigait,ilievski2011soft,mosadegh2014pneumatic}.
Such systems are typically fabricated through multi-step workflows involving
molds or cores, casting, curing, demolding, bonding, pneumatic integration, and
testing \cite{marchese2015recipe,holland2014softtoolkit}. More recent integrated
fabrication strategies aim to reduce the amount of manual iteration and
supervised process complexity in pneumatic actuator fabrication
\cite{wehner2016integrated,silva2024singlecasting}.

Consider a pneumatic body cast from silicone using a printed mold and removable
or sacrificial internal cores. A typical route is
\[
\mathcal{O}_{\phi}^{\mathrm{pneu}}
=
\{
\mathrm{print},
\mathrm{cast},
\mathrm{cure},
\mathrm{demold},
\mathrm{bond},
\mathrm{integrate},
\mathrm{test}
\}.
\]
The manufacturing cost per accepted robot is
\[
\begin{aligned}
\hat c_{\mathrm{mfg}}^{\mathrm{pneu}}(x,\phi;B)
=
&\frac{C^{\mathrm{tool}}(x,\phi)}{B}
+
\chi_{\mathrm{print}}C_{\mathrm{print}}
+
\chi_{\mathrm{cast}}C_{\mathrm{cast}}
+
\chi_{\mathrm{cure}}C_{\mathrm{cure}} \\
&+
\chi_{\mathrm{demold}}C_{\mathrm{demold}}
+
\chi_{\mathrm{bond}}C_{\mathrm{bond}}
+
\chi_{\mathrm{int}}C_{\mathrm{int}}
+
\chi_{\mathrm{test}}C_{\mathrm{test}} .
\end{aligned}
\]

The printing cost for molds and cores is estimated from slicer output,
\[
C_{\mathrm{print}}
=
c_{\mathrm{print}}^{\mathrm{mat}} \, m_{\mathrm{print}}(x,\phi)
+
a_{\mathrm{print}}^{\mathrm{equip}} \, t_{\mathrm{print}}(x,\phi)
+
a_{\mathrm{print}}^{\mathrm{lab}} \, t_{\mathrm{setup,print}},
\]
where $m_{\mathrm{print}}$ includes mold, core, and support material. The casting
cost is
\[
C_{\mathrm{cast}}
=
c_{\mathrm{sil}} \, m_{\mathrm{sil}}(x)
+
c_{\mathrm{release}} \, m_{\mathrm{release}}(x,\phi)
+
a_{\mathrm{lab}} \, t_{\mathrm{cast}}(x,\phi)
+
a_{\mathrm{vac}} \, t_{\mathrm{degas}}(\phi),
\]
where $m_{\mathrm{sil}}$ is the silicone mass, $m_{\mathrm{release}}$ captures
release agent or surface preparation, and $t_{\mathrm{cast}}$ includes mixing,
pouring, and cleanup. Curing is separated because it often requires little active
labor but can occupy an oven or controlled-temperature area:
\[
C_{\mathrm{cure}}
=
a_{\mathrm{oven}}t_{\mathrm{cure}}(\phi)
+
a_{\mathrm{lab}}t_{\mathrm{load/unload}}.
\]

Demolding and bonding costs depend strongly on geometry,
\[
C_{\mathrm{demold}}
=
a_{\mathrm{lab}}
\left(
t_{\mathrm{demold},0}
+
n_{\mathrm{core}}(x)t_{\mathrm{core}}
+
L_{\mathrm{core}}(x)t_{\mathrm{core}}'
\right),
\]
\[
C_{\mathrm{bond}}
=
c_{\mathrm{adh}}m_{\mathrm{adh}}(x)
+
a_{\mathrm{lab}}
\left(
t_{\mathrm{align}}
+
L_{\mathrm{seal}}(x)t_{\mathrm{seal}}'
+
n_{\mathrm{patch}}(x)t_{\mathrm{patch}}
\right),
\]
where $n_{\mathrm{core}}$ is the number of internal cores, $L_{\mathrm{core}}$ is
the total core length, $L_{\mathrm{seal}}$ is the total bonded or sealed length,
and $n_{\mathrm{patch}}$ counts local sealing patches or reinforcements.

The pneumatic integration cost includes ports, fittings, tubing, valves, and
sensors,
\[
\begin{aligned}
C_{\mathrm{int}}
=
&n_{\mathrm{port}}(x)c_{\mathrm{port}}
+
n_{\mathrm{fit}}(x)c_{\mathrm{fit}}
+
L_{\mathrm{tube}}(x)c_{\mathrm{tube}}
+
n_{\mathrm{valve}}(x)c_{\mathrm{valve}} \\
&+
n_{\mathrm{sens}}(x)c_{\mathrm{sens}}
+
a_{\mathrm{lab}}
\left(
n_{\mathrm{port}}(x)t_{\mathrm{port}}
+
n_{\mathrm{fit}}(x)t_{\mathrm{fit}}
+
L_{\mathrm{tube}}(x)t_{\mathrm{tube}}'
\right).
\end{aligned}
\]
The testing cost is based on the number of independently pressurized chambers,
\[
C_{\mathrm{test}}
=
\left(a_{\mathrm{lab}}+a_{\mathrm{rig}}\right)
n_{\mathrm{ch}}(x)
\left(
t_{\mathrm{press}}
+
t_{\mathrm{hold}}
+
t_{\mathrm{inspect}}
\right),
\]
where the test protocol may include a leak test at operating pressure
$p_{\mathrm{op}}(x)$ and a proof-pressure test
$p_{\mathrm{proof}}=\alpha_{\mathrm{proof}}p_{\mathrm{op}}(x)$.

The yield multipliers are computed from simple pass probabilities associated
with the same operations,
\[
\chi_{\mathrm{print}}
=
\left(
y_{\mathrm{print}}
y_{\mathrm{cast}}
y_{\mathrm{demold}}
y_{\mathrm{bond}}
y_{\mathrm{leak}}
y_{\mathrm{proof}}
\right)^{-1},
\qquad
\chi_{\mathrm{test}}
=
\left(
y_{\mathrm{leak}}
y_{\mathrm{proof}}
\right)^{-1}.
\]
These pass probabilities can initially be defined by threshold rules. For
example,
\[
y_{\mathrm{print}}
=
f_{\mathrm{print}}
\left(
t_{\mathrm{wall,min}}(x),
d_{\mathrm{chan,min}}(x),
V_{\mathrm{support}}(x,\phi)
\right),
\]
\[
y_{\mathrm{demold}}
=
f_{\mathrm{demold}}
\left(
\theta_{\mathrm{draft,min}}(x),
\frac{L_{\mathrm{core,max}}(x)}{d_{\mathrm{core,min}}(x)}
\right),
\]
\[
y_{\mathrm{bond}}
=
f_{\mathrm{bond}}
\left(
w_{\mathrm{bond,min}}(x),
A_{\mathrm{bond}}(x),
L_{\mathrm{seal}}(x)
\right),
\]
and
\[
y_{\mathrm{proof}}
=
f_{\mathrm{proof}}
\left(
\frac{t_{\mathrm{wall,min}}(x)}
{p_{\mathrm{proof}}(x)r_{\mathrm{ch,max}}(x)}
\right).
\]
Here, $t_{\mathrm{wall,min}}$ is the minimum chamber wall thickness,
$d_{\mathrm{chan,min}}$ is the minimum channel diameter,
$V_{\mathrm{support}}$ is the support volume, $\theta_{\mathrm{draft,min}}$ is
the minimum draft angle, $L_{\mathrm{core,max}}/d_{\mathrm{core,min}}$ measures
core slenderness, $w_{\mathrm{bond,min}}$ is the minimum bond width,
$A_{\mathrm{bond}}$ is the bond area, and $r_{\mathrm{ch,max}}$ is the maximum
local chamber radius. These functions can be simple piecewise rules from a
fabrication handbook and later replaced by empirical fits from small test
campaigns. In this way, the pneumatic example remains specific and computable:
geometry determines material usage, operation time, purchased components, and
yield-loss multipliers, all of which enter the single scalar computational estimate
$\hat c_{\mathrm{mfg}}^{\mathrm{pneu}}$.

\subsection{Durability}\label{si:sub:durability}

Within our holistic co-design view, durability is not treated only as catastrophic failure after a certain number of cycles, but more broadly as the ability of a soft robot to preserve repeatable mechanical behavior over prolonged operation. This is especially important for soft robots, where wear, hysteresis, plastic deformation, and history-dependent effects can progressively degrade behavior even before structural failure occurs~\cite{stella2024toward,ouyoucef2025durability}.

The true durability metrics $c_{\mathrm{dur},r}=m_{\mathrm{dur},r}(x)$ are experimentally grounded quantities obtained from repeated-use tests, cyclic loading, recovery experiments, or accelerated-life protocols. During computational co-design, they are replaced by estimates $\hat c_{\mathrm{dur},r}=\hat m_{\mathrm{dur},r}(x)$ obtained from simulations, reduced-order models, surrogate hysteresis models, or finite-element analyses. We consider three complementary quantities,
\[
\hat c_{\mathrm{dur},E}(x), \qquad
\hat c_{\mathrm{dur,hys}}(x), \qquad
\hat c_{\mathrm{dur,rec}}(x),
\]
which estimate, respectively, trajectory-induced mechanical demand, cyclic dissipation, and recovery after extreme deformation. These estimated quantities are intended as computational proxies for corresponding measured durability costs $c_{\mathrm{dur},E}$, $c_{\mathrm{dur,hys}}$, and $c_{\mathrm{dur,rec}}$.

This decomposition is motivated by two complementary views from the literature. On the one hand, repeated motions can accumulate strain energy and mechanically load the structure over time, making trajectory-dependent quantities useful proxies for fatigue-related demand~\cite{ouyoucef2025durability}. On the other hand, recent work by Stella \emph{et al.} emphasizes experimentally accessible durability metrics for architected soft robots based on hysteresis during cyclic loading and transient/steady-state recovery after large deformations~\cite{stella2024toward}. We therefore retain these quantities as distinct, interpretable estimates rather than aggregating them into a single scalar score.

\begin{itemize}
    \item \emph{Trajectory-induced strain energy}
    $\hat c_{\mathrm{dur},E}(x)$ quantifies how mechanically demanding an executed policy is for the soft structure. Following~\cite{ouyoucef2025durability}, a natural proxy is the strain energy accumulated along the motion. Denoting by $\hat q(t)$ the simulated or model-predicted strain coordinates, by $\bar{q}$ the zero-strain configuration and by $K$ the corresponding stiffness matrix, we define
    \[
    \hat E_\mathrm{K}(t)= \frac{1}{2} \, (\hat q(t)-\bar{q})^{\top}K(\hat q(t)-\bar{q}),
    \]
    and use the time-averaged trajectory cost
    \[
    \hat c_{\mathrm{dur},E}(x)
    =
    \frac{1}{H}\int_{0}^{H} \hat E_\mathrm{K}(t)\,dt.
    \]
    For systems where local stress concentration is critical, the maximum local strain energy can be used instead of, or alongside, this global quantity~\cite{ouyoucef2025durability}.

    $\hat c_{\mathrm{dur},E}(x)$ can be estimated computationally either via a closed-loop simulation with a ROM that exhibits a physical structure~\cite{mathew2025reduced, valadas2025learning, alkayas2025soft, stolzle2024input} or with a high-fidelity simulator (e.g., FEM).

    \item \emph{Cyclic dissipation}
    $\hat c_{\mathrm{dur,hys}}(x)$ quantifies the energy dissipated through hysteresis during a representative loading cycle. In line with the compression-test metric proposed by Stella \emph{et al.}~\cite{stella2024toward} for architected soft robotic modules, we define
    \[
    \hat c_{\mathrm{dur,hys}}(x)
    =
    \frac{\hat E_{\mathrm{hyst}}(x)}{\hat E_{\mathrm{in}}(x)},
    \]
    where $\hat E_{\mathrm{hyst}}$ is the estimated area enclosed by the loading--unloading curve and $\hat E_{\mathrm{in}}$ is the estimated input mechanical energy. This quantity captures viscoelastic losses and provides an experimentally accessible proxy for repeatability degradation under repeated loading.

    $\hat c_{\mathrm{dur,hys}}(x)$ can be computationally estimated via surrogate models (e.g., Bouc-Wen hysteresis model~\cite{bouc1967forced, wen1976method}) or FEM simulations.

    \item \emph{Recovery after extreme deformation}
    $\hat c_{\mathrm{dur,rec}}(x)$ measures how well the robot returns toward its nominal state after a large deformation or pre-strain event. Consistent with the step-response characterization proposed by Stella \emph{et al.}~\cite{stella2024toward}, a practical formulation combines a transient recovery term with a residual-deformation term:
    \[
    \hat c_{\mathrm{dur,rec}}(x)
    =
    \lambda_{1}\,\frac{\hat T_{80}(x)}{T_{\mathrm{ref}}}
    +
    \lambda_{2}\,\frac{\hat e_{\mathrm{ss}}(x)}{e_{\mathrm{ref}}},
    \]
    where $\hat T_{80}$ is the estimated time needed to recover $80\%$ of the undeformed state, $\hat e_{\mathrm{ss}}$ is the estimated residual steady-state deformation after a prescribed settling interval, and $\lambda_{1},\lambda_{2}$ tune the relative importance of transient and residual effects.

    $\hat c_{\mathrm{dur,rec}}$ can be computationally estimated via surrogate elasto-plastic models or FEM simulations.
\end{itemize}


Importantly, durability is not purely a material property: it emerges from the interaction between geometry, fabrication route, and control. The same morphology may exhibit very different long-term behavior depending on the executed trajectory and controller, which is why durability-aware planning and control should be included in the co-design problem rather than treated only after fabrication~\cite{ouyoucef2025durability}.

A particularly important direction for future work will be to connect these experimentally grounded durability metrics to explicit lifetime estimates, for instance, through accelerated-life testing and reliability models that map hysteresis, recovery behavior, and trajectory-induced mechanical demand to expected cycles-to-failure or useful operating hours.

\subsection{Sustainability}
\label{si:sub:sustainability}

Sustainability is inherently multidimensional. A soft robot design may differ
from another design in carbon footprint, toxicity, water consumption, critical
material use, non-recyclable waste, repairability, and end-of-life recovery.
A single scalar sustainability score across all of these dimensions would
require application-dependent value judgments. Therefore, we focus here on one
physically interpretable impact category: the life-cycle carbon footprint,
measured in $\mathrm{kg\,CO_2e}$. This choice is aligned with product-level
life-cycle assessment and product carbon-footprint standards
\cite{iso14040,iso14044,iso14067,ghgprotocol2011product}. Other sustainability
dimensions, such as toxicity, water use, or critical-material consumption, can be
treated as additional objectives or constraints in the holistic co-design problem.

Let $x\in\mathcal{X}$ denote the soft robot design, including geometry, material
selection, actuator layout, manufacturing route, assembly structure, and control
parameters. Let $H$ denote the mission horizon, specified either by a number of
task attempts $N_H$ or by an operating time $T_H$. Following the life-cycle
assessment principle that products should be compared with respect to a common
functional unit \cite{iso14040,iso14044}, we define the expected useful output
\[
Y(x;H)
=
\mathbb{E}\left[
\sum_{\ell=1}^{N_H} y_\ell(x)
\right],
\]
where $y_\ell(x)$ is the useful service delivered during task attempt $\ell$.
For a repeated manipulation, inspection, or harvesting task, this can be
approximated by
\[
Y(x;H)
=
N_H\,p_{\mathrm{succ}}(x)\,y_{\mathrm{task}}(x),
\]
where $p_{\mathrm{succ}}(x)$ is the probability of successful task execution and
$y_{\mathrm{task}}(x)$ is the useful output per successful task, such as the
number of successful picks, harvested mass, inspected area, or operating hours.

We define the true functional life-cycle carbon intensity as
\[
c_{\mathrm{sust}}(x;H)
=
m_{\mathrm{sust}}(x;H)
=
\frac{C_{\mathrm{life}}(x;H)}{Y(x;H)}
\qquad
\left[
\frac{\mathrm{kg\,CO_2e}}{\mathrm{functional\ unit}}
\right],
\]
where $C_{\mathrm{life}}$ and $Y$ would be obtained from complete product carbon accounting and realized functional output. During computational co-design, we instead use
\[
\hat c_{\mathrm{sust}}(x;H)
=
\hat m_{\mathrm{sust}}(x;H)
=
\frac{\hat C_{\mathrm{life}}(x;H)}{\hat Y(x;H)}.
\]
If all feasible designs are required to deliver the same functional output, the
co-design problem can equivalently minimize the estimated total life-cycle footprint
$\hat C_{\mathrm{life}}(x;H)$.

\subsubsection{Life-cycle carbon footprint}

Product carbon-footprint accounting typically includes raw materials,
manufacturing, transportation, use, and end-of-life treatment
\cite{iso14067,ghgprotocol2011product}. For soft robots, these life-cycle
stages should be resolved at the module level because local damage to a soft
component may either require only local repair or force replacement of a larger
assembly. Let $M_\mathrm{rep}(x)$ denote the set of replaceable modules or components of the
soft robot, such as soft fingers, tendon-driven segments, pneumatic chambers,
valves, electronics, sensors, batteries, pumps, or structural frames. We
decompose the life-cycle carbon footprint as
\[
C_{\mathrm{life}}(x;H)
=
\sum_{j\in M_\mathrm{rep}(x)}
n_j(x;H)
\left(
C_j^{\mathrm{make}}(x)
+
C_j^{\mathrm{eol}}(x)
\right)
+
C^{\mathrm{use}}(x;H)
+
C^{\mathrm{repair}}(x;H)
+
C^{\mathrm{trans}}(x;H).
\]
Here, $C_j^{\mathrm{make}}$ is the carbon footprint of manufacturing module $j$,
$C_j^{\mathrm{eol}}$ is the end-of-life footprint of module $j$,
$C^{\mathrm{use}}$ is the use-phase footprint, $C^{\mathrm{repair}}$ is the
footprint of repair events, and $C^{\mathrm{trans}}$ accounts for transportation
and logistics. All terms are expressed in $\mathrm{kg\,CO_2e}$; hence no
arbitrary weights are introduced.

The expected number of manufactured copies of module $j$ over the mission
horizon is
\[
n_j(x;H)
=
1+
\sum_{k\in K(x)}
A_{jk}(x)R_k(x;H).
\]
The first term accounts for the initially manufactured module. The term
$R_k(x;H)$ denotes the expected number of replacement-triggering failures of
module $k$, while $A_{jk}(x)$ describes how a failure of module $k$ propagates
through the assembly structure. Specifically,
\[
A_{jk}(x)=
\begin{cases}
1, & \text{if failure of module } k \text{ requires replacing module } j,\\
0, & \text{otherwise}.
\end{cases}
\]
Fractional entries may be used to describe probabilistic repair policies. Thus,
a modular design has mostly diagonal entries in $A(x)$, whereas a monolithic or
permanently bonded design may require many modules to be replaced after a single
local failure. This connects the carbon metric to circular-design principles such
as maintenance, repair, reuse, and material recovery
\cite{ellenmacarthur2015mci,saidani2019taxonomy}.

The expected number of replacement-triggering failures of module $k$ is modeled as
\[
R_k(x;H)
=
\int_0^H
\lambda_k(h,x)
\left(
1-p_k^{\mathrm{repair}}(h,x)
\right)
\,dh,
\]
where $h$ denotes either time or cycle count, $\lambda_k(h,x)$ is the damage or
failure hazard rate, and $p_k^{\mathrm{repair}}(h,x)$ is the probability that the
damage can be repaired without replacing the module. This term captures the
sustainability benefit of durable, repairable, modular, patchable, or
self-healing soft robotic materials and components. Self-healing soft robots
\cite{terryn2017selfhealing}, on-demand degradable soft robotic materials
\cite{oh2023photodegradable}, and broader efforts toward sustainable soft robotics
\cite{giordano2023sustainable} provide examples of design directions that affect
$\lambda_k$, $p_k^{\mathrm{repair}}$, or end-of-life treatment.

\subsubsection{Manufacturing footprint}

The manufacturing footprint of module $j$ is
\[
C_j^{\mathrm{make}}(x)
=
C_j^{\mathrm{mat}}(x)
+
C_j^{\mathrm{fab}}(x)
+
C_j^{\mathrm{tool}}(x).
\]
The material footprint is
\[
C_j^{\mathrm{mat}}(x)
=
\sum_{i\in I}
\epsilon_i^{\mathrm{mat}}
m_{ij}^{\mathrm{in}}(x)
+
\sum_{i\in I}
\epsilon_i^{\mathrm{scrap}}
m_{ij}^{\mathrm{scrap}}(x),
\]
where $I$ denotes the set of materials, $\epsilon_i^{\mathrm{mat}}$ is the
upstream material carbon intensity in $\mathrm{kg\,CO_2e/kg}$, and
$\epsilon_i^{\mathrm{scrap}}$ is the additional carbon intensity associated with
scrap treatment, recycling processing, incineration, or landfill. The total
material input is
\[
m_{ij}^{\mathrm{in}}(x)
=
m_{ij}^{\mathrm{part}}(x)
+
m_{ij}^{\mathrm{neg}}(x)
+
m_{ij}^{\mathrm{support}}(x)
+
m_{ij}^{\mathrm{failed}}(x).
\]
Here, $m_{ij}^{\mathrm{part}}$ is the material retained in the final robot,
$m_{ij}^{\mathrm{neg}}$ is negative material removed during fabrication,
$m_{ij}^{\mathrm{support}}$ denotes support, mold, or sacrificial material, and
$m_{ij}^{\mathrm{failed}}$ is the expected material lost due to failed builds or
quality-control rejection. If $\zeta_j(x)\in(0,1]$ denotes the fabrication yield
of module $j$, one possible estimate is
\[
m_{ij}^{\mathrm{failed}}(x)
=
\left(
\frac{1}{\zeta_j(x)}-1
\right)
\left(
m_{ij}^{\mathrm{part}}(x)
+
m_{ij}^{\mathrm{neg}}(x)
+
m_{ij}^{\mathrm{support}}(x)
\right).
\]

The fabrication energy footprint is
\[
C_j^{\mathrm{fab}}(x)
=
\sum_{\nu\in\mathcal{E}}
\gamma_\nu
E_{\nu j}^{\mathrm{fab}}(x),
\]
where $\mathcal{E}$ indexes energy carriers such as electricity, heat, natural
gas, compressed air, or process fuel. The coefficient $\gamma_\nu$ is the carbon
intensity of energy carrier $\nu$. The consumed fabrication energy is computed as
\[
E_{\nu j}^{\mathrm{fab}}(x)
=
\int_0^{T_j^{\mathrm{fab}}(x)}
P_{\nu j}^{\mathrm{fab}}(t,x)\,dt.
\]
This term can include printer operation, curing, heating, vacuum pumps,
compressors, post-processing, and cleaning. Finally, tooling emissions are
amortized over the expected batch size $B_j$:
\[
C_j^{\mathrm{tool}}(x)
=
\frac{\bar{C}_j^{\mathrm{tool}}(x)}{B_j},
\]
where $\bar{C}_j^{\mathrm{tool}}$ is the footprint of molds, fixtures, jigs, or
other reusable process hardware. In practice, material and process emission
factors can be obtained from life-cycle inventory databases, environmental
product declarations, supplier-specific product carbon footprints, or public
electricity-emission-factor databases such as eGRID for U.S. electricity
\cite{wernet2016ecoinvent,epa2026egrid}.

\subsubsection{Use-phase footprint}

The use-phase footprint is
\[
C^{\mathrm{use}}(x;H)
=
\sum_{\nu\in\mathcal{E}}
\gamma_\nu
E_{\nu}^{\mathrm{use}}(x;H)
+
\sum_{c\in C}
\epsilon_c^{\mathrm{cons}}
m_c^{\mathrm{cons}}(x;H),
\]
where $C$ indexes consumables such as compressed gas cartridges, hydraulic
fluids, lubricants, cleaning agents, sterile covers, disposable skins, adhesives,
or replacement batteries. The use-phase energy is
\[
E_{\nu}^{\mathrm{use}}(x;H)
=
\sum_{\ell=1}^{N_H}
\int_0^{T_\ell(x)}
P_{\nu}^{\mathrm{use}}(t,x)\,dt.
\]
For electrically powered robots, this can be written as
\[
E_{\mathrm{elec}}^{\mathrm{use}}(x;H)
=
\sum_{\ell=1}^{N_H}
\int_0^{T_\ell(x)}
\left(
P_{\mathrm{elec}}^{\mathrm{act}}(t,x)
+
P_{\mathrm{elec}}^{\mathrm{aux}}(t,x)
\right)
dt,
\]
where $P_{\mathrm{elec}}^{\mathrm{act}}$ is the actuation-related electrical
power and $P_{\mathrm{elec}}^{\mathrm{aux}}$ includes sensing, computation,
communication, cooling, lighting, and other auxiliary loads. The corresponding
carbon footprint is
\[
C^{\mathrm{use}}(x;H)
=
\gamma_{\mathrm{elec}}
E_{\mathrm{elec}}^{\mathrm{use}}(x;H)
+
\sum_{c\in C}
\epsilon_c^{\mathrm{cons}}
m_c^{\mathrm{cons}}(x;H).
\]
The electricity carbon intensity $\gamma_{\mathrm{elec}}$ should be selected
according to the deployment region, time horizon, and accounting convention; for
U.S. applications, eGRID provides regional electricity emission rates that are
commonly used for electricity carbon-footprint calculations
\cite{epa2026egrid}.

\paragraph{Example: tendon-driven actuation.}

For tendon-driven soft robots and continuum robots, the actuation term can be
instantiated directly from tendon forces and tendon velocities. Tendon-driven
continuum robots are commonly modeled using tendon routing, generalized
coordinates, and virtual-work relationships between actuator forces and
generalized forces \cite{rucker2011statics,burgnerkahrs2015continuum}. Let
$q\in\mathbb{R}^{n}$ denote the generalized coordinates of the reduced-order
robot model and let $u\in\mathbb{R}^{m}$ denote the tendon tensions. Suppose the
generalized actuation force is
\[
\tau(q,x)=A_\mathrm{u}(q,x)u,
\]
where $A_\mathrm{u}(q,x)\in\mathbb{R}^{n\times m}$ is the tendon actuation matrix. By
virtual work \cite{murray1994mathematical,siciliano2009robotics}, the robot-side
tendon power is
\[
P_{\mathrm{ten}}^{\mathrm{rob}}(t,x)
=
\dot{q}^{\top}A(q,x)u
=
u^{\top}v_{\mathrm{ten}}(q,\dot{q},x),
\]
where
\[
v_{\mathrm{ten}}(q,\dot{q},x)
=
A^{\top}(q,x)\dot{q}
\]
is the tendon contraction velocity. If $l(q,x)$ denotes the physical tendon
length, the sign convention may instead give
\[
\dot{l}(q,\dot{q},x)
=
-A^{\top}(q,x)\dot{q}.
\]
Thus, $A^{\top}(q,x)\dot{q}$ should be interpreted as positive tendon shortening
velocity.

Let $r_k(x)$ denote the spool radius of tendon $k$ and let $g_k(x)$ denote the
gear ratio,
\[
g_k(x)=\frac{\omega_{\mathrm{m},k}}{\omega_{\mathrm{s},k}},
\]
where $\omega_{\mathrm{m},k}$ is the motor shaft speed and
$\omega_{\mathrm{s},k}$ is the spool speed. The motor speed is then
\[
\omega_{\mathrm{m},k}(t,x)
=
g_k(x)\frac{v_{\mathrm{ten},k}(t,x)}{r_k(x)}.
\]
The spool torque required to generate tendon tension $u_k$ is
\[
\tau_{\mathrm{s},k}(t,x)
=
r_k(x)u_k(t),
\]
and the corresponding motor torque can be approximated by
\[
\tau_{\mathrm{m},k}(t,x)
=
\frac{r_k(x)u_k(t)}
{g_k(x)\eta_{\mathrm{gb},k}(x)},
\]
where $\eta_{\mathrm{gb},k}$ is the gearbox efficiency. A more detailed model can
make $\eta_{\mathrm{gb},k}$ depend on torque, speed, and direction of power flow.

The electrical power drawn by the tendon actuation system is then
\[
P_{\mathrm{elec}}^{\mathrm{ten}}(t,x)
=
\sum_{k=1}^{m}
\Pi_k\left(
\tau_{\mathrm{m},k}(t,x),
\omega_{\mathrm{m},k}(t,x)
\right),
\]
where $\Pi_k$ is a motor-driver power map. This map can be obtained from
datasheets, bench-top measurements, or a motor model. If such a map is not
available, a first-order approximation is
\[
\Pi_k(\tau_{\mathrm{m},k},\omega_{\mathrm{m},k})
=
P_{0,k}
+
R_{\Omega,k}
\left(
\frac{\tau_{\mathrm{m},k}}{K_{t,k}}
\right)^2
+
\frac{1}{\eta_{\mathrm{drv},k}}
\left[
\tau_{\mathrm{m},k}\omega_{\mathrm{m},k}
\right]_+
-
\eta_{\mathrm{reg},k}
\left[
\tau_{\mathrm{m},k}\omega_{\mathrm{m},k}
\right]_- ,
\]
with
\[
[z]_+=\max(z,0),
\qquad
[z]_-=\max(-z,0).
\]
Here, $P_{0,k}$ captures idle electronics and holding losses, $R_{\Omega,k}$ is
the motor winding resistance, $K_{t,k}$ is the motor torque constant,
$\eta_{\mathrm{drv},k}$ is the driver efficiency, and $\eta_{\mathrm{reg},k}$ is
the regenerative braking efficiency. For non-regenerative tendon actuation, one
sets $\eta_{\mathrm{reg},k}=0$, so negative mechanical work is dissipated rather
than credited.

The tendon-specific use-phase energy over the mission horizon is
\[
E_{\mathrm{elec}}^{\mathrm{ten}}(x;H)
=
\sum_{\ell=1}^{N_H}
\int_0^{T_\ell(x)}
P_{\mathrm{elec}}^{\mathrm{ten}}(t,x)\,dt.
\]
The full electrical use-phase energy is then
\[
E_{\mathrm{elec}}^{\mathrm{use}}(x;H)
=
E_{\mathrm{elec}}^{\mathrm{ten}}(x;H)
+
E_{\mathrm{elec}}^{\mathrm{aux}}(x;H),
\]
where $E_{\mathrm{elec}}^{\mathrm{aux}}$ accounts for sensing, computation,
communication, and other auxiliary electrical loads. This formulation makes the
sustainability metric directly sensitive to tendon routing, moment arms, spool
radii, gear ratios, actuator efficiencies, controller choice, and task-specific
motion.

\subsubsection{Repair footprint}

Repairability reduces replacement but may still consume materials and energy.
The repair footprint over the mission horizon is
\[
C^{\mathrm{repair}}(x;H)
=
\sum_{k\in K(x)}
\int_0^H
\lambda_k(h,x)
p_k^{\mathrm{repair}}(h,x)
C_k^{\mathrm{repair}}(h,x)
\,dh,
\]
where
\[
C_k^{\mathrm{repair}}(h,x)
=
\sum_{\nu\in\mathcal{E}}
\gamma_\nu
E_{\nu k}^{\mathrm{repair}}(h,x)
+
\sum_{i\in I}
\epsilon_i^{\mathrm{mat}}
m_{ik}^{\mathrm{repair}}(h,x).
\]
For a self-healing elastomer, $E_{\nu k}^{\mathrm{repair}}$ may represent the
heat, light, or electrical energy required to trigger healing
\cite{terryn2017selfhealing}. For conventional repair, $m_{ik}^{\mathrm{repair}}$
may include patches, adhesives, tubing, replacement seals, or surface coatings.

\subsubsection{End-of-life footprint}

The end-of-life footprint of module $j$ is
\[
C_j^{\mathrm{eol}}(x)
=
\sum_{i\in I}
m_{ij}^{\mathrm{part}}(x)
\left[
\alpha_{ij}^{\mathrm{sep}}(x)
\left(
r_i\epsilon_i^{\mathrm{rec}}
+
(1-r_i)\epsilon_i^{\mathrm{disp}}
-
r_iq_i\epsilon_i^{\mathrm{sub}}
\right)
+
\left(
1-\alpha_{ij}^{\mathrm{sep}}(x)
\right)
\epsilon_i^{\mathrm{mixed}}
\right].
\]
Here, $\alpha_{ij}^{\mathrm{sep}}(x)\in[0,1]$ is the fraction of material $i$ in
module $j$ that can be separated into a recoverable material stream. This
depends on the assembly graph, adhesives, fasteners, embedded sensors, coatings,
and disassembly process. The parameter $r_i$ is the recycling or reuse fraction,
$q_i$ is the substitution yield of recovered material, $\epsilon_i^{\mathrm{rec}}$
is the recycling-process carbon intensity, $\epsilon_i^{\mathrm{disp}}$ is the
disposal intensity, $\epsilon_i^{\mathrm{mixed}}$ is the end-of-life intensity of
nonseparable mixed-material waste, and $\epsilon_i^{\mathrm{sub}}$ is the avoided
burden of displaced virgin material. If a conservative cut-off approach is
desired, the avoided-burden term $r_iq_i\epsilon_i^{\mathrm{sub}}$ can be set to
zero. This term is conceptually related to product circularity metrics, but it is
kept in carbon units rather than converted into a dimensionless circularity score
\cite{ellenmacarthur2015mci,saidani2019taxonomy}.

\subsubsection{Transportation footprint}

If transportation is non-negligible, it can be modeled as
\[
C^{\mathrm{trans}}(x;H)
=
\sum_{s\in S}
\kappa_s
D_s(x;H)
M_s(x;H),
\]
where $s$ indexes shipping modes, $\kappa_s$ is the carbon intensity per
mass-distance, $D_s$ is the transport distance, and $M_s$ is the transported
mass. This term can include material supply, part shipment, replacement modules,
maintenance logistics, and end-of-life transport.

\subsubsection{Resulting co-design metric}

The final sustainability metric used in computational co-design is therefore the estimate
\[
\hat c_{\mathrm{sust}}(x;H)
=
\frac{
\sum_{j\in M_\mathrm{rep}(x)}
\hat n_j(x;H)
\left(
\hat C_j^{\mathrm{make}}(x)
+
\hat C_j^{\mathrm{eol}}(x)
\right)
+
\hat C^{\mathrm{use}}(x;H)
+
\hat C^{\mathrm{repair}}(x;H)
+
\hat C^{\mathrm{trans}}(x;H)
}
{\hat Y(x;H)}.
\]
This estimator is physically grounded because it is constructed from material
masses, fabrication energies, actuation energies, lifetimes, repair
probabilities, recycling rates, and carbon-intensity factors. It avoids
arbitrary sustainability weights while still exposing the main design levers:
material choice, negative and sacrificial material use, fabrication yield,
tooling, actuation efficiency, controller energy consumption, durability,
repairability, modularity, and end-of-life separability.

The exact value of $c_{\mathrm{sust}}$ is generally only available after product-level accounting and realized operation. In early computational co-design, $\hat c_{\mathrm{sust}}$ can be computed from CAD volumes, bill-of-materials data, process plans, reduced-order actuation models, motor power maps, fatigue or reliability models, and life-cycle inventory databases. Prototyping and measurement can then refine the uncertain quantities, such as fabrication yield, actual motor power, failure rates, and repair success probabilities.

\suppnote{Case Study for Indoor Tomato Harvesting}\label{si:sec:case_study}
\subsection{Motivation and Current Gaps}

Before introducing the co-design pipeline for tomato harvesting, it is important to
highlight the deficiencies of existing solutions and clarify how holistic co-design
addresses them, as well as the limitations that remain intrinsic.

\subsubsection{Deficiencies of Current Fruit Harvesting Solutions}
\begin{itemize}
    \item \textbf{Fragile manipulation}: Rigid grippers or simple suction cups often bruise delicate fruits or fail to detach them cleanly \cite{bac2014harvesting} \cite{elfferich2022soft} \cite{wang2017development}.
    \item \textbf{Limited adaptability}: Current robotic harvesters struggle with crop variability (clustered growth, occlusions by leaves) and require extensive hand-tuning \cite{shamshiri2024digitalization}.
    \item \textbf{Sequential design}: Traditional design processes separate hardware, perception, and control development, leading to poor integration and inefficient iteration \cite{oliveira2021advances}.
    \item \textbf{Sim-to-real gap}: Designs validated in simulation frequently underperform in real greenhouse conditions (lighting, humidity, variability) \cite{jun2021towards, kootstra2021selective}.
    \item \textbf{Cost and scalability issues}: Many prototypes are too expensive, complex to maintain, or fail to meet hygiene and food-contact regulations \cite{bethi2023challenges}.
\end{itemize}

\subsubsection{Challenges Addressed by Holistic Co-Design}
\begin{itemize}
    \item \textbf{Integrated body--brain optimization}: By jointly designing morphology, grippers, and control, co-design can yield manipulators that are both gentle and effective in dense tomato clusters.
    \item \textbf{Probabilistic treatment of uncertainty}: Co-design formalizes uncertainty (e.g., in bruise prediction or detachment force) and guides targeted prototyping, narrowing the sim-to-real gap.
    \item \textbf{Stakeholder-driven requirements}: Including growers, operators, and safety assessors from the outset ensures that labor costs, usability, and compliance are integrated early, avoiding late redesigns.
    \item \textbf{Balanced refinement vs.\ realization}: Computational design is continuously informed by physical twins (e.g., artificial peduncles), preventing costly over-prototyping while reducing reliance on purely simulated metrics.
    \item \textbf{Biological variability}: Tomato cultivars differ in peduncle strength, growth patterns, and ripening characteristics, which cannot be fully standardized within any design framework. In simulation and computational evaluation, this variability can be addressed through \emph{domain randomization}, i.e., randomly sampling across these biological differences to ensure that the design performs robustly across most cultivars.

\end{itemize}

\subsubsection{Challenges Not Fully Solved by Co-Design}
\begin{itemize}
  
    \item \textbf{Unpredictable environments}: Greenhouse conditions (condensation, sudden humidity spikes, lighting changes) will always introduce residual uncertainty.
    \item \textbf{Hardware maturity}: Some limitations, such as long-term durability of soft actuators or cleaning under high humidity (IP65+), depend on material science advances beyond co-design.
    \item \textbf{Economic constraints}: Market acceptance and cost--benefit thresholds are influenced by labor markets and crop economics, which no technical framework can fully control.
\end{itemize}

To illustrate how the holistic co-design pipeline is applied in practice, we 
consider the case of an autonomous system for harvesting tomatoes in 
greenhouse environments. Indoor tomato production presents specific 
challenges: fruits grow in clusters at varying heights, aisles are narrow, 
and environmental conditions include high humidity, condensation, and 
delicate plant structures such as support strings and stems. The system 
must therefore combine precise perception, gentle manipulation, and safe 
navigation within tight constraints.

\subsection{Holistic Co-Design Components}
In the following, we provide a step-by-step procedure on how holistic co-design could be adopted in this case and how the components necessary for our proposed framework can be derived and implemented.

\subsubsection{Overview}
The starting point is to capture the primary system-level specifications within a \emph{Library of Design Specifications}. In a typical greenhouse, aisles are about 50--60\,cm wide, which constrains the mobile base geometry and center-of-gravity limits, while storage must safely accommodate bins of 10--15\,kg. The manipulator is required to be lightweight to minimize plant disturbance yet capable of reaching up to 2.5\,m to access high-wire clusters. The gripper design employs compliant, food-safe materials to minimize bruise risk during detachment and handling.

Beyond these physical constraints, the system is guided by explicit \emph{design values} that define the targets of the holistic co-design process. These include: (i) \textbf{fruit integrity and food safety}, ensuring bruise-free handling with certified food-contact materials; (ii) \textbf{throughput and efficiency}, maximizing harvested fruits per hour within the mechanical and energetic limits of the platform; (iii) \textbf{energy and resource sustainability}, minimizing energy per pick and consumables use; (iv) \textbf{operational robustness}, guaranteeing safe operation across biological and environmental variability; and (v) \textbf{maintainability and cleanability}, allowing rapid disassembly and hygienic operation under high-humidity conditions.

Each of these design values is associated with a computational metric—such as pick success rate, bruise probability, cycle time, or energy per pick—that can be probabilistically evaluated and progressively refined through targeted realization, as detailed in \Cref{subsec:harvesting:metrics}. This establishes the quantitative foundation upon which the co-design process optimizes both functional performance and deployability.

A reduced-order design space is sampled and decoded into full CAD/BOM 
models, and surrogate models are used to predict detachment success 
(force vs.\ peduncle strength), bruise risk (contact pressure proxies), 
and energy per pick. 

\paragraph{Closed-loop operation}
Closed-loop control is essential in this environment: perception informs updates of the robot’s motion plan, adjusting for fruit pose uncertainty, leaf occlusions, and small plant motions.

\paragraph{Evaluation Procedure}  
Evaluation proceeds in three stages: (i) bench tests with physical twins; (ii) pilot trials in a controlled greenhouse section with representative occlusions and lighting; and (iii) full operational pilots in commercial greenhouses. Aligned with the concept of \emph{Realization}, each stage is used to measure the chosen evaluation metrics under increasingly realistic conditions and to iteratively update the probabilistic belief over these metrics, thereby refining the system’s expected performance in real-world deployment.

\paragraph{Audit and compliance}
From the beginning, the design record documents requirements, decisions, and results. Food-contact compliance is ensured by selecting materials certified under food-grade regulations (e.g., EU 10/2011 or FDA CFR 21). In practice, this restricts the palette of usable materials to specific silicones (platinum-cured, medical or food-grade), thermoplastic elastomers (e.g., TPU with certification), and certain polyurethanes and fluoropolymers. Popular materials in soft robotics that are not allowed include tin-cured silicones, Dragon Skin\textsuperscript{\textregistered}, Ecoflex\textsuperscript{\textregistered}, and many non-coated fabrics or composites, as they may leach plasticizers or absorb contaminants.

Cleaning procedures are integrated into the design to ensure hygienic operation. This requires (i) smooth and non-porous surfaces to prevent biofilm formation, (ii) chemical compatibility with standard sanitization agents such as ethanol, isopropanol, hydrogen peroxide, or chlorine-based solutions, (iii) modular components that can be rapidly disassembled for sterilization, and (iv) sealing strategies (e.g., overmolding, welded joints) that prevent ingress of dirt or fluids. Additionally, actuation chambers and fluidic lines must be designed to avoid stagnant zones that are difficult to flush.

Co-design is necessary because food safety and cleanability constraints cannot be simply applied after the functional prototype exists. Instead, they must be jointly embedded in the selection of materials, actuator geometries, surface finishes, and connection mechanisms. For example, a gripper shape that optimizes for grasping alone may introduce crevices that are impossible to sanitize; conversely, redesigning with cleaning in mind may reduce grasp efficiency unless both aspects are optimized simultaneously. Thus, compliance, functionality, and maintainability are achieved only through an integrated co-design process.

\paragraph{Known Limitations and Efficiency Levers}
Known issues are leaf and branch occlusions, variable peduncle 
strength across varieties, generalization to new tomato cultivars, 
and delays due to bin cycles. The framework addresses these by using 
surrogate models early, closed-loop contact evaluation, informative 
prototype selection, and joint optimization of throughput and fruit 
quality.

\subsubsection{Stakeholders and TRL Gates}
Stakeholders include the grower (focused on labor costs and yields), 
operators (ease of use and maintenance), agronomists (knowledge of 
crop characteristics), safety assessors (risk analysis), and retailers 
(concerned with consistent fruit quality). Development is aligned with 
TRL gates: TRL-3 at bench, TRL-5 in greenhouse pilots, and TRL-7 in 
pre-series trials with service contracts.

In our holistic co-design setting, stakeholder input is not collected as informal feedback but translated into explicit design values (i.e., functionalities, resources, and constraints), informing the choice of the design space and motivating the definition of useful design priors through a structured workflow led by a cross-functional product development team (mechatronics/robotics, soft-materials \& fabrication, perception/ML, control, systems engineering, and regulatory/QA). We operationalize this workflow with three recurring activities, executed at project start and repeated (with tighter thresholds) at each TRL gate.

\paragraph{(A) Design values and requirements elicitation (stakeholder workshop).}
The team runs a facilitated workshop with growers/operators/agronomists/safety assessors/retailers to elicit: (i) \emph{primary outcomes} (e.g., revenue-relevant throughput and quality), (ii) \emph{failure modes} (e.g., bruising, stem damage, contamination risk, missed fruit, plant damage, downtime), and (iii) \emph{operational constraints} (aisle geometry, shift patterns, cleaning routines, allowable chemicals, humidity, lighting, acceptable maintenance intervals). Each item is then converted into a measurable statement with a target, tolerance, and test condition (e.g., ``bruise rate $\le 2\%$ after 24/48h under standard storage''; ``success $\ge 90\%$ in clusters with occlusion level X''; ``washdown time $\le 10$ min per shift''). The outputs are stored in the \emph{Library of Design Specifications} with traceability IDs linking stakeholder needs $\rightarrow$ engineering requirements $\rightarrow$ verification protocols.

\paragraph{(B) Definition of the co-design optimization problem (design requirements $\rightarrow$ co-design optimization problem).}
The product team decomposes the design specifications/requirements into:
\emph{(1) functionalities and resources} optimized along the Pareto front (e.g., fruits/h, energy per pick, cost-per-output),
\emph{(2) acceptance constraints} (often reliability constraints e.g., $\Pr[\text{bruise}\le 0.02]\ge 0.95$), and
\emph{(3) strict feasibility constraints} that act as filters (e.g., certified food-contact materials, cleanability geometry rules, IP rating constraints, electrical safety).

\paragraph{(C) Design values-to-metrics mapping (design values $\rightarrow$ probabilistic metrics).}
For each metric, the team defines (i) a computational proxy/surrogate (simulation, learned model, or analytical estimate), (ii) the uncertainty model (i.e., prior belief) resulting from the sim-to-real gap, and (iii) an experimental protocol that produces ground-truth measurements at the current TRL (bench twins $\rightarrow$ pilot greenhouse $\rightarrow$ pre-series). This turns ``values'' (e.g., sustainability, usability, hygiene) into quantities the co-design loop can evaluate and progressively refine via realization.

\paragraph{(D) Design-space definition \& governance (what the optimizer may change).}
The team explicitly defines the decision variables and their bounds, grouped by subsystem, e.g.:
(i) mobile base (footprint, mass distribution, turning radius, bin handling),
(ii) manipulator (reach, compliance/stiffness distribution, payload margin, routing),
(iii) end-effector (contact geometry, variable stiffness mechanism, finger materials/coatings, suction+finger hybridization),
(iv) sensing/perception (sensor modalities, placements, lighting strategy, calibration routines),
(v) control (impedance/admittance parameters, detachment strategies, recovery behaviors), 
and (vi) hygiene/maintenance design (sealing strategy, surface finishes, disassembly time, avoid-crevice constraints).
This governance step also fixes which parameters are \emph{frozen} at a given TRL (to reduce complexity) and which are \emph{free} to optimize, enabling staged co-design rather than an unbounded search.

\paragraph{TRL gates as decision points.}
At each TRL gate the same stakeholders participate in a structured review where the team: (i) updates metric beliefs using realized data, (ii) tightens constraints and narrows uncertainty bounds, and (iii) selects a preferred point on the Pareto front using an explicit preference model (e.g., weighted utility agreed with the grower/retailer, or ``must-hit'' constraints + secondary objectives). Concretely, TRL-3 prioritizes the feasibility of gentle detachment and basic hygiene compliance; TRL-5 prioritizes robustness to occlusions/lighting/humidity and maintainability; TRL-7 prioritizes deployability metrics (uptime, cleaning time, service intervals) and evidence suitable for certification and commercial pilots.

\subsubsection{Design Space}
We define the co-design variables as a single vector $x\in\mathcal{X}$ that spans the subsystems that must be tuned together in a greenhouse harvester: the mobile base and bin handling, the arm, the end-effector, the sensing/perception layout, the control stack, and hygiene/maintainability. 
We optimize over 
\begin{equation}
    x=(x_\mathrm{base},x_\mathrm{arm},x_\mathrm{ee},x_\mathrm{sense},x_\mathrm{ctrl},x_\mathrm{hyg}),
\end{equation}
where $x_\mathrm{base}$ covers footprint/wheelbase (fit within 50--60\,cm aisles), turning radius and speed limits, and the bin interface and placement for 10--15\,kg loads with center-of-gravity margins; $x_\mathrm{arm}$ includes reach (up to $\sim$2.5\,m), link lengths/DoF, mass budget, compliance placement (e.g., at the wrist), and hose/cable routing to reduce snagging on strings and canopy; $x_\mathrm{ee}$ sets finger layout and geometry, contact materials/coatings, optional suction integration, stiffness-tuning settings, and the detachment primitive (twist--pull vs.\ cut-assisted); $x_\mathrm{sense}$ selects and places the sensors (e.g., RGB-D, close-up stem view, optional NIR), plus tactile/F\&T ranges and protective housings; $x_\mathrm{ctrl}$ includes navigation and contact-control parameters (e.g., impedance gains, detachment thresholds, recovery behaviors); and $x_\mathrm{hyg}$ encodes sealing, surface finish, connectors, and serviceability (e.g., disassembly time proxies) together with chemical compatibility for cleaning. 

To keep optimization manageable, we progressively freeze validated blocks at TRL gates (e.g., base envelope early, material/sealing choices after pilots) and continue refinement in a reduced-order subset (typically end-effector geometry, compliance distribution, and control gains).

\subsubsection{From Design Values To Co-Design Functionalities, Resources, and Constraints}
To avoid ambiguity, we separate \emph{objectives} (optimized along the Pareto front) from \emph{acceptance requirements} (soft constraints) and \emph{strict feasibility constraints} (hard filters that discard non-compliant designs).

\paragraph{Strict constraints }
These capture non-negotiable conditions for deployment in commercial greenhouses: the platform must fit within 50--60\,cm aisles (including turning and passing clearances), satisfy IP65+ protection for humidity and washdown, use certified fruit-contact materials and compatible sealing/assembly methods, and respect transport/handling limits for moving through greenhouse corridors. Designs that violate food-contact or hygiene constraints are discarded rather than ``optimized''.

\paragraph{Acceptance requirements}
These are belief-dependent deployability gates rather than Pareto objectives: they specify minimum quality, safety, hygiene, and operational levels agreed with growers, operators, and downstream buyers.
These reflect minimum quality and operational levels agreed with growers, operators, and downstream buyers, and are implemented as reliability constraints that can be tightened across TRL gates. Examples include cluster pick success $\Pr[\text{success}\ge 0.90]\ge 0.95$, selectivity $\Pr[\text{selectivity}\le 0.03]\ge 0.95$, bruising $\Pr[\text{bruise}\le 0.02]\ge 0.95$ measured at 24/48\,h, and endurance $\Pr[\text{autonomy}\ge 16\text{ h}]\ge 0.95$ (allowing battery swaps). Additional acceptance requirements can include limits on plant disturbance (forces near stems/strings), maximum allowable cycle-time variance, and bounded cleaning time per shift.

\paragraph{Objectives}
Within the feasible set, we optimize farm-level utility by maximizing throughput (fruits/h) while minimizing energy per pick and an \emph{economic viability} objective. We express the latter as a cost-per-picked-fruit metric that aggregates CAPEX amortization, OPEX (energy, consumables), expected maintenance and cleaning labor, and downtime. Quality premiums (e.g., reduced bruising) enter either as (i) a penalty term in cost-per-output or (ii) an explicit revenue uplift in the viability model.

In a co-design formulation~\cite{zardini2023co}, the harvesting throughput represents a \emph{functionality} the system provides while the costs are \emph{resources} that the developed design requires.

\paragraph{Economic Viability}
Economic viability is assessed through a cost-per-output model. Savings are computed per kg or per bin, annualized over expected harvest volume, and compared against CAPEX and OPEX. Additional checks include utilization within the harvest window, maintenance scheduling, and the value of reduced bruising/defects. This model is used as an optimization objective (or as an acceptance constraint via break-even time), and its parameters are refined at TRL reviews as operational data becomes available.

\subsubsection{Metrics}\label{subsec:harvesting:metrics}
In our framework, we estimate design values probabilistically via (estimated) metrics that represent computational surrogates of the real-world behavior that can be calibrated via prototyping and experimental validation. In \Cref{tab:harvesting:metrics}, we specify some metrics that could be adopted in this case, including an associated experimental protocol to obtain the \emph{true} value of the metric value/cost, which enables reducing the uncertainty in the computational metrics via targeted realization.

\begin{table}[h!]
\centering
\footnotesize
\caption{Probabilistic evaluation metrics with experimental protocols and computational implementations.}
\label{tab:harvesting:metrics}
\begin{tabular}{p{3cm} p{6cm} p{6cm}}
\hline
\textbf{Metric} & \textbf{Experimental protocol} & \textbf{Computational implementation} \\
\hline
Pick success rate & Count successful picks over total attempts in greenhouse trials. & Simulate grasp dynamics and apply success probability models under domain-randomized variability. \\
\hline
Selectivity error (ripe vs.\ unripe) & Compare picked fruits with human-annotated ripeness ground truth. & Classifier accuracy on synthetic or augmented datasets representing different ripeness states. \\
\hline
Cycle time & Measure time per pick from actuation start to fruit release. & Model task scheduling with actuator and robot dynamics; stochastic timing models. \\
\hline
Throughput per hour & Number of fruits successfully harvested per hour in pilot trials. & Extrapolation from cycle time distributions under randomized task sequences. \\
\hline
Dropped fruit percentage & Ratio of fruits lost during transfer to total picked fruits. & Simulated grasp stability analysis with randomized external perturbations. \\
\hline
Bruise percentage (24/48h) & Visual inspection and pressure mapping after storage at 24/48\,h. & Contact force simulation with tissue deformation models and probabilistic damage thresholds. \\
\hline
Energy per pick & Log power consumption of actuators and robot during trials. & Sum of modeled actuator work and system idle energy under stochastic duty cycles. \\
\hline
Bin fill efficiency & Measure packing density and damage after bin loading. & Virtual bin packing simulations with randomized fruit geometries and compliance models. \\
\hline
Economic viability (cost per picked fruit / euro\,kg)
& From pilot deployments, log output (kg or fruits), cycle time, uptime/downtime, labor for operation/cleaning, consumables, and maintenance events; combine with accounting data for CAPEX amortization and energy costs.
& Stochastic cost-per-output model that propagates uncertainty from cycle-time and uptime distributions, energy-per-pick, consumables, and maintenance/cleaning time; includes amortized CAPEX and optional quality premium/penalty terms tied to bruise/defect rates. \\
\hline
\end{tabular}
\end{table}

\subsubsection{Design Priors}
From a biological standpoint, fruit harvesting behaviors in nature demonstrate precisely the kind of multimodal and adaptive interactions that soft robotics aims to reproduce. As highlighted by the review\cite{liang2025bio}, effective grasping in biological systems often arises from coupling distinct contact-force modes—enveloping, frictional, and adhesive—depending on the object’s compliance, surface texture, and attachment strength. For instance, \emph{primates} combine enveloping and frictional grasping through variable finger stiffness and torsional motion to detach soft fruits, while \emph{avian beaks} and \emph{insect appendages} employ shear-assisted and adhesive contact strategies to manipulate delicate or irregular items. Similarly, \emph{octopus suckers} and \emph{gecko toe pads} exemplify multimodal adhesion coupled with local compliance control, offering inspiration for hybrid suction–friction interfaces. These biological templates inform the co-design process by providing motion and morphology priors that blend different grasping modalities—enveloping for support, frictional for stability, and suction or adhesion for gentle retention—within a single end-effector architecture. In our framework, such priors populate the Library of Design Specifications (LDS) as candidate archetypes for the decoder, supporting optimization toward gentle yet robust detachment behaviors under biological variability in peduncle strength and fruit geometry.

\subsubsection{Perception and Sensing}
Perception combines multi-view RGB-D (eye-in-hand and fixed cameras), 
NIR or hyperspectral sensing for ripeness cues, peduncle/stem detection, 
wrist force/torque and tactile sensing for slip and force, bin fill 
estimation, and periodic calibration of both cameras and arms.

\begin{table}[h!]
\centering
\scriptsize
\setlength{\tabcolsep}{3pt} 
\caption{Perception stack for indoor tomato harvesting: modality, role, example sensors, and typical processing algorithms.}
\begin{tabular}{p{2.7cm} p{4.3cm} p{4.3cm} p{4.1cm}}
\hline
\textbf{Sensor modality} & \textbf{Function} & \textbf{Example commercial sensors} & \textbf{Typical algorithms} \\
\hline
Multi-view RGB-D / stereo & 3D fruit pose, occlusions, approach planning & Intel RealSense (e.g., D455); Stereolabs ZED2i; industrial RGB + structured light & Instance segmentation + tracking; depth fusion; 6D pose refinement \\
\hline
NIR / hyperspectral & Ripeness/defect cues under lighting variability & Specim FX10/FX17; multispectral cameras (agri/food) & Spectral features + classifier; domain adaptation; calibrated confidence \\
\hline
Close-up stem/peduncle vision & Detachment-point localization and cut/twist planning & Eye-in-hand RGB(-D); structured light & Keypoints/skeletonization; stem tracing; local surface normal estimation \\
\hline
Wrist force/torque & Contact monitoring, detachment detection, safety limits & ATI Mini45 (or similar); integrated F/T at flange & Contact-state estimation; impedance/admittance control; force thresholds \\
\hline
Tactile / fingertip sensing & Slip/contact patch; bruise-risk proxies & GelSight-type tactile; capacitive/optical tactile arrays & Slip detection; contact-area/pressure proxy; tactile servoing \\
\hline
Bin fill estimation & Packing density, drop/impact risk mitigation & Overhead RGB-D/stereo; depth camera & 3D occupancy/fill-level estimation; anomaly detection \\
\hline
\end{tabular}
\end{table}

\subsubsection{Foundation Models}
In this case study, foundation models are used \emph{only} for high-level perception-to-task decisions and operator-facing interaction, not for direct low-level control of the arm or gripper. For instance, as discussed by Stella \emph{et al.}~\cite{stella2023can}, large language models (LLMs) can augment the robotic design and operation process by supporting conceptual reasoning, task selection, and human–AI co-design workflows rather than replacing deterministic controllers. A vision–language model provides \emph{semantic scene understanding} in cluttered canopies by fusing RGB(-D) observations into actionable labels (ripe vs.\ unripe fruit, occluding leaves, visible peduncle/stem segments) and by outputting calibrated confidence levels that can trigger re-plan behaviors.  They enable \emph{task selection and constraint setting}: given a goal (e.g., ``harvest only red-ripe''), the model chooses which fruit in a cluster to attempt next and instantiates discrete action options and constraints for the downstream planner (e.g., choose twist--pull vs.\ cut-assisted, enforce a no-pull constraint when the peduncle is not visible). 

Third, they support \emph{exception handling and human-in-the-loop supervision} by generating short, auditable explanations for failure cases (``stem not localized'', ``occlusion too high'', ``ripeness uncertain'') and suggesting the next safe recovery action (reposition camera, change approach side, skip fruit). Low-level safety monitors (force/torque limits and collision checks) and contact controllers remain model-based and deterministic to preserve certifiability and predictable behavior.

\subsubsection{Refinement, Realization, and Physical Twins}
Rapid simulations (e.g., grasp stability under uncertainty) are used 
to refine candidate designs. Realization steps, such as instrumented 
fruit tests, provide ground truth and reduce uncertainty in bruise 
and detachment predictions. The balance avoids both over-reliance on 
simulation and costly over-prototyping.

The refinement path integrates \emph{physical twins}, such as 
calibrated artificial peduncles that replicate measured detachment 
forces. These reduce uncertainty in surrogate metrics before costly 
pilots. The process progresses from bench to pilot cells to 
operational pilots, where the system’s performance is validated 
in real harvest conditions.

\suppnote{Co-Design Formulation Demonstration}\label{ssec:codesign_demo}

We instantiate the holistic co-design formulation on a trimmed-helicoid (TH) tendon-driven soft manipulator. The goal is not to define a complete benchmark, but to show how one computational co-design query can jointly expose morphology, reduced-order model (ROM), and model-based controller variables while keeping the distinction between functionalities, resources, strict feasibility constraints, and acceptance requirements. We consider a representative 3D regulation task with $m$ tendons and a task coordinate $\chi\in\mathbb R^{p}$. 

\subsection{Design Variables and TH Decoder}
The implementation optimized by the co-design algorithm is
\begin{equation}
    x=
    \left(
        x_{\mathrm{morph}},
        x_{\mathrm{ROM}},
        x_{\mathrm{ctrl}}
    \right)
    \in\mathcal X .
\end{equation}
The morphology variables are chosen from the trimmed-helicoid family introduced in~\cite{guan2023trimmed},
\begin{equation}
    x_{\mathrm{morph}}
    =
    \left(
        R_h,
        h_{\mathrm w},
        w,
        \alpha,
        d_{1:m},
        N,
        N_{\mathrm{helix}}
    \right),
\end{equation}
where $R_h$ is the helicoid radius, $w$ is the strip width, $h_{\mathrm w}$ is the strip thickness, $\alpha$ is the trimming angle, $d_{1:m}$ collects the tendon-routing offsets for the $m$ tendons, $N$ is the number of TH modules, and $N_{\mathrm{helix}}$ is the number of helicoid cells per module. Representative bounds for the toy example are
\begin{equation}\label{eq:morphology_params_constraints}
\begin{aligned}
    R_h &= 30~\mathrm{mm},
    & h_{\mathrm w} &\in [1,5]~\mathrm{mm},
    & w &\in [6,30]~\mathrm{mm},\\
    \alpha &\in \{52.6^\circ,33.2^\circ,23.1^\circ,18.6^\circ\},
    & N &= 6,
    & N_{\mathrm{helix}} &= 6 .
\end{aligned}
\end{equation}
The ROM variables $x_{\mathrm{ROM}}$ specify the control-oriented finite-dimensional model. For a PCS-based ROM, they may include the number and length of segments,
\begin{equation}
    x_{\mathrm{ROM}}^{\mathrm{PCS}}
    =
    \left(n_s,\ell_1,\ldots,\ell_{n_s}\right),
    \qquad
    \ell_j>0,
    \qquad
    \sum_{j=1}^{n_s} \ell_j=L(x_{\mathrm{morph}}).
\end{equation}
For a GVS-based ROM, they may instead define the retained strain basis,
\begin{equation}
    \xi(s,t)
    \approx
    \bar\xi(s;x_{\mathrm{morph}})
    +
    \Phi(s;x_{\mathrm{ROM}}) \, q(t),
    \qquad
    \Phi=\left[\phi_1,\ldots,\phi_{n_{\mathrm{ROM}}}\right],
\end{equation}
where $\Phi$ can be selected from a candidate basis library or identified from high-fidelity simulation snapshots. The scalar $n_{\mathrm{ROM}}(x_{\mathrm{ROM}})$ denotes the resulting ROM dimension. Finally, the controller variables correspond to the feedback control gains
\begin{equation}
    x_{\mathrm{ctrl}}
    =
    \left(
        \Gamma_{\mathrm p},
        \Gamma_{\mathrm i},
        \Gamma_{\mathrm d}
    \right),
    \qquad
    \Gamma_{\mathrm p},\Gamma_{\mathrm i},\Gamma_{\mathrm d}\succeq 0.
\end{equation}
The strict feasible set $\mathcal X\subseteq\mathcal X_0$ contains non-negotiable validity conditions, including positive geometry variables, integer module and ROM dimensions, admissible tendon routing, positive semidefinite controller gains, and process-feasibility limits such as
\begin{equation}
    h_{\mathrm w}^{\min} \leq h_{\mathrm w} \leq h_{\mathrm w}^{\max},
    \qquad
    \lambda_{\mathrm{slender}}(x_{\mathrm{morph}})
    :=
    \frac{\ell_{\mathrm{mod}}(R_h,\alpha,N_{\mathrm{helix}})}{w}
    \leq \lambda_{\max}.
\end{equation}
These hard constraints are not traded against performance; designs violating them are excluded from the search.

\subsection{From TH Morphology to Simulation and ROM Dynamics}
The morphology decoder $\psi_{\mathrm{TH}}$ maps $x_{\mathrm{morph}}$ to a tendon-driven Cosserat-rod description. The backbone length is obtained by stacking TH modules,
\begin{equation}
    L(x_{\mathrm{morph}})
    =
    N\,\ell_{\mathrm{mod}}(R_h,\alpha,N_{\mathrm{helix}}),
    \qquad
    \ell_{\mathrm{mod}}
    =
    N_{\mathrm{helix}}\ell_{\mathrm{cell}}(R_h,\alpha,N_{\mathrm{helix}}).
\end{equation}
As a geometric envelope, we approximate the TH cross section by a circular annulus with $R_{\mathrm o}=R_h$ and $R_{\mathrm i}=R_h-w$,
\begin{equation}
    A_{\mathrm{geom}}
    =
    \pi\left(R_{\mathrm o}^{2}-R_{\mathrm i}^{2}\right),
    \qquad
    I_{\mathrm{geom}}
    =
    \frac{\pi}{4}\left(R_{\mathrm o}^{4}-R_{\mathrm i}^{4}\right),
\end{equation}
and use a filling factor $c_f(x_{\mathrm{morph}})\in(0,1]$ to account for the fact that the helicoidal lattice occupies only part of this annulus. This gives $\tilde A=c_fA_{\mathrm{geom}}$ and $\tilde I=c_fI_{\mathrm{geom}}$. The empirical stiffness relations reported in~\cite{guan2023trimmed},
\begin{equation}
\begin{gathered}
    \lambda_{\mathrm{stiff}}
    \approx
    12.76\left(\frac{w}{R_h}\right)^{3.42}+1.58,\\
    S_{\mathrm{bend}}(h_{\mathrm w})\approx 29.79 \, h_{\mathrm w}^{2.97},
    \qquad
    S_{\mathrm{axial}}(h_{\mathrm w})\approx 0.05 \, h_{\mathrm w}^{3.16}.
\end{gathered}
\end{equation}
are then used to instantiate the stiffness matrix $K$ for the Cosserat model. The resulting high-fidelity simulator provides closed-loop rollouts $z^{\mathrm{HF}}(t;x)$ and configurations $q^{\mathrm{HF}}(t;x)$, while the ROM specified by $x_{\mathrm{ROM}}$ provides the corresponding control-oriented rollout $z^{\mathrm{ROM}}(t;x)$.

For the controller derivation, the ROM dynamics can be written as
\begin{equation}\label{eq:codesign_demo:open_loop_dynamics}
    M(q)\ddot q
    +
    C(q,\dot q)\dot q
    +
    G(q)
    +
    K \, q
    +
    D \, \dot q
    =
    A(q)\tau,
\end{equation}
where $q\in\mathbb R^{n_{\mathrm{ROM}}}$ and $\tau\in\mathbb R^m$. For tendon actuation, the collocated coordinates are defined by $\varphi=h_\varphi(q)=[\varphi_{\mathrm a}^{\top},\varphi_{\mathrm u}^{\top}]^{\top}$ with $\varphi_{\mathrm a}=\ell_\mathrm{tend}(q)\in\mathbb R^m$ representing the tendon lengths and $\dot\varphi=J_{\mathrm{h}_\varphi}(q)\dot q$~\cite{pustina2024input,pustina2025analysis}. Given a statically feasible desired configuration $q^{\mathrm d}$ and $\varphi^{\mathrm d}=h_\varphi(q^{\mathrm d})$, we use the P-satI-D controller
\begin{equation}\label{eq:codesign_demo:controller}
\begin{split}
    \tau
    =
    \pi(t,q)
    ={}&
    \Gamma_{\mathrm p}
    \left(\varphi_{\mathrm a}^{\mathrm d}-\varphi_{\mathrm a}\right)
    -
    \Gamma_{\mathrm d} \, \dot\varphi_{\mathrm a}
    +
    \Gamma_{\mathrm i}
    \int_0^t
    \tanh\!\left(
        \gamma\left(\varphi_{\mathrm a}^{\mathrm d}(t')-\varphi_{\mathrm a}(t')\right)
    \right)
    \mathrm dt'
    \\
    &+
    S_{\mathrm a}J_{\mathrm{h}_\varphi}^{-\top}(q^{\mathrm d})
    \left(G(q^{\mathrm d})+K \, q^{\mathrm d} \right),
\end{split}
\end{equation}
where $S_{\mathrm a}=[I_m\;0]$ selects the actuated collocated coordinates. A statics-aware inverse-kinematics procedure can be used to map the task-space setpoint $\chi^{\mathrm d}$ to a feasible $q^{\mathrm d}$ for the current morphology and ROM~\cite{della2025pushing,perfetta2026reconciling}.

\subsection{Computational Metric Instantiations}
We use the cost convention of \Cref{si:note:probabilistic_codesign}: true metrics are written as $c_j=m_j(x)$ and computational estimates as $\hat c_j=\hat m_j(x)$. In a probabilistic refinement step, each estimate may be replaced by the selected risk summary $\widetilde c_{j,\iota}(x)$; in a deterministic demonstration, one can simply set $\widetilde c_{j,\iota}(x)=\hat c_j(x)$.

\paragraph{Task performance.}
For a closed-loop rollout over $H$ time steps, the task loss is
\begin{equation}
    \hat c_{\mathrm{task}}^{\mathrm{TH}}(x)
    =
    \sqrt{
        \frac{1}{H}
        \sum_{k=1}^{H}
        \left\|
            \chi^{\mathrm d}
            -
            \operatorname{FK}\!\left(q_k^{\mathrm{HF}}(x)\right)
        \right\|_{W_\chi}^{2}
    } .
\end{equation}
This metric is a cost; higher task functionality is obtained by negating it in the provided-functionality vector.

\paragraph{ROM fidelity and complexity.}
The ROM should be accurate enough for model-based control while remaining low-dimensional. We quantify prediction error by comparing high-fidelity and ROM rollouts through a common output $y$, such as sampled backbone poses, strain samples, or task-space coordinates:
\begin{equation}
    \hat c_{\mathrm{ROM}}^{\mathrm{TH}}(x)
    =
    \sqrt{
        \frac{1}{H}
        \sum_{k=1}^{H}
        \left\|
            y_k^{\mathrm{HF}}(x)
            -
            y_k^{\mathrm{ROM}}(x)
        \right\|_{W_y}^{2}
    } .
\end{equation}
The dimension $n_{\mathrm{ROM}}(x_{\mathrm{ROM}})$ is treated separately as a required resource, because increasing ROM dimension may improve fidelity but also increases sensing, state-estimation, and control complexity.

\paragraph{Safety.}
We reuse the safety metric introduced in \Cref{si:sub:safety}. For the TH manipulator and controller $\pi$, the computational safety cost is
\begin{equation}
    \hat c_{\mathrm{safety}}^{\mathrm{TH}}(x)
    =
    \widehat{\mathrm{SRDHC}}
    \left(
        x;
        \Sigma_{\mathrm c}^{\mathrm{TH}}(x,\pi)
    \right),
\end{equation}
where $\Sigma_{\mathrm c}^{\mathrm{TH}}(x,\pi)$ denotes the admissible set of closed-loop collision scenarios induced by the morphology, tendon routing, ROM-based controller, actuator limits, and operating envelope. Safety is used below as an acceptance requirement rather than reintroduced as a separate metric definition.

\paragraph{Manufacturability.}
The TH-specific manufacturability cost follows the operation-based expected-cost metric in \Cref{si:sub:manufacturability}. For a process route $\phi$, batch size $B$, and operation set
\begin{equation}
    \mathcal O_{\phi}^{\mathrm{TH}}
    =
    \{\mathrm{print},\mathrm{post},\mathrm{assemble},\mathrm{route},\mathrm{integrate},\mathrm{inspect}\},
\end{equation}
we define
\begin{equation}
    \hat c_{\mathrm{mfg}}^{\mathrm{TH}}(x_{\mathrm{morph}},\phi;B)
    =
    \frac{\hat C_{\mathrm{tool}}^{\mathrm{TH}}(x_{\mathrm{morph}},\phi)}{B}
    +
    \sum_{o\in\mathcal O_{\phi}^{\mathrm{TH}}}
    \hat\chi_o^{\mathrm{TH}}(x_{\mathrm{morph}},\phi)
    \hat C_o^{\mathrm{TH}}(x_{\mathrm{morph}},\phi).
\end{equation}
For example, the print and assembly contributions can be estimated as
\begin{align}
    \hat C_{\mathrm{print}}^{\mathrm{TH}}
    &=
    c_{\mathrm{TPU}}\rho_{\mathrm{TPU}}V_{\mathrm{TPU}}(x_{\mathrm{morph}})
    +
    a_{\mathrm{print}}^{\mathrm{equip}}\hat t_{\mathrm{print}}(x_{\mathrm{morph}},\phi)
    +
    a_{\mathrm{print}}^{\mathrm{lab}}\hat t_{\mathrm{setup,print}},
    \\
    \hat C_{\mathrm{assemble}}^{\mathrm{TH}}
    &=
    a_{\mathrm{lab}}
    \left(
        N\hat t_{\mathrm{stack}}
        +
        m\hat t_{\mathrm{route}}
        +
        \hat t_{\mathrm{preload}}
    \right),
\end{align}
with
\begin{equation}
    V_{\mathrm{TPU}}(x_{\mathrm{morph}})
    \approx
    \tilde A(x_{\mathrm{morph}})L(x_{\mathrm{morph}})
    =
    c_f(x_{\mathrm{morph}})A_{\mathrm{geom}}(R_h,w)L(x_{\mathrm{morph}}).
\end{equation}
The yield-loss multipliers retain the general form
\begin{equation}
    \hat\chi_o^{\mathrm{TH}}
    =
    \left(
        \prod_{r\succeq o}\hat y_r^{\mathrm{TH}}
    \right)^{-1},
\end{equation}
where pass probabilities can initially be simple rules depending on quantities such as strip thickness, support-removal difficulty, tendon-routing clearance, and the slenderness ratio $\lambda_{\mathrm{slender}}$. Thus, coarse hard limits enter $\mathcal X$, while residual material, labor, equipment, and yield costs enter the resource estimate $\hat c_{\mathrm{mfg}}^{\mathrm{TH}}$.

\paragraph{Durability.}
We reuse the durability framework in \Cref{si:sub:durability} and instantiate only the trajectory-induced strain-energy proxy for the TH manipulator. For a representative task horizon $H$,
\begin{equation}
    \hat c_{\mathrm{dur},E}^{\mathrm{TH}}(x)
    =
    \frac{1}{H}
    \int_0^H
    \hat E_K(t;x,\pi)\,\mathrm dt,
\end{equation}
where
\begin{equation}
    \hat E_K(t;x,\pi)
    =
    \frac{1}{2} \, \hat q(t)^{\top} \, K(x_{\mathrm{morph}}) \, \hat q(t).
\end{equation}
Here, $\hat q(t)$ is obtained from the selected simulation pipeline. Lower accumulated strain energy corresponds to lower predicted durability cost.

\subsection{Co-Design Problem Formulation}
We now map the TH-specific computational costs into the co-design quantities used in the main text. The provided functionalities are chosen as negative costs,
\begin{equation}
    \operatorname{prov}_{\mathrm{TH},\iota}(x)
    =
    \widetilde f_{\mathrm{TH},\iota}(x)
    =
    \begin{bmatrix}
        -\widetilde c_{\mathrm{task},\iota}^{\mathrm{TH}}(x)\\
        -\widetilde c_{\mathrm{ROM},\iota}^{\mathrm{TH}}(x)\\
        -\widetilde c_{\mathrm{dur},E,\iota}^{\mathrm{TH}}(x)
    \end{bmatrix}.
\end{equation}
The required resources are
\begin{equation}
    \operatorname{req}_{\mathrm{TH},\iota}(x)
    =
    \widetilde r_{\mathrm{TH},\iota}(x)
    =
    \begin{bmatrix}
        \widetilde c_{\mathrm{mfg},\iota}^{\mathrm{TH}}(x)\\
        n_{\mathrm{ROM}}(x_{\mathrm{ROM}})
    \end{bmatrix},
\end{equation}
and safety is imposed as an acceptance requirement,
\begin{equation}
    \widetilde b_{\mathrm{TH},\iota}(x)
    =
    \widetilde c_{\mathrm{safety},\iota}^{\mathrm{TH}}(x)
    -
    c_{\mathrm{safety}}^{\max}
    \leq 0 .
\end{equation}
The resulting fixed-resource, maximum-functionality query is
\begin{equation}\label{eq:codesign_demo:problem}
\begin{aligned}
    \max_{x}\quad
    &
    \operatorname{prov}_{\mathrm{TH},\iota}(x)
    \\
    \textrm{s.t.}\quad
    &
    x\in\mathcal X,
    \\
    &
    \operatorname{req}_{\mathrm{TH},\iota}(x)
    \leq
    \begin{bmatrix}
        c_{\mathrm{mfg}}^{\max}\\
        n_{\mathrm{ROM}}^{\max}
    \end{bmatrix},
    \\
    &
    \widetilde c_{\mathrm{safety},\iota}^{\mathrm{TH}}(x)
    \leq
    c_{\mathrm{safety}}^{\max}.
\end{aligned}
\end{equation}
The vector-valued maximization can be interpreted as a Pareto query or replaced by an application-specific scalarization. A chance-constrained version can be obtained by replacing the final inequality with $\Pr_{\iota}[\hat c_{\mathrm{safety}}^{\mathrm{TH}}(x)\leq c_{\mathrm{safety}}^{\max}]\geq 1-\delta_{\mathrm{safety}}$.

\subsection{Conclusion}
This example shows how the proposed holistic co-design framework can be instantiated for a TH soft manipulator without collapsing the problem into a single task-performance objective. Morphology, ROM structure, and controller gains are optimized jointly; task performance, ROM fidelity, and durability are treated as provided functionalities; manufacturability and ROM dimension are treated as required resources; and safety is imposed as an acceptance requirement. Solving and validating this full problem is beyond the scope of this perspective, but the formulation illustrates how existing modeling, control, safety, manufacturability, and durability tools can be connected within a single auditable co-design query.

  \clearpage
\fi

\end{document}